\documentclass[10pt,twocolumn,letterpaper]{article}

\usepackage{cvpr}
\usepackage{times}
\usepackage{epsfig}
\usepackage{graphicx}
\usepackage{amsmath}
\usepackage{amssymb}
\usepackage{multirow}
\usepackage{booktabs}
\usepackage[normalem]{ulem}

\usepackage{array}
\newcolumntype{C}{>{\raggedleft\let\newline\\\arraybackslash\hspace{0pt}}p{1cm}}

\usepackage[pagebackref=true,breaklinks=true,letterpaper=true,colorlinks,bookmarks=false]{hyperref}

 \cvprfinalcopy 

\def\cvprPaperID{1} 

\newcommand{\dpmvocvp}{DPM-VOC+VP\xspace}

\newcommand{\vdpm}{VDPM\xspace}
\newcommand{\rcnnmv}{RCNN-MV\xspace}
\newcommand{\cnnmv}{CNN-MV\xspace}
\newcommand{\cnn}{CNN\xspace}

\newcommand{\rcnnRidge}{RCNN-Ridge\xspace}
\newcommand{\rcnnLasso}{RCNN-Lasso\xspace}
\newcommand{\rcnnElNet}{RCNN-ElNet\xspace}
\newcommand{\rcnnl}{RCNN-L\xspace}

\newcommand{\rcnnRidgeL}{RCNN-Ridge-L\xspace}
\newcommand{\rcnn}{RCNN\xspace}
\newcommand{\keyReg}{RCNN-KeyReg\xspace}

\newcommand{\aavp}{AAVP\xspace}

\newcommand{\argmin}{\operatornamewithlimits{argmin}}

\newcommand{\myparagraph}[1]{{{\vspace{0.1cm}\textbf{#1}\quad}}}

\ifcvprfinal\fi
\begin{document}

\title{3D Object Class Detection in the Wild}

\author{Bojan Pepik$^1$\\
\and Michael Stark$^{1}$\\
\and Peter Gehler$^2$\\
\and Tobias Ritschel$^1$\\
\and Bernt Schiele$^1$\\
\and
\begin{tabular}{c}
$^1$Max Planck Institute for Informatics, $^2$Max Planck Institute for Intelligent Systems\\
\end{tabular}
}


\maketitle

\begin{abstract}
Object class detection has been a synonym for 2D bounding box
localization for the longest time, fueled by the success of powerful
statistical learning techniques, combined with robust image
representations.  Only recently, there has been a growing interest in
revisiting the promise of computer vision from the early days: to
precisely delineate the contents of a visual scene, object by object,
in 3D. In this paper, we draw from recent advances in object detection
and 2D-3D object lifting in order to design an object class detector
that is particularly tailored towards 3D object class detection. Our
3D object class detection method consists of several stages gradually
enriching the object detection output with object viewpoint, keypoints
and 3D shape estimates. Following careful design, in each stage it
constantly improves the performance and achieves state-of-the-art
performance in simultaneous 2D bounding box and viewpoint estimation
on the challenging Pascal3D+~\cite{xiang14wacv} dataset.


\end{abstract}
%
\section{Introduction}

\begin{figure}
\centering
\includegraphics[width=2.6667cm]{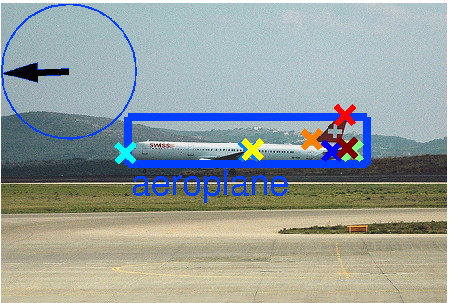}
\includegraphics[width=2.6667cm]{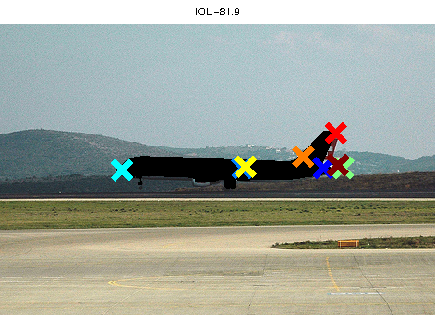}
\includegraphics[width=2.6667cm]{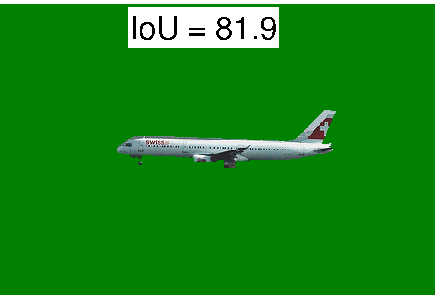}\\
 \includegraphics[width=2.6667cm]{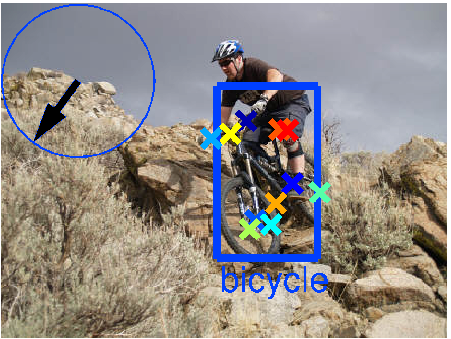}
 \includegraphics[width=2.6667cm]{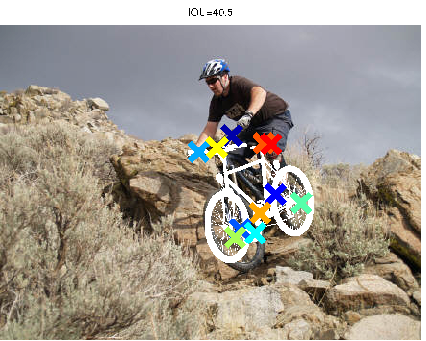}
 \includegraphics[width=2.6667cm]{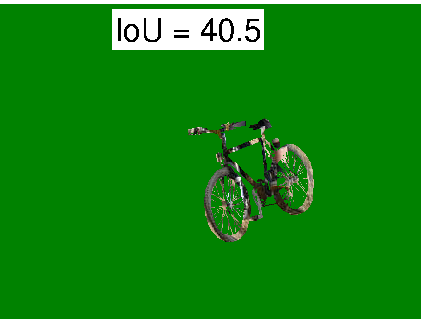}\\
 \includegraphics[width=2.6667cm]{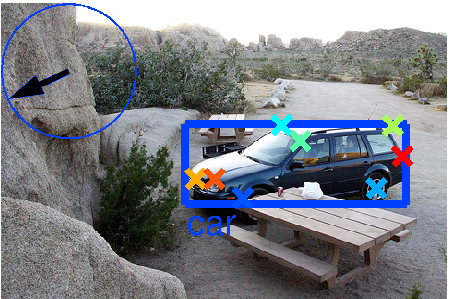}
 \includegraphics[width=2.6667cm]{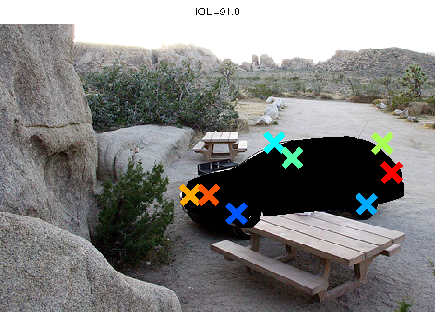}
 \includegraphics[width=2.6667cm]{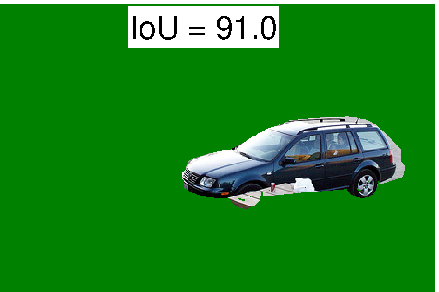}\\
 \includegraphics[width=2.6667cm]{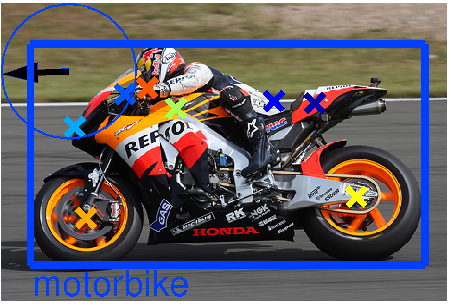}
 \includegraphics[width=2.6667cm]{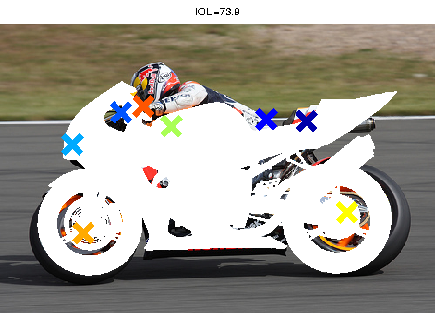}
 \includegraphics[width=2.6667cm]{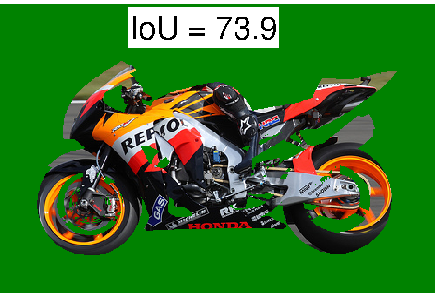}\\
 \caption{Output of our 3D object class detection method. (Left) BB,
   keypoints and viewpoint estimates, (center) aligned 3D CAD prototype,
   (right) segmentation mask.}
\label{fig:teaser}
\end{figure}

Estimating the precise 3D shape and pose of objects in a scene from
just a single image has been 
a long standing goal of computer vision since its early
days~\cite{marr78,brooks81,pentland86,lowe87}. It has been argued that
higher-level tasks, such as scene understanding or object tracking,
can benefit from detailed, 3D object
hypotheses~\cite{ess09pami,wojek10eccv,geiger14pami} that allow to
explicitly reason about
occlusion~\cite{pepik13cvpr,zia13cvpr,li14eccv} or establish
correspondences across multiple frames~\cite{xiang14eccv}. As a
consequence, there has been an increasing interest in designing object
class detectors that predict more information than just 2D bounding
boxes, ranging from additional viewpoint
estimates~\cite{stark10bmvc,gu10eccv,lopez11iccv,xiang14wacv} over 3D
parts that correspond across
viewpoints~\cite{pepik12cvpr,thomas06cvpr} to the precise 3D shape of
the object instance observed in a test
image~\cite{zia13pami,yoruk133drr,hejrati12nips}.

So far, these efforts have lead to two main results.
First, it has been shown that simultaneous 2D bounding box
localization and viewpoint estimation, often in the form of
classification into angular bins, are feasible for rigid object
classes~\cite{thomas06cvpr,savarese07iccv,liebelt08cvpr,ozuysal09cvpr,arie-nachimson09iccv,liebelt10cvpr}.
These {\em multi-view object class detectors} typically use
view-based~\cite{lopez11iccv,pepik12cvpr} or coarse 3D geometric
~\cite{xiang12cvpr,pepik12eccv,fidler12nips,hejrati12nips,hejrati14cvpr}
object class representations that are designed to generalize across
variations in object shape and appearance. While these
representations have shown remarkable performance through the use of
joint training with structured losses~\cite{pepik12cvpr,pepik12eccv},
they are still limited with respect to the provided geometric detail.

Second, and more recently, it has been shown that highly detailed 3D
shape hypotheses can be obtained by {\em aligning 3D CAD model
  instances} to an
image~\cite{zia13pami,lim13iccv,aubry14cvpr,lim14eccv}. These
approaches are based on a large database of 3D CAD models that ideally
spans the entire space of object instances expected at recognition
time. Unfortunately, the added detail comes at a cost: first, these
approaches are targeted only towards specific object classes like cars
and bicycles~\cite{zia13pami}, chairs~\cite{aubry14cvpr}, or pieces of
IKEA furniture~\cite{lim13iccv,lim14eccv}, limiting their
generality. Second, they are typically evaluated on datasets with
limited clutter and occlusion~\cite{zia13pami}, such as 3D Object
Classes~\cite{savarese07iccv}, EPFL Multi-View
Cars~\cite{ozuysal09cvpr}, or particular subsets of PASCAL
VOC~\cite{everingham10ijcv} without truncation, occlusion, or
``difficult'' objects~\cite{aubry14cvpr}.
%

In this work, we aim at joining the two directions, multi-view
detection and 3D instance alignment, into {\em 3D object class
  detection in the wild} -- predicting the precise 3D shape and pose
of objects of various classes in challenging real world images.
We achieve this by combining a robust, part-based object class
representation based on RCNNs~\cite{girshick14cvpr} with a small
collection of 3D prototype models, which we align to the observed
image at recognition time. The link between a 2D image and a 3D
prototype model is established by means of 2D-3D keypoint
correspondences, and facilitated by a pose regression step that
precedes rigid keypoint alignment.
%

As a result, the presented method predicts the precise 3D shape and
pose of all PASCAL3D+~\cite{xiang14wacv} classes
(Fig.~\ref{fig:teaser}), at no loss in performance with respect to 2D
bounding box localization: our method improves over the previous best
results on this dataset~\cite{felzenszwalb10pami} by $21.2\%$ in
average precision (AP) while simultaneously improving $12.5\%$ in AAVP
(Sect.~\ref{sec:expLifting}) in joint object localization and
viewpoint estimation~\cite{pepik12cvpr}. In addition, projecting the
3D object hypotheses provided by our system onto the image plane
result in segmentation masks that are competitive with native
segmentation approaches, highlighting the accuracy of our 3D shape
estimates.

This paper makes the following contributions.
First, to our knowledge, we present the first method for 3D object
class detection in the wild, achieving precise 3D shape and pose
estimation at no loss of 2D bounding box localization accuracy
compared to state-of-the-art RCNN detectors. 
%
Second, we design a four-stage detection pipeline that is explicitly
tailored towards 3D object class detection, based on a succession of
{\em (i)} robust 2D object class detection, {\em (ii)} continuous
viewpoint regression, {\em (iii)} object keypoint detection and {\em
  (iv)} 3D lifting through rigid keypoint alignment. 
Third, we give an in-depth experimental study that validates the
design choices at each stage of our system. Crucially, and in
contrast to previous work~\cite{pepik12cvpr,pepik12eccv}, we
demonstrate that 
enriching the output of the object detector 
does not incur any
performance loss: the final 3D detections yield the same AP as stage
(i) and improved AAVP over stage (ii), even though significant
geometric detail is added.
And fourth, we demonstrate superior performance compared to
state-of-the-art in 2D bounding box localization, simultaneous
viewpoint estimation, and segmentation based on 3D prototype
alignment, on all classes of the PASCAL3D+ dataset~\cite{xiang14wacv}.
\section{Related work}
Our approach draws inspiration from four different lines of work, each
of which we review briefly now. 

\myparagraph{2D Object class detection.}  Recently, RCNNs (regions
with convolutional neural network features) have shown impressive
performance in image classification and 2D BB
localization~\cite{girshick14cvpr}, outperforming the previous
de-facto standard, the deformable part model
(DPM)~\cite{felzenszwalb10pami}, by a large margin. Our pipeline is
hence built upon an RCNN-based detector that provides a solid
foundation to further stages. To our knowledge, our 
model is
the first to extend a
\rcnn-based object class detector towards 3D
detection.

\myparagraph{Multi-view object class detection.}  In recent years,
computer vision has seen significant progress in multi-view object
class detection. Successful approaches are mostly extensions of proven
2D detectors, such as the implicit shape
model~\cite{thomas06cvpr,yan07iccv,liebelt08cvpr,sun10eccv}, the
constellation model~\cite{su09iccv,stark10bmvc}, an the deformable
part
model~\cite{gu10eccv,lopez11iccv,pepik12cvpr,pepik12eccv,xiang12cvpr,fidler12nips,hejrati12nips,hejrati14cvpr},
resulting in both
view-based~\cite{stark10bmvc,gu10eccv,lopez11iccv,pepik12cvpr} and
integrated, 3D
representations~\cite{payet11iccv,pepik12eccv,xiang12cvpr,fidler12nips,hejrati12nips,zia13pami,yoruk133drr,hejrati14cvpr,bakry14eccv}
that reflect the 3D nature of object classes.

Our work follows a different route, and decomposes the 3D detection
problem into a sequence of simple, but specialized pipeline stages,
each optimized for performance. From the multi-view detection
literature, we take inspiration mostly from continuous viewpoint
regression~\cite{ozuysal09cvpr,gu10eccv,pepik12eccv,zia13pami,bakry14eccv},
which we use as the second stage of our pipeline. In contrast to
previous work, however, our pipeline does not end with a viewpoint
estimate, but rather uses it to guide the next stage, 3D lifting. As
we show in our experiments, the last stage benefits from the
intermediate viewpoint regression, and even improves the regressor's
estimate.

\myparagraph{3D Instance alignment.}  Methods that align 3D CAD model
instances to a test
image~\cite{zia13pami,lim13iccv,aubry14cvpr,lim14eccv} are receiving
increasing attention, due to their ability to yield highly precise
estimates of 3D object shape and pose (sometimes referred to as fine
pose estimation~\cite{lim13iccv,lim14eccv}). These methods are based
on a large number of 3D CAD model instances that are rendered from a
large set of viewpoints, in order to sufficiently cover appearance
variations. While the resulting complexity can be alleviated by
selecting discriminative exemplar patches~\cite{aubry14cvpr} or
sharing of 3D parts~\cite{lim14eccv}, it is still linear in the
cross-product of instances and viewpoints, 
limiting scalability. In contrast, we focus on capturing only the
major modes of shape variation in the form of a hand-full of
prototypical 3D CAD models per object class.
In addition, our representation is based on only a small number of 3D
keypoints (on average $10$ per object class) that are not only shared
among instances, but also matched to image evidence in a
viewpoint-invariant way~\cite{zia13cvpr}. As a result, we
can increase the accuracy of our 3D lifting stage by adding more CAD
models, without retraining our pipeline.
%
%

\myparagraph{Keypoint-based methods.}
The concept of deriving 3D information from predicted 2D keypoints is
well known in human body pose
estimation~\cite{andriluka09cvpr,bourdev09iccv}, and has also been
successfully applied to estimating the rough pose of birds
~\cite{farrell11iccv,zhang14eccv} or
fitting deformable 3D shape models~\cite{zia13pami,zia13cvpr}. Our
work draws from this idea in order to find a rigid alignment of a
prototypical 3D CAD model to an image.

\section{3D Object class detection}
\begin{figure*}[t]
\centering
\includegraphics[width=0.9\textwidth]{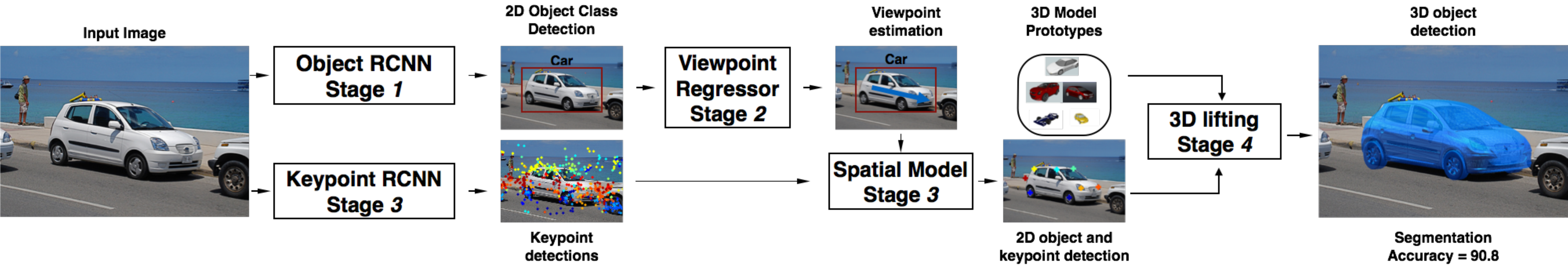}
\caption{Our 3D object class detection pipeline.}
\label{fig:modelIlustration}
\end{figure*}

In this section, we describe our 3D object class detection
pipeline. Given a single test image as an input, it can not only
predict the 2D bounding box (BB) of each object in the image, but also
yields estimates of their 3D poses as well as their 3D shape,
represented relative to a set of prototypical 3D CAD
models. Fig.~\ref{fig:teaser} gives example results. A schematic
overview of our method is shown in Fig.~\ref{fig:modelIlustration}.

The following subsections provide a walk-through of our pipeline.  We
start with robust 2D object class detection
(Sect.~\ref{sec:objdet}). We then add viewpoint information 
(Sect.~\ref{sec:view}). Next, we localize a set of 3D object keypoints
in the 2D image plane (Sect.~\ref{sec:partdet}) that provides the
basis for our last stage: 3D lifting (Sect.~\ref{sec:lift}). It
combines all estimates of the previous stages into a final, 3D object
class detection result.
Since this last step depends crucially on the quality of the
intermediate stages, we highlight the important design choices that
have to be made in each subsection.

\subsection{2D Object class detection}
\label{sec:objdet}
RCNNs~\cite{krizhevsky12nips,girshick14cvpr} have shown remarkable
performance in image classification and 2D BB localization, leading to
state-of-the-art results on the Pascal VOC~\cite{everingham10ijcv} and
ImageNet~\cite{deng09cvpr} datasets. As precise BB detection and 2D
alignment are crucial requirements for being able to infer 3D
geometry, we adopt RCNNs as the first stage of our pipeline.

Specifically, we use the implementation of Girshick et
al.~\cite{girshick14cvpr} (\rcnn).  It consists of three steps:
generation of BB proposals, feature extraction using the intermediate
layers of a CNN, and subsequent training of a one-vs-all
SVMs. 

The selective search method~\cite{uijlings13ijcv} provides several
object candidate regions $o \in {\mathcal{O}}$ in an image. These are
passed into a CNN~\cite{krizhevsky12nips} and its unit activations in
separate layers are extracted as feature representation for each
region. The \rcnn~\cite{girshick14cvpr} uses the responses of either
the last convolutional (conv5) or one of the two fully connected
layers (fc6, fc7). A linear SVM is trained for every object class,
with the positive examples being the regions with a certain
intersection-over-union (IoU) overlap $R$ with the ground truth
and the negative examples the regions with IoU $\le 0.3$ with the
ground truth.  At test time, the \rcnn provides for each image $I$ a
set of object detections $o = [o^b, o^c, o^s]$, where $o^b$ is the BB,
$o^c$ the object class, and $o^s$ the score.

Empirical results in~\cite{girshick14cvpr} on the Pascal VOC 2007 and
2010 datasets identify fc7 features and $R=1$ as the best set of
parameters. We compared the combination of intermediate feature
responses and values of $R$ on the Pascal3D+~\cite{xiang14wacv}
dataset and found the same setting to perform best.
%
%

\subsection{Viewpoint estimation}
\label{sec:view}
An essential cue for performing the transition from 2D to 3D is an
accurate estimate of the 3D pose of the object, or, equivalently, of
the viewpoint under which it is imaged.
%
We represent the viewpoint of an object $o^v\in[0,360)$ in terms of
azimuth angle $a$. 
Several approaches can be taken to obtain a
viewpoint estimate, treating it either as a discrete or continuous
quantity.
We discuss the discrete version first, mainly to be comparable with
recent work. However we argue that due to the continuous nature of the
viewpoint the problem should be treated as a continuous regression
problem.  As the experiments will show
(Sect.~\ref{sec:expObjViewpoint}), this treatment outperforms the
discrete variants allowing for a much finer resolution of the
viewpoint estimate.


\myparagraph{Discrete viewpoint prediction.}
A large body of previous work and datasets on multi-view object class
detection~\cite{savarese07iccv,glasner11iccv,pepik12cvpr,xiang14wacv}
use a discretization of the viewpoint into a discrete set of $V$
classes, typically focusing on a single angle (azimuth).  The task is
then to classify an object hypothesis into one of the
$v\in\{1,\ldots,V\}$ classes. 
While this defeats the continuous nature
of the problem, it has the benefit of giving a reduction to a
multi-class classification problem for which efficient methods exist.

We conjecture that a CNN representation will be discriminative also
for viewpoint estimation and explore two different CNN variants to
test this hypothesis. First, we use the pre-trained CNN from
Section~\ref{sec:objdet} and replace the last linear SVM layer for
object detection with one for viewpoint estimation. Discretizing the
viewpoints in $V$ classes results in $V$ different classifiers for
every object category. During test time, we choose the class with the
maximum score. We refer to this method as \rcnnmv. 
We explore a second variant (\cnnmv), a multi-view CNN trained
end-to-end to jointly predict category and viewpoint. The CNN
parameters are initialized from a network trained on
ImageNet~\cite{deng09cvpr} for object category classification and is
then trained
using logistic loss and backpropagation~\cite{jia2014caffe}.


 


\myparagraph{Continuous viewpoint prediction.} 
While discrete viewpoint prediction is the de-facto standard today, 
we believe that angular accurate viewpoint estimation is both more 
natural and leads to better performance, which is confirmed by the
empirical results in Sect.~\ref{sec:expObjViewpoint}.

We again use the intermediate layer responses of a CNN, pretrained for
detection (Section~\ref{sec:objdet}), as the feature representation
for this task. From these features, we regress the
azimuth 
angle directly. More formally, let us denote with $\phi_i$ the
features provided by a CNN on region $o_i$ depicting an object of
category $c$. Let $o^a$ represent the azimuth of the region and $w^a$
the azimuth regressor for class $c$. We use a least squares objective
\begin{equation}
w^a = \argmin_{w} ||o^a_i - \phi_i^{\top}w ||_2^2 + \lambda||w||^2_p,
\label{eq:regLearn}
\end{equation}
and test three different regularizers: ridge regression ($p=2$), lasso
($p=1$), and elastic net. 
We refer to the regressors as \rcnnRidge, \rcnnLasso and
\rcnnElNet. 
In our experiments, we found that these are the best performing
methods, confirming that the CNN features are informative for
viewpoint estimation, and that the continuous nature of the problem
should be modeled directly.
%
%

\subsection{Object keypoint detection}
\label{sec:spatial}
While an estimate of the 3D object pose in terms of azimuth angle
(Sect.~\ref{sec:view}) already conveys significant geometric
information beyond a 2D BB, it is not enough to precisely delineate a
3D prototype model, which is the desired final output of our 3D object
class detection pipeline. In order to ultimately do the lifting to 3D
(Sect.\ref{sec:lift}), our model relies on additional geometric
information in the form of object keypoints. They establish precise
correspondences between 3D object coordinates and the 2D image plane.

To that end, we design a set of object class specific
keypoint detectors that can accurately localize keypoints in the 2D
image plane. In connection with a spatial model spanning multiple
keypoints, these detectors can deliver reliable anchor points for
2D-3D lifting.



%

\myparagraph{Keypoints proposal and detection.}
\label{sec:partdet}
Recently, it has been shown that powerful part detectors can be
obtained by training full-blown object class detectors for
parts~\cite{chen14cvpr}. Inspired by
these findings, we once more turn to the \rcnn as the most powerful
object class detector to date, but train it for keypoint detection
rather than entire objects. Since keypoints have quite different characteristics in
terms of image support and feature statistics, we have to perform the
following adjustments to make this work.

First, we find that the standard \rcnn mechanism for obtaining candidate
regions, selective search~\cite{uijlings13ijcv}, is sub-optimal for
our purpose (Sect.~\ref{sec:expKeypoint}), since it provides only
limited recall for object keypoints. This is not surprising, since it
has been designed to reliably propose regions for entire objects: it
starts from a super-pixel segmentation of the test image, which tends
to undersegment parts in most cases~\cite{hosang14bmvc}.
%
We hence propose an alternative way of generating candidate regions,
by training  a separate DPM~\cite{felzenszwalb10pami} detector for
each keypoint. To generate positive training examples we need to define a BB around
each keypoint. We use a squared region centered
at the keypoint that covers $30\%$ of the relative size of the object
BB. At test time, we can then choose an appropriate number
of candidate keypoint regions by thresholding the DPM's dense sliding
window detections.
%

Second, we find that fine-tuning the CNN on task-specific training
data makes a difference for keypoint detection
(Sect.~\ref{sec:expKeypoint}). We compare two variants of RCNN
keypoint detectors, both scoring DPM keypoint proposal regions using a
linear SVM on top of  CNN features. The first variant re-uses the CNN
features trained for 2D object class detection
(Sect.~\ref{sec:objdet}). The second one fine-tunes the CNN on
keypoint data prior to feature computation. 
%
%

\myparagraph{Spatial model.}
Flexible part-based models are among the most successful
approaches for object class recognition in numerous
incarnations~\cite{fergus03cvpr,felzenszwalb05ijcv,felzenszwalb10pami},
since they constrain part positions to overall plausible
configurations while at the same time being able to adapt to
intra-class shape variation -- both are crucial traits for the 3D
lifting stage of our pipeline.
Here, we start from the spatial model suggested by~\cite{aubry14cvpr}
in the context of localizing mid-level exemplar patches, and extend it
for 3D instance alignment. This results in a simple, effective, and
computationally efficient spatial model relating object with keypoint
detections.

We define a spatial model that relates the position of keypoints to
the position of the object center in the 2D image plane, resulting in
a star-shaped dependency structure as in previous
work~\cite{leibe08ijcv,felzenszwalb10pami}. Specifically, for every
different keypoint class $p$ we estimate on the training data the
average relative position around the object center $o$.  Around this
estimated mean position we define a rectangular region $N(o,p)$ of
size proportional to the standard deviation of the relative keypoint
positions in the training set.
At test time, for a given 
object center $o$,
for every part $p$ we perform max-pooling in $N(o,p)$. This prunes out
all keypoint detections outside of $N(o,p)$ and only retains the
strongest one inside.

As the visibility and relative locations of keypoints changes
drastically with object viewpoint, we introduce a number of
viewpoint-specific components of this spatial model. During training,
these components 
are obtained by clustering the viewpoints into $C$ clusters, and
estimating the mean relative keypoint position 
on each component.

At test time we resort to two strategies to decide on which component
to use. We either use the viewpoint estimation (Sect.\ref{sec:view})
as a guidance for which one to use, or we use the one with the
best 3D detection objective (Sect.~\ref{sec:lift}). Indeed, the
guided version performs better (Sect.~\ref{sec:expKeypoint}).

\subsection{3D Object class detection}
\label{sec:lift}
The result of the previous stages is a combination of a 2D object BB
(Sect.~\ref{sec:objdet}) plus a set of 2D keypoint locations
(Sect.~\ref{sec:partdet}) specific to the object class.  Optionally,
the keypoint locations are also specific to viewpoint, by virtue of
the viewpoint estimation (Sect.~\ref{sec:view}) and the corresponding
spatial model component. This input can now be used to lift the 2D
object class detection to 3D, resulting in a precise estimate of 3D
object shape and pose.


We choose a non-parametric representation of 3D object
shape, based on prototypical 3D CAD models for the object class of
interest. Assuming known correspondences between keypoints defined on
the surface of a particular model and 2D image locations, we can
estimate the parameters of the projective transformation that gives
rise to the image.

\myparagraph{3D Lifting.}  We adopt the camera model
from~\cite{xiang14wacv} and use a pinhole camera $P$ always facing the
center of the world, assuming the object is located there.
%
Assuming a fixed field of view, the camera model
consists of 3D rotation (pose) and 3D translation parameters.  We
parameterize the 3D pose as
$o^v\in[0,360)\times[-90,+90)\times[-180,180)$, in terms of azimuth
angle $a$, elevation angle $e$ and the in-plane rotation
$\theta$. These three continuous parameters, fully specify the pose of
a rigid object. The 3D translation parameters consist of the distance
of the object to the camera $D$ and the in-plane translation $t$.
%

The 3D lifting procedure jointly estimates the camera and
the 3D shape of the object. Let us denote with $\{k^i\}$ the set of 2D
keypoint predictions. Let $\{K^i_j\}$ be the corresponding 3D keypoints
on the CAD model $j$ and $\tilde{k}^i_j=PK^i_j$ denote the image
projection of $K^i_j$. Then the CAD prototype $c^*$ and camera $P^*$
are obtained by solving
\begin{equation}
\label{eq:lift}
c^*, P^*=\argmin_{c,P}\sum_i^L ||k^i - \tilde{k}^i_c||.
\end{equation}
We perform exhaustive search over the set of CAD models and solve for
$P$ using an interior point solver as in~\cite{xiang14wacv}.

\myparagraph{Initialization.} The object viewpoint estimate is used to
initialize the azimuth. The elevation is initialized using the
category mean. We initialize $\theta=0$. For the in-plane translation
and 3D distance parameters, we solve Eq.~\ref{eq:lift} optimizing only
for these parameters.  This gives a good coarse initialization of the
distance and the in-plane translation that is used later for the joint
optimization of all parameters.
%

\section{Experiments}
\label{sec:experiments}

\begin{figure*}
\centering
\includegraphics[height=1.83cm]{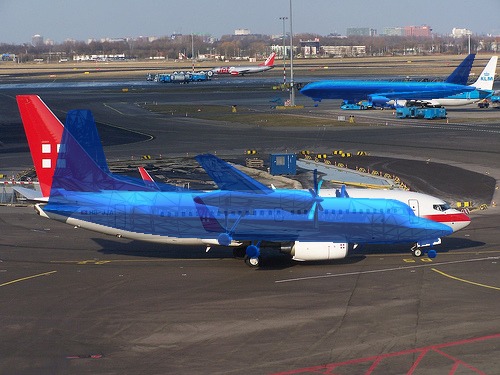}~
\includegraphics[height=1.83cm]{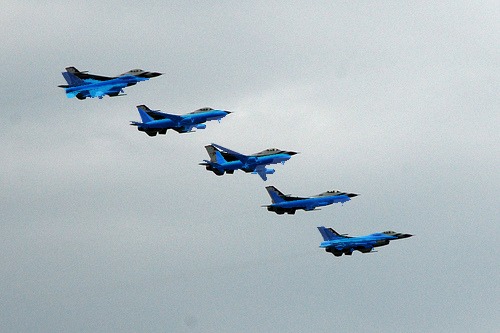}~
\includegraphics[height=1.83cm]{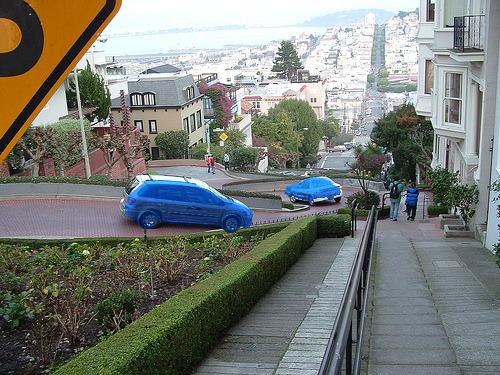}~
\includegraphics[height=1.83cm]{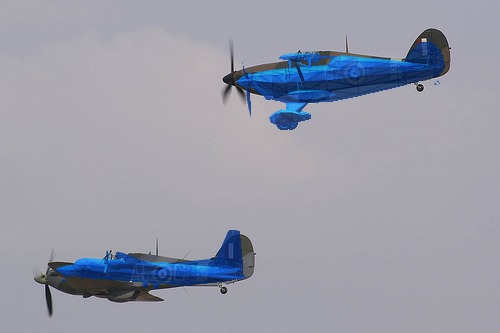}~
\includegraphics[height=1.83cm]{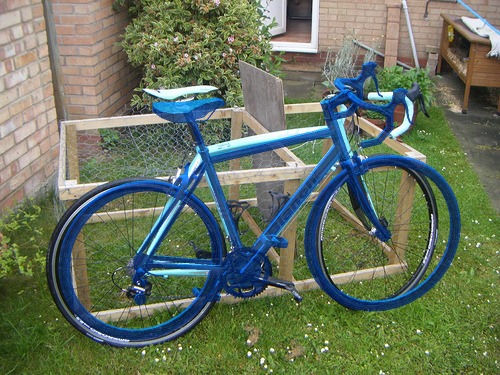}~
\includegraphics[height=1.83cm]{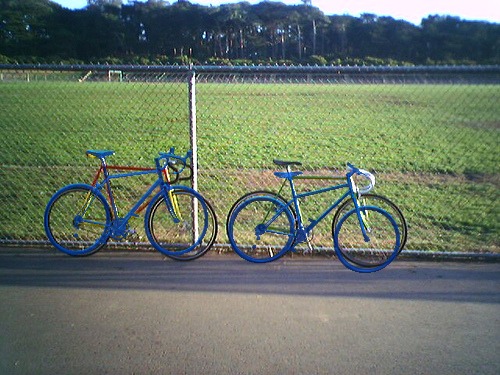}~
\includegraphics[height=1.83cm]{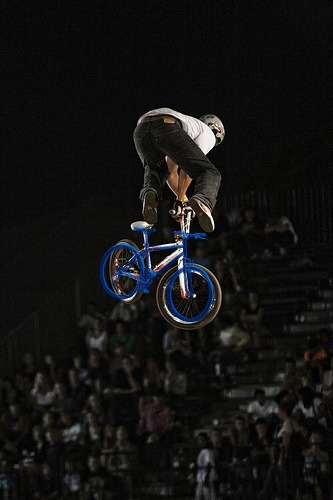}\\[2pt]
\includegraphics[height=1.9cm]{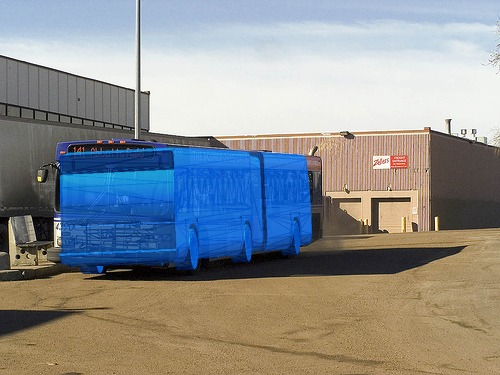}~
\includegraphics[height=1.9cm]{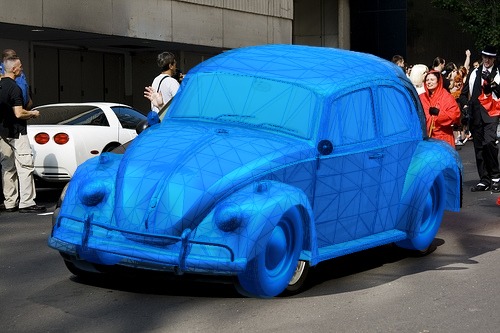}~
\includegraphics[height=1.9cm]{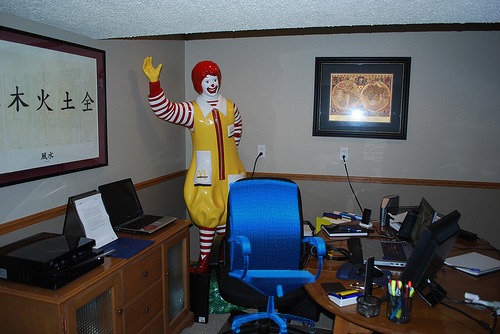}~
\includegraphics[height=1.9cm]{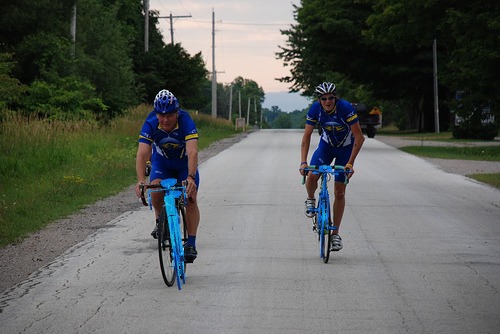}~
\includegraphics[height=1.9cm]{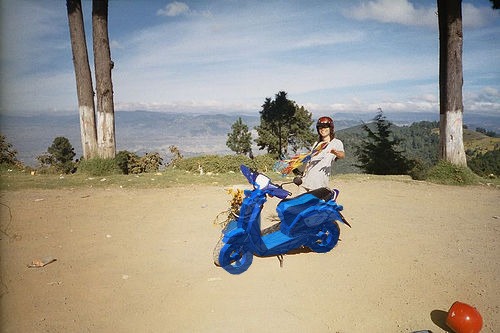}~
\includegraphics[height=1.9cm]{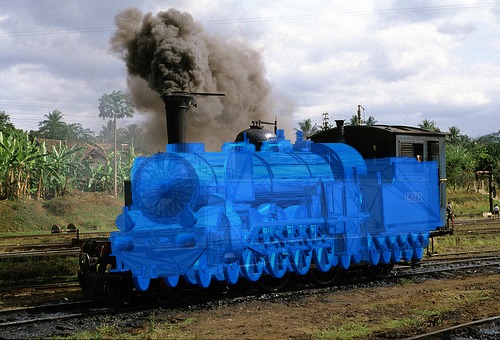}\\[2pt]
\includegraphics[height=1.8cm]{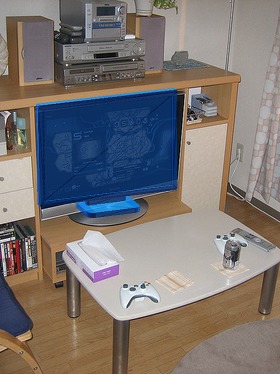}~
\includegraphics[height=1.8cm]{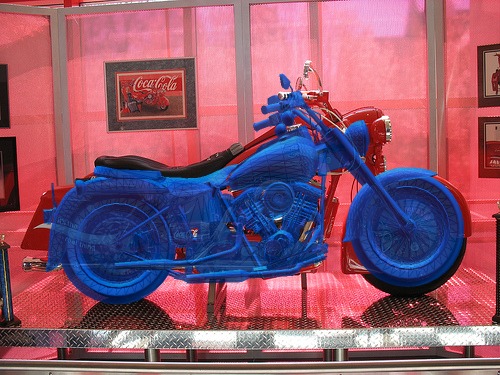}~
\includegraphics[height=1.8cm]{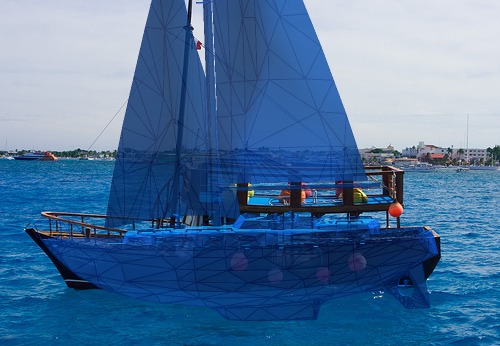}~
\includegraphics[height=1.8cm]{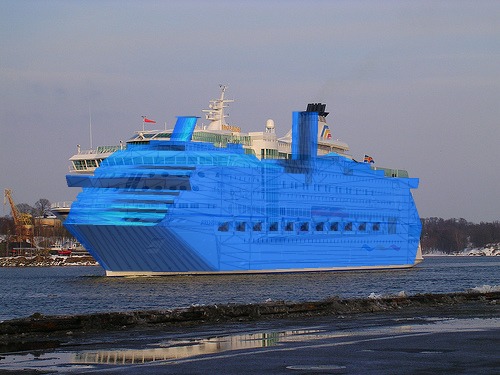}~
\includegraphics[height=1.8cm]{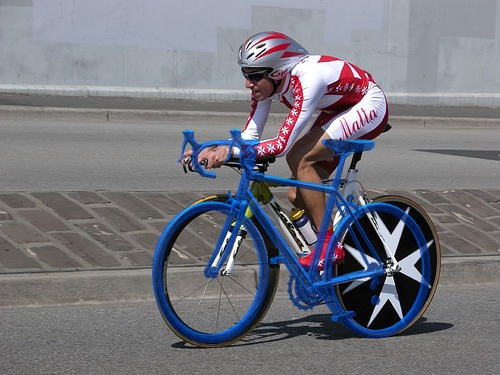}~
\includegraphics[height=1.8cm]{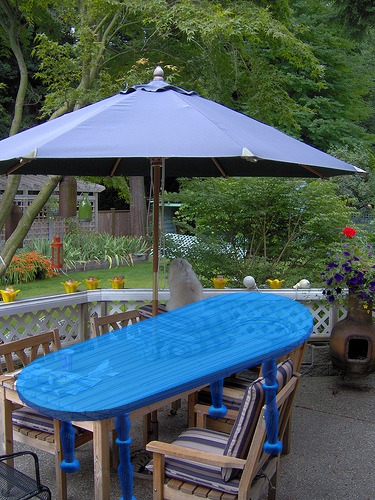}~
\includegraphics[height=1.8cm]{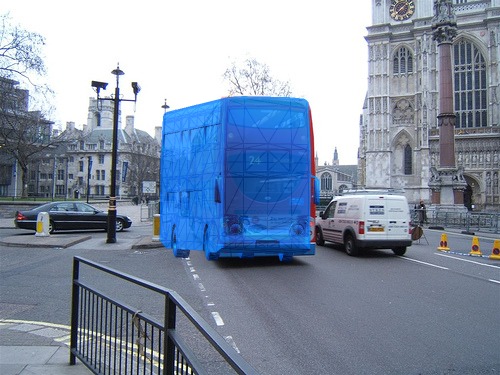}~
\includegraphics[height=1.8cm]{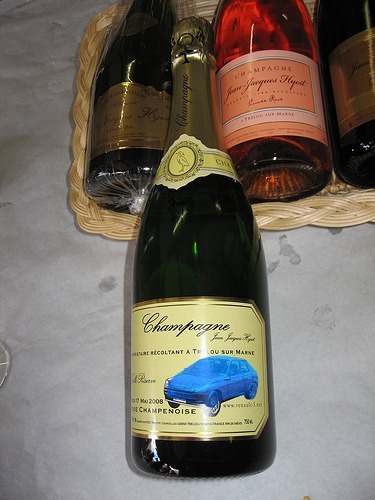}\\[2pt]
\includegraphics[height=1.897cm]{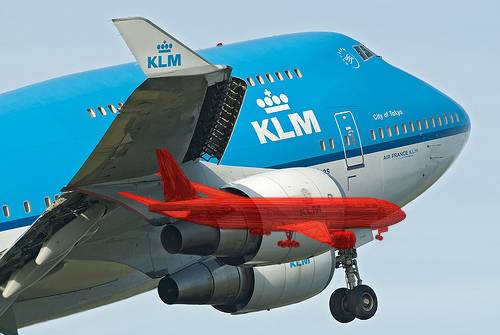}~
\includegraphics[height=1.897cm]{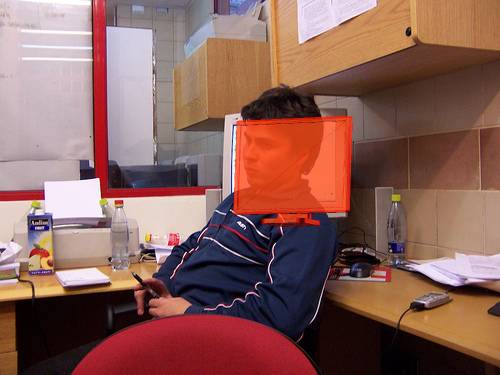}~
\includegraphics[height=1.897cm]{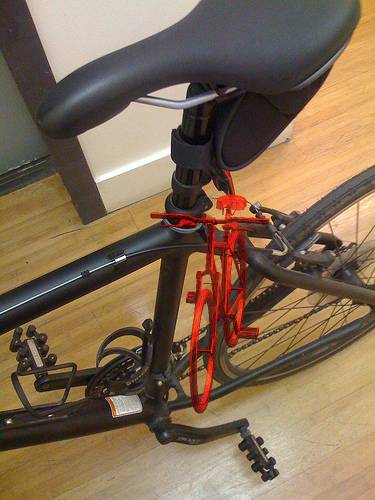}~
\includegraphics[height=1.897cm]{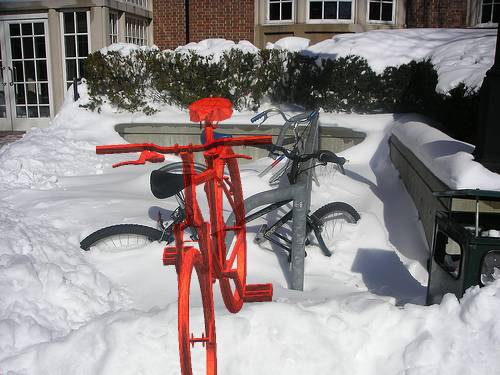}~
\includegraphics[height=1.897cm]{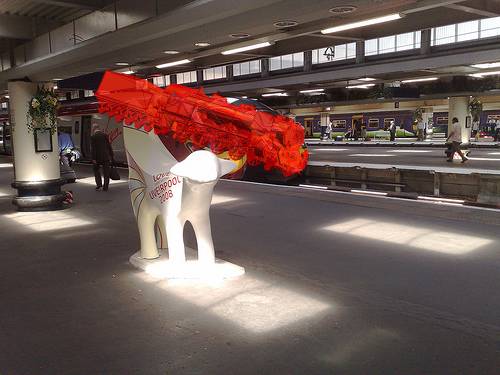}~
\includegraphics[height=1.897cm]{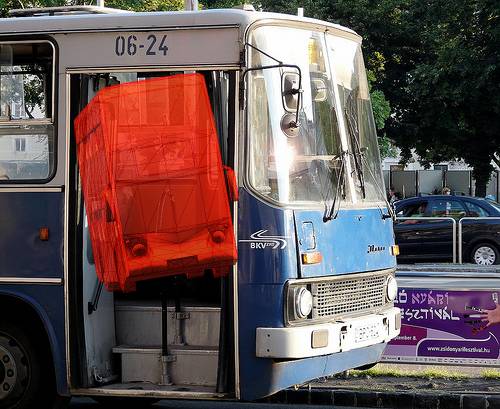}~
\includegraphics[height=1.897cm]{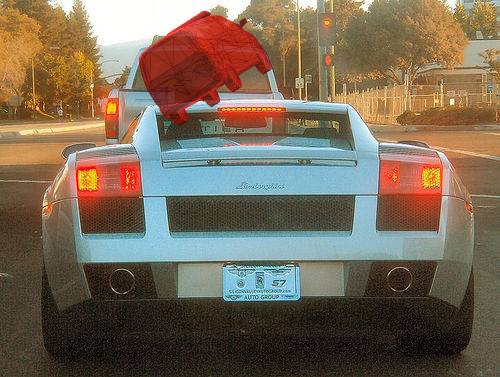}~
\caption{3D CAD prototype alignment examples. (Blue) good alignments,
  (red) bad alignments. \rcnnRidgeL fails mainly on truncated and
  occluded cases. For more 3D alignment visualizations please see the
  supplemental material.}
\label{fig:ex}
\end{figure*}

In this section, we give an in-depth experimental study of the
performance of our 3D object class detection pipeline, highlighting
three distinct aspects. First, we validate the design choices at each
stage of our pipeline, 2D object class detection
(Sect.~\ref{sec:expObjDet}), continuous viewpoint regression
(Sect.~\ref{sec:expObjViewpoint}), keypoint detection
(Sect.~\ref{sec:expKeypoint}) and 3D lifting
(Sect.~\ref{sec:expLifting}), ensuring that each stage delivers
optimal performance when considered in isolation. Second, we verify
that adding geometric detail through adding more pipeline stages does
not come at the cost of losing any performance, as it is often
observed in previous
work~\cite{liebelt10cvpr,pepik12cvpr,pepik12eccv,zia13pami}. And
third, we compare the performance of our method to the previous
state-of-the-art art, demonstrating significant performance gains in
2D BB localization, simultaneous localization and viewpoint
estimation, and segmentation based on 3D prototype alignment.
In contrast to previous
work~\cite{zia13pami,lim13iccv,aubry14cvpr,lim14eccv}, we evaluate the
performance of our method for a variety of classes on challenging,
real-world images of PASCAL VOC~\cite{everingham10ijcv,xiang14wacv}.
%
%
%

\myparagraph{Dataset.}  We focus our evaluation on the recently
proposed Pascal3D+~\cite{xiang14wacv} dataset. It enriches PASCAL VOC
2012~\cite{everingham10ijcv} with 3D annotations in the form of
aligned 3D CAD models. 
The dataset provides aligned CAD models for 11 rigid classes ({\em
  aeroplane}, {\em bicycle}, {\em boat}, {\em bus}, {\em car}, {\em
  chair}, {\em dining table}, {\em motorbike}, {\em sofa}, {\em
  train}, and {\em tv monitor}) of the {\em train} and {\em val}
subsets of PASCAL VOC 2012. 
The alignments are obtained through human supervision, by first
selecting the visually most similar CAD model for each instance, and
specifying the correspondences between a set of 3D CAD model keypoints
and their image projections, which are used to compute the 3D pose of
the instance in the image.
Note that,  while the 3D lifting stage of our pipeline
(Sect.~\ref{sec:lift}) is in fact inspired by this procedure, it is
entirely automatic, and selects the best fitting 3D CAD model
prototype without any human supervision.
Throughout the evaluation, we use the {\em train} set for training and
the {\em val} set for testing, as suggested by the
Pascal3D+~\cite{xiang14wacv}.

\myparagraph{State-of-the-art.}  We compare the performance of our
pipeline to previous state-of-the-art results on the PASCAL3D+ dataset
as reported in~\cite{xiang14wacv}. Specifically, we compare our
results to two variants of the deformable part model
(DPM~\cite{felzenszwalb10pami}) that predict viewpoint estimates in
the form of angular bins in addition to 2D BBs: {\em (i)}
\vdpm~\cite{xiang14wacv} trains dedicated mixture components for each
angular viewpoint bin, using standard hinge-loss, and {\em (ii)}
\dpmvocvp~\cite{pepik12cvpr} optimizes mixture components jointly
through a combined localization and viewpoint estimation
loss~\footnote{The \dpmvocvp detections were provided by the authors
  of~\cite{pepik12cvpr}.}. This method has been shown to outperform
previous work in multi-view detection by significant margins on 3D
Object Classes~\cite{savarese07iccv} and PASCAL VOC 2007 cars and
bicycles.
%
%

\subsection{2D Bounding box localization}
\label{sec:expObjDet}
%
%
We start by evaluating the first stage of our pipeline, 2D object
class detection (Sect.~\ref{sec:objdet}), in the classical 2D BB
localization task, as defined by PASCAL
VOC~\cite{everingham10ijcv}. Fig~\ref{exp:vp} (left) compares the
performance of our \rcnn in its discrete multi-view variant \rcnnmv
(cyan) to \cnnmv (green) and the state-of-the-art methods on this
dataset, \vdpm~\cite{xiang14wacv} (blue) and
\dpmvocvp~\cite{pepik12cvpr} (light blue). It reports the mean average
precision (mAP) over all 11 classes of Pascal3D+ (per-class results
are part of the supplemental material) for different numbers of
discrete azimuth bins, as suggested by the PASCAL3D+ benchmark:
VP$_1$, VP$_4$, VP$_8$, VP$_{16}$ and VP$_{24}$ denote the number of
discrete viewpoint-dependent components of the respective model. Note
that for the VP$_1$ case, the \vdpm model reduces to the standard
DPM~\cite{felzenszwalb10pami} and \rcnnmv to the standard \rcnn.

\myparagraph{Results.}
We make the following observations.
First, for VP$_1$, both \rcnn ($51.2\%$) and \cnn ($47.6\%$)
outperform the previous state-of-the-art result of \vdpm ($29.6\%$) by
significant margins of $21.6\%$ and $18.0\%$, respectively, in line
with prior reports concerning the superiority of CNN- over DPM-based
detectors~\cite{girshick14cvpr}.
Second, we observe that the performance of \vdpm and \dpmvocvp remains
stable or even slightly increases when increasing the number of
components (e.g., from $29.6\%$ to $30.0\%$ for \vdpm and from
$27.0\%$ to $28.3\%$ for \dpmvocvp and VP$_{16}$). Curiously, this
tendency is essentially inverted for \rcnn and \cnn: performance drops
dramatically from $51.2\%$ to $30.8\%$ and from $47.6\%$ to $27.6\%$
for AP$_{24}$, respectively.

\myparagraph{Conclusion.}  We conclude that, while the training of
per-viewpoint components is a viable strategy for DPM-based methods,
\rcnnmv and \cnnmv both suffer from the decrease in training data
available per component. We hence elect \rcnn as the first stage of
our 3D detection pipeline, leaving us with the need for another
pipeline stage capable of estimating viewpoint.
%

\subsection{Simultaneous 2D BB and viewpoint estimation}
\label{sec:expObjViewpoint}
\begin{figure*}[t]
\centering
\includegraphics[width=0.32\textwidth]{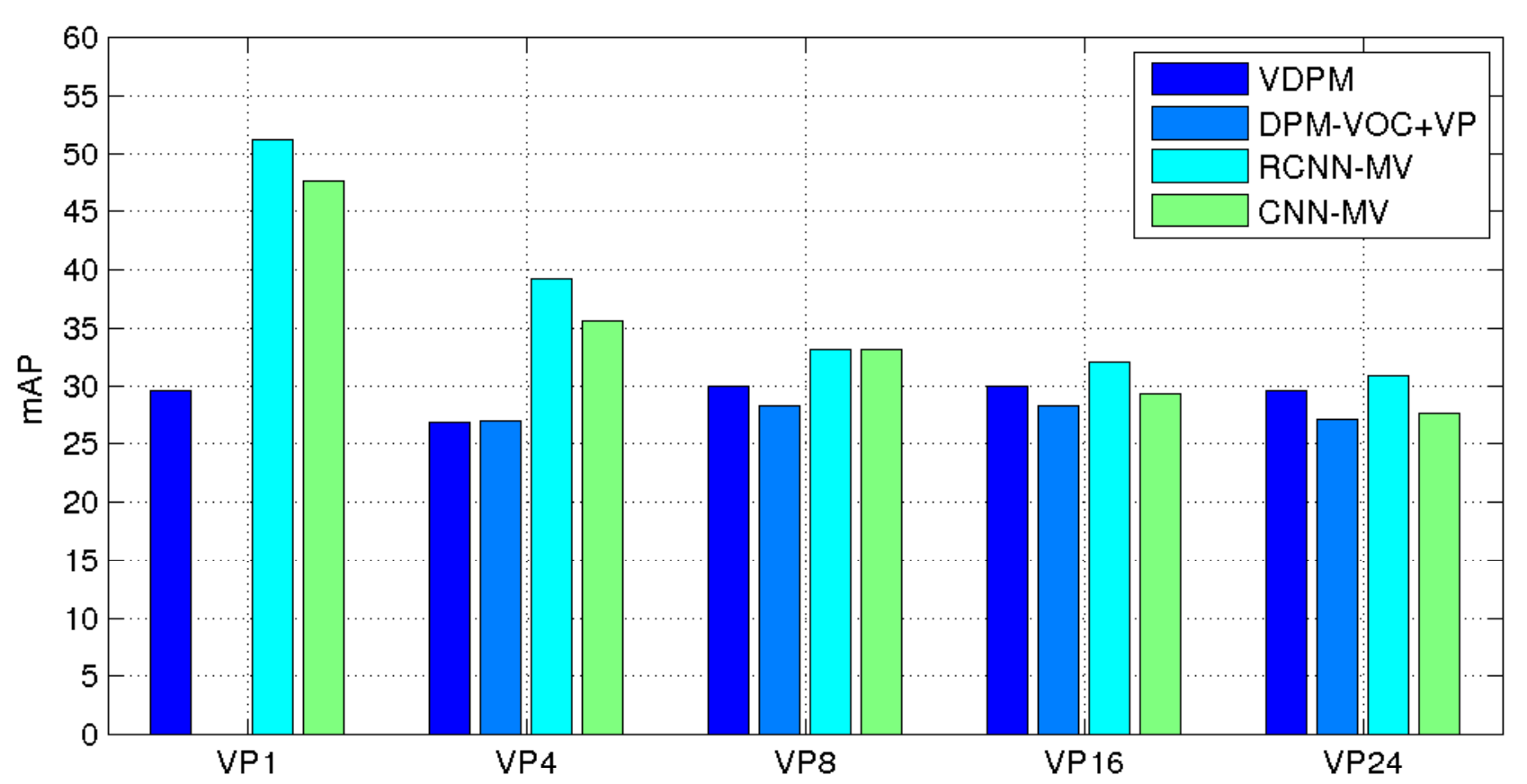}
\includegraphics[width=0.35\textwidth]{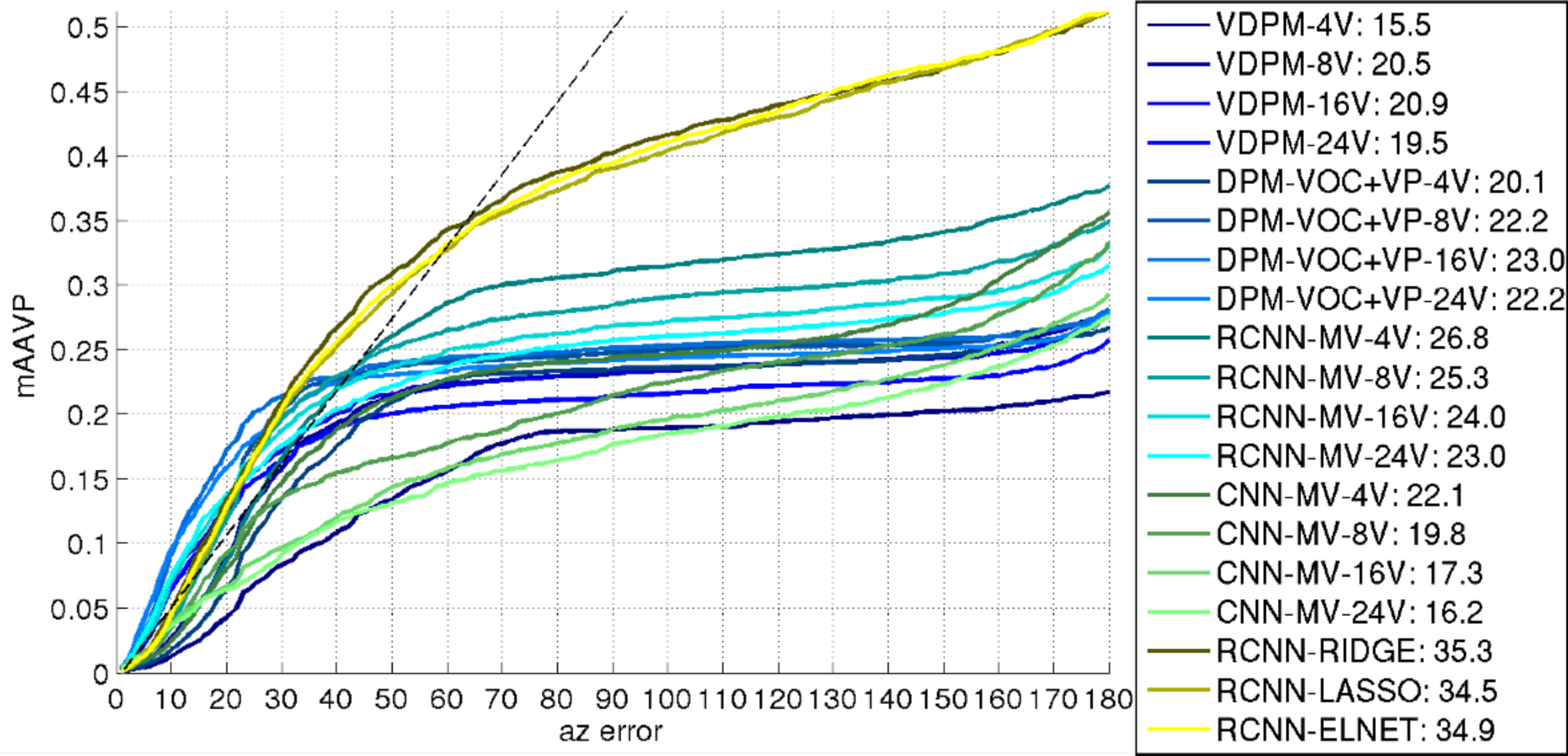}
\includegraphics[width=0.31\textwidth]{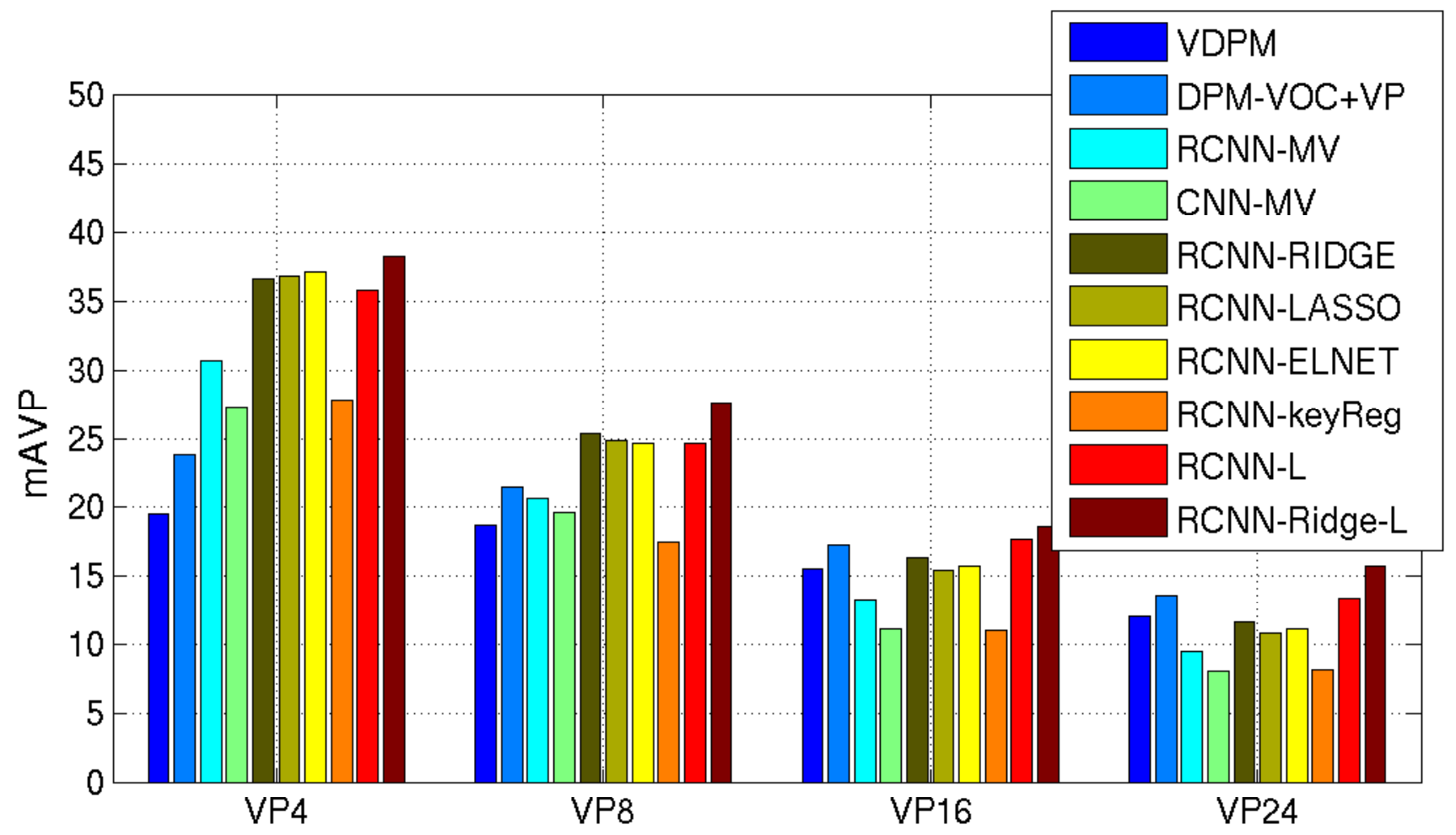}
\caption{(Left) 2D BB localization on
  Pascal3D+~\cite{xiang14wacv}. (Center, right) Simultaneous 2D
  BB localization and viewpoint estimation. (Center)
  continuous mAAVP performance, (right) discrete mAVP performance for
  VP$_{4}$, VP$_{8}$, VP$_{16}$ and VP$_{24}$.}
\label{exp:vp}
\end{figure*}
The original PASCAL3D+ work~\cite{xiang14wacv} suggests to quantify
the performance of simultaneous 2D BB localization and
viewpoint estimation via a combined measure, average viewpoint
precision (AVP). It extends the traditional PASCAL
VOC~\cite{everingham10ijcv} detection criterion to only consider a
detection a true positive if it satisfies both the IoU BB
overlap criterion {\em and} correctly predicts the ground truth
viewpoint bin (AVP $\le$ AP). This evaluation is repeated for different numbers of
azimuth angle bins VP$_4$, VP$_8$, VP$_{16}$ and
VP$_{24}$. While this is a step in the right direction, we believe
that viewpoint is inherently a continuous quantity that should be
evaluated accordingly. We hence propose to consider the entire
continuum of possible azimuth angle errors $D \in
[0^{\circ},\ldots,180^{\circ}]$, and count a detection as a true
positive if it satisfies the IoU and is within $D$ degrees of the
ground truth. We then plot a curve over $D$, and aggregate the result
as the average AVP (\aavp). This measure has the advantage that it
properly quantifies angular errors rather than equalizing all
misclassified detections, and it alleviates the somewhat arbitrary
choice of bin centers.

Fig.~\ref{exp:vp}~(center) gives the results according to this
measure, averaged over all 11 classes of PASCAL3D+ (per-class results
are part of the supplemental material). 
Fig.~\ref{exp:vp}~(right) gives the
corresponding results in the original AVP measure for discrete azimuth
angle binnings~\cite{xiang14wacv} as a reference.
In both cases, we compare the performance of our different
RCNN-viewpoint regressor combinations, \rcnnRidge, \rcnnLasso, and
\rcnnElNet, to the discrete multi-view \rcnnmv and \cnnmv, and the
state-of-the-art methods \vdpm and \dpmvocvp.

\myparagraph{Results.}  We observe that in the mAAVP measure
(Fig.~\ref{exp:vp}~(left)), the \rcnn-viewpoint regressor combinations
outperform the previous state-of-the-art methods \vdpm and \dpmvocvp
by large margins. The best performing combination \rcnnRidge
($35.3\%$, brown) outperforms the best \vdpm-16V ($20.9\%$) by
$14.4\%$ and the best \dpmvocvp-16V ($23.0\%$) by $12.3\%$,
respectively. 

The performance of \vdpm and \dpmvocvp is stable or increasing for
increasing numbers of components: \vdpm-4V ($15.5\%$) improves to
\vdpm-16V ($20.9\%$), and \dpmvocvp-4 ($20.1\%$) improves to
\dpmvocvp-16V ($23.0\%$). In contrast, performance decreases for
\rcnnmv and \cnnmv: \rcnnmv-4V ($26.8\%$) decreases to \rcnnmv-24V
($23.0\%$), and \cnnmv-4V ($22.1\%$) decreases to \cnnmv-24V
($16.2\%$).  Even though the best performing \rcnnmv-4V ($26.8\%$)
outperforms the previous state-of-the-art \dpmvocvp-16V ($23.0\%$), it
can not compete with the \rcnn-viewpoint regressor combinations.

The same tendencies are also reflected in the original mAVP
measure~\cite{xiang14wacv} (Fig.~\ref{exp:vp}~(right)). While
\dpmvocvp has a slight edge for the fine binnings (it outperforms
\rcnnRidge by $0.9\%$ for VP$_{16}$ and $1.9\%$ for VP$_{24}$),
\rcnn-viewpoint regressor combinations dominate for the coarser
binnings VP$_4$ and VP$_8$, followed by \rcnnmv, \cnnmv, \vdpm, and
\dpmvocvp.
%
%

\myparagraph{Conclusion.} The combination of \rcnn and viewpoint
regressor \rcnnRidge provides a pronounced improvement in simultaneous
2D BB localization and viewpoint estimation compared to previous
state-of-the-art ($12.3\%$ in mAAVP). Notably, it retains the original
performance in 2D BB localization of the \rcnn ($51.2\%$ in AP).
%

\subsection{2D Keypoint detection}\label{sec:expKeypoint}
%
We proceed by evaluating the basis for our 3D lifting stage, 2D
keypoint detection (Sect.~\ref{sec:spatial}), in isolation.
We use the keypoint annotations provided as part of
Pascal3D+~\cite{xiang14wacv}, and train an \rcnn keypoint detector for
each of 117 types of keypoints distributed over 11 object
categories. Since the keypoints are only characterized by their
location (not extent), we evaluate localization performance in a way
that is inspired by human body pose estimation~\cite{yang13pami}. For computing
a precision-recall curve, we replace the standard BB IoU criterion for
detection with an allowed distance $P$ from the keypoint annotation,
normalized to a reference object height $H$. We refer to this measure
as Average Pixel Precision (APP). In all experiments, we use $H=100$
and $P=25$.

\begin{figure}
\begin{minipage}{\columnwidth}
  \centering
  \raisebox{-0.5\height}{\includegraphics[width=0.345\columnwidth]{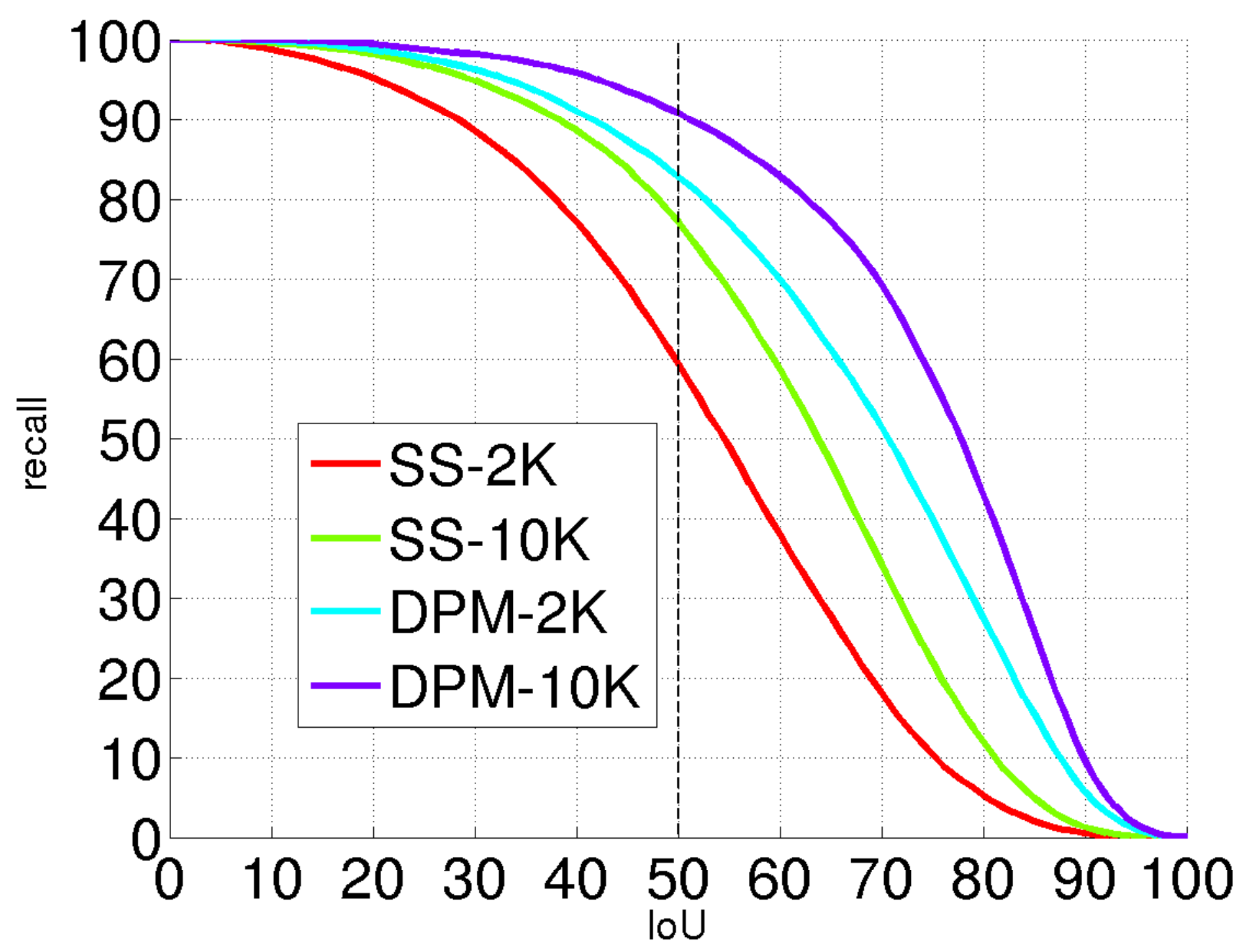}}
  \raisebox{-0.5\height}{\includegraphics[width=0.645\columnwidth]{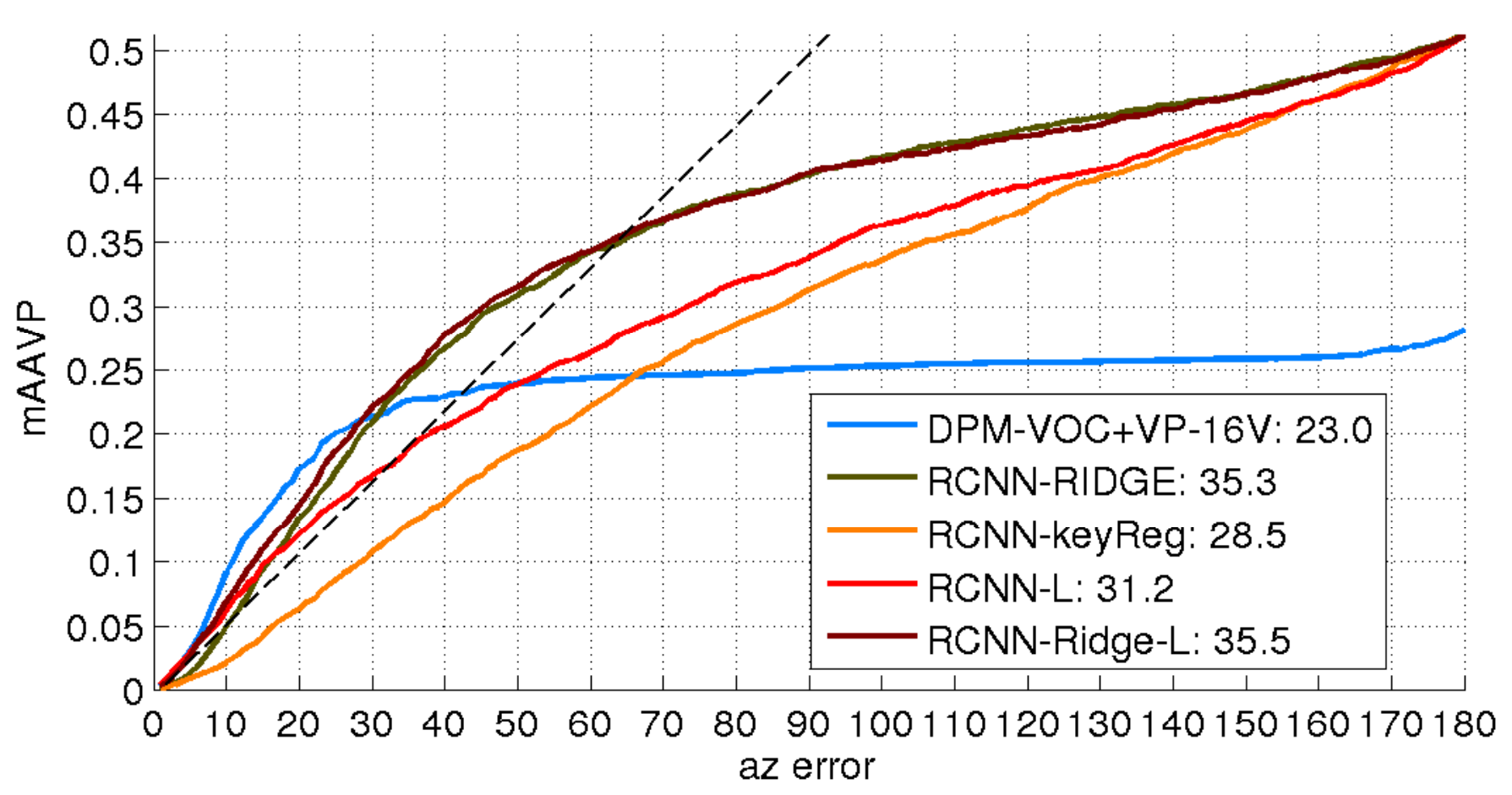}}
\end{minipage}
\caption{Left: 2D Keypoint region proposal quality. Right:
  Simultaneous 2D BB and viewpoint estimation with 3D
  lifting.}
\label{exp:liftPlot}
\end{figure}

\myparagraph{Region proposals.}  We first evaluate the keypoint region
proposal method (Fig.~\ref{exp:liftPlot}~(left)), comparing selective
search (SS) with the deformable part model
(DPM~\cite{felzenszwalb10pami}) at $K=2000$ and $K=10000$ top-scoring
regions per image.  The DPM is trained independently for each keypoint
(for that purpose, we define the BB of each keypoint to be a square
centered at the keypoint with area equal to $30\%$ of the object
area).
%
%
Both DPM versions outperform the corresponding SS methods by large
margin: at 70\% IoU DPM with $K=10000$ gives $30\%$ more recall than SS-$10K$
 which is why we stick with these keypoint proposals for our 3D
object class detection pipeline.

\myparagraph{Part localization.}
%
%
%
%
%
Tab.~\ref{tab:partdet} compares the performance of our \rcnn keypoint
detectors with the DPM keypoint proposal detectors alone, in APP.
On average, the \rcnn-FT keypoint detectors trained using the features from the
CNN fine-tuned on keypoint detection ($27.3\%$) outperform the 
DPM ($14.4\%$) by $12.9\%$ APP 
providing a solid basis for our 3D lifting procedure.
%

\subsection{2D to 3D lifting}
\label{sec:expLifting}
\begin{table}[t]
\scriptsize
\centering
\setlength{\tabcolsep}{1.5pt}
\begin{tabular}{cp{0.5cm}p{0.5cm}p{0.5cm}p{0.5cm}p{0.5cm}p{0.5cm}p{0.5cm}p{0.5cm}p{0.5cm}p{0.5cm}p{0.5cm}p{0.5cm}p{0.5cm}}
   \tiny{APP}                 & aero plane       & bike          & boat          & bus           & car           & chair        & din. table      & mot. bike        & sofa          & train         & tv            & AVG           \\
\toprule
\tiny{DPM}                    & 19.2          & 36.2          & 8.9           & 26.4          & 14.3          & 3.1          & 4.0          & 24.2          & 7.6           & 8.5           & 6.1           & 14.4          \\
\tiny{RCNN}                   & 24.6          & 43.1          & 9.8           & 47.8          & 34.1          & 5.7          & 4.6          & 36.7          & \textbf{14.3} & 22.5          & 21.5          & 24.1          \\
\tiny{RCNN FT}                & \textbf{30.4} & \textbf{48.9} & \textbf{12.4} & \textbf{50.8} & \textbf{39.5} & \textbf{9.5} & \textbf{6.3} & \textbf{41.6} & 14.0          & \textbf{24.5} & \textbf{22.8} & \textbf{27.3} \\
\bottomrule
\end{tabular}
\caption{Part detection performance in APP.}
\label{tab:partdet}
\end{table}
Finally, we evaluate the performance of our full 3D object class
detection pipeline that predicts the precise 3D shape and pose.  We
first give results on simultaneous 2D BB localization and viewpoint
estimation as before, but then move on to measuring the quality of our
predicted 3D shape estimates, in the form of a segmentation task. We
generate segmentation masks by simply projecting the predicted 3D
shape (Fig.~\ref{fig:teaser} (right)).
We compare the performance of a direct 3D lifting (\rcnnl) of detected
2D keypoints with a viewpoint guided 3D lifting (\rcnnRidgeL), and a
baseline that regresses keypoint positions (\keyReg) on top of an
\rcnn object detector rather than using keypoint detections.

\myparagraph{Simultaneous 2D BB \& VP estimation.}
Fig.~\ref{exp:liftPlot}~(right) compares the mAAVP performance of the
lifting methods with the best viewpoint regressor \rcnnRidge and the
best previously published method
\dpmvocvp-16V. Fig.~\ref{exp:vp}~(right) gives the
AVP$_V$~\cite{xiang14wacv} performance in comparison with all
viewpoint classifiers and regressors.

\rcnnl ($31.2\%$ mAAVP) and \rcnnRidgeL ($35.5\%$) outperform both the
\keyReg ($28.5\%$) and the \dpmvocvp-16V ($23.0\%$) by considerable
margins. \rcnnRidgeL consistently outperforms \rcnnRidge in terms of
AVP$_V$ (by $1.6\%$, $2.2\%$, $2.2\%$, and $4.1\%$ for increasing
$V$), thus improving over the previous pipeline stage. Furthermore,
with $18.6\%$ AVP$_{16}$ and $15.8\%$ AVP$_{24}$ it also outperforms
\dpmvocvp-16V ($17.3\%$,$13.6\%$, respectively),
%
and achieving state-of-the-art simultaneous BB localization and
viewpoint estimation results on Pascal3D+.

\myparagraph{Segmentation.} Tab.~\ref{exp:segm1} reports the
segmentation accuracy on Pascal3D+. We use the evaluation protocol
of~\cite{xiang14wacv} with two differences. First, we evaluate inside
the ground truth BB only to account for truncated and occluded
objects. Second, we focus the evaluation on objects with actual ground
truth 3D prototype alignment as that constitutes the relevant set of
objects we want to compare on. Therefore, we report the performance of
the ground truth aligned 3D CAD prototypes (GT) as well.

With 41.4\% performance across all classes, \rcnnRidgeL outperforms
\rcnnl (36.9\%) and the baseline \keyReg (36.4\%) by 4\%, confirming
the quality of the alignment. Fig.~\ref{fig:ex} illustrates successful
2D-3D alignments for different object classes, along with failure
cases. Truncated and occluded objects represent a major part of the
failures.

In Tab.~\ref{tab:segm2} we go one step further and compare to native
state-of-the-art segmentation methods (O$_2$P~\cite{carreira12eccv}),
this time on the Pascal-context~\cite{mottaghi14cvpr} dataset. We
report the performance on the 11 classes from Pascal3D+
only. \rcnnRidgeL with 31.5\% is only slightly worse than O$_2$P+
(35.9\%) although the latter is designed for segmentation.

%
%

%
%
\myparagraph{Conclusion.}  We conclude that \rcnnRidgeL achieves
state-of-the-art simultaneous BB localization and viewpoint estimation
performance on Pascal3D+~\cite{xiang14wacv}, outperforming the
\dpmvocvp and the \rcnnRidge regressor. It successfully predicts the
3D object shape which is confirmed by it's segmentation performance.

\begin{table}
\centering
\scriptsize
\setlength{\tabcolsep}{1pt}
\begin{tabular}{cp{0.5cm}p{0.5cm}p{0.5cm}p{0.5cm}p{0.5cm}p{0.5cm}p{0.5cm}p{0.5cm}p{0.5cm}p{0.5cm}p{0.5cm}p{0.5cm}p{0.5cm}}
sAcc               & aero plane       & bike          & boat          & bus           & car           & chair         & din. table       & mot. bike        & sofa          & train         & tv            & AVG           \\
\toprule
 GT                & 58.3          & 32.0          & 57.9          & 84.9          & 79.6          & 53.5          & 63.1          & 69.3          & 64.7          & 70.5          & 80.7          & 65.0          \\
 \tiny{\keyReg}    & 27.1          & 20.2          & 19.1          & 56.2          & 47.7          & 23.0          & 18.6          & 41.3          & \textbf{46.4} & 30.9          & \textbf{70.0} & 36.4          \\
\tiny{\rcnnl}      & 30.3          & 22.0          & \textbf{27.9} & 60.5          & 44.2          & 24.9          & 24.4          & 46.3          & 41.9          & 37.5          & 45.6          & 36.9          \\
\tiny{\rcnnRidgeL} & \textbf{35.1} & \textbf{22.2} & 26.9          & \textbf{66.4} & \textbf{53.9} & \textbf{26.8} & \textbf{28.6} & \textbf{49.0} & 44.8          & \textbf{42.5} & 58.7          & \textbf{41.4} \\
\bottomrule
\end{tabular}
\caption{Segmentation accuracy on Pascal3D+.}
\label{exp:segm1}
\end{table}

\begin{table}
\centering
\scriptsize
\setlength{\tabcolsep}{1pt}
\begin{tabular}{cp{0.5cm}p{0.5cm}p{0.5cm}p{0.5cm}p{0.5cm}p{0.5cm}p{0.5cm}p{0.5cm}p{0.5cm}p{0.5cm}p{0.5cm}p{0.5cm}p{0.5cm}}
 sAcc                                & aero plane       & bike          & boat          & bus           & car           & chair         & din. table       & mot. bike        & sofa          & train        & tv            & AVG  \\
\toprule
GT                                   & 40.3          & 27.9          & 36.2          & 75.0          & 59.3          & 34.9          & 16.0          & 59.0          & 25.2          & 57.0         & 72.5          & 45.7 \\
\tiny{O$_2$P~\cite{carreira12eccv}}  & 48.2          & 32.5          & 29.6          & \textbf{61.1} & 46.7          & 12.4          & 12.4          & \textbf{46.0} & 17.0          & 36.7         & 41.6          & 34.9 \\ 
\tiny{$O_2$P+~\cite{mottaghi14cvpr}} & \textbf{52.4} & \textbf{32.8} & \textbf{33.1} & 60.5          & \textbf{47.8} & 12.8          & \textbf{13.0} & 44.5          & 16.7          & \textbf{40.1} & 40.7          & \textbf{35.9} \\
\tiny{\keyReg}                       & 21.9          & 17.2          & 15.1          & 49.5          & 39.2          & 16.4          & 11.8          & 37.3          & \textbf{21.9} & 28.2         & \textbf{60.9} & 29.0 \\
\tiny{\rcnnl}                        & 26.7          & 18.8          & 17.5          & 53.9          & 36.7          & 16.2          & 6.4           & 43.5          & 16.3          & 35.5         & 49.7          & 29.2 \\
\tiny{\rcnnRidgeL}                   & 27.7          & 20.1          & 19.9          & 59.0          & 41.7          & \textbf{18.2} & 7.8           & 44.4          & 18.5          & 37.9         & 51.1          & 31.5 \\
\bottomrule
\end{tabular}
\caption{Segmentation accuracy on Pascal-context~\cite{mottaghi14cvpr} dataset.}
\label{tab:segm2}
\end{table}

\section{Conclusions}

In this work we have build a 3D object class detector, capable of
detecting objects of multiple object categories in the wild
(Pascal3D+).
It consists of four main stages: {\em (i)} object detection, {\em
  (ii)} viewpoint estimation, {\em (iii)} keypoint detection and {\em
  (iv)} 2D-3D lifting. Based on careful design choices, 
our 3D object class detector improves the performance in each stage,
achieving state-of-the-art object BB localization and simultaneous BB
localization and viewpoint estimation performance on the challenging
Pascal3D+ dataset. At the same time, it predicts the 3D shape of the
objects, as confirmed by it's segmentation quality. The final result
is a rich 3D representation, consisting of 3D shape, 3D viewpoint, and
3D position automatically estimated using only 2D image evidence.
%
%
\clearpage
\clearpage
{\small
\bibliographystyle{ieee}
\bibliography{paper}
}
\clearpage
\clearpage
\section{Supplementary material}

This supplementary material provides more detailed, per-class object
detection (Sect.~\ref{sec:supplDet}), simultaneous localization and
viewpoint estimation (Sect.~\ref{sec:supplVp}) and 2D-3D CAD prototype
alignment results (Sect.~\ref{sec:suppl3Dlift}). We also provide
additional qualitative results.

\section{2D Bounding box localization}\label{sec:supplDet}
Tab.~\ref{tab:supplDet} reports the per-class 2D BB localization
results in terms of average precision (AP), in addition to the average
results presented in Fig.~\ref{exp:vp} (left) in the paper. We compare
the first stage of our pipeline (\rcnn~\cite{girshick14cvpr}) with
\cnn and DPM~\cite{felzenszwalb10pami}, with the multi-view variants
\rcnnmv and \cnnmv and the state-of-the-art \vdpm~\cite{xiang14wacv}
and \dpmvocvp~\cite{pepik12cvpr}.

As observed previously (Sect.~\ref{sec:expObjDet} in the paper) \rcnn
with $51.2\%$ outperforms the \cnn ($47.2\%$) by $4.0\%$, which in
turn outperforms the DPM ($29.6\%$). In addition, in
Tab.~\ref{tab:supplDet} we observe that this trend is consistent
across all the classes: \rcnn is better than \cnn on every class,
while both methods are superior in comparison to the DPM and
outperform it by large margin on each class ($\sim 20\%$) . As shown
before (Fig.~\ref{exp:vp} (left) in the paper), introducing viewpoint
information in terms of viewpoint bins causes \rcnn and \cnn to
drastically drop in performance.  \rcnn-24V (30.8\%) and \cnn-24V
($27.6\%$) are by $20.4\%$ and $20.0\%$ worse compared to their
respective variants that do not include viewpoint
information. Tab.~\ref{tab:supplDet} confirms that this trend is
consistent across all the Pascal3D+ classes. However, the \rcnnmv
variants on average are still better than the state-of-the-art \vdpm
multi-view variants by $12.4\%$, $3.3\%$, $2.0\%$ and $1.3\%$ on
VP$_4$, VP$_8$, VP$_{16}$, VP$_{24}$ respectively. This is consistent
across most of the object classes, with {\em car} and {\em bicycle}
being the only notable exception.

\begin{table*}
\centering
\small
\setlength{\tabcolsep}{3pt}
\begin{tabular}{cCCCCCCCCCCCC}
  AP          & aeroplane     & bicycle       & boat          & bus           & car           & chair         & diningtable   & motorbike     & sofa          & train         & tvmonitor     & AVG           \\ 
\toprule
DPM 	      & 42.2          & 49.6          & 6.0           & 54.1          & 38.3          & 15.0          & 9.0           & 33.1          & 18.9          & 36.4          & 33.2          & 29.6          \\
\vdpm-4V      & 40.0          & 45.2          & 3.0           & 49.3          & 37.2          & 11.1          & 7.2           & 33.0          & 6.8           & 26.4          & 35.9          & 26.8          \\
\vdpm-8V      & 39.8          & 47.3          & 5.8           & 50.2          & 37.3          & 11.4          & 10.2          & 36.6          & 16.0          & 28.7          & 36.3          & 29.9          \\
\vdpm-16V     & 43.6          & 46.5          & 6.2           & 54.6          & 36.6          & 12.8          & 7.6           & 38.5          & 16.2          & 31.5          & 35.6          & 30.0          \\
\vdpm-24V     & 42.2          & 44.4          & 6.0           & 53.7          & 36.3          & 12.6          & 11.1          & 35.5          & 17.0          & 32.6          & 33.6          & 29.5          \\
\hline                                                                       
\dpmvocvp-4V  & 41.5          & 46.9          & 0.5           & 51.5          & 45.6          & 8.7           & 5.7           & 34.3          & 13.3          & 16.4          & 32.4          & 27.0          \\
\dpmvocvp-8V  & 40.5          & 48.1          & 0.5           & 51.9          & 47.6          & 11.3          & 5.3           & 38.3          & 13.5          & 21.3          & 33.1          & 28.3          \\
\dpmvocvp-16V & 38.0          & 45.6          & 0.7           & 55.3          & 46.0          & 10.2          & 6.2           & 38.1          & 11.8          & 28.5          & 30.7          & 28.3          \\
\dpmvocvp-24V & 36.0          & 45.9          & 5.3           & 53.9          & 42.1          & 8.0           & 5.4           & 34.8          & 11.0          & 28.2          & 27.3          & 27.1          \\
\hline                                                                                    
RCNN          & \textbf{68.8} & \textbf{63.5} & \textbf{29.0} & \textbf{64.3} & \textbf{55.5} & \textbf{26.1} & \textbf{33.2} & \textbf{68.9} & \textbf{39.1} & \textbf{54.0} & \textbf{60.5} & \textbf{51.2} \\
\rcnnmv-4V    & 55.0          & 50.1          & 18.9          & 55.9          & 38.1          & 13.9          & 20.7          & 51.2          & 29.4          & 41.4          & 56.1          & 39.2          \\ 
\rcnnmv-8V    & 53.4          & 38.6          & 15.3          & 44.5          & 33.7          & 13.5          & 21.5          & 45.5          & 21.5          & 34.5          & 43.7          & 33.2          \\ 
\rcnnmv-16V   & 46.7          & 39.2          & 13.6          & 51.2          & 31.0          & 12.9          & 18.5          & 43.9          & 19.9          & 35.1          & 40.3          & 32.0          \\ 
\rcnnmv-24V   & 48.5          & 39.2          & 13.0          & 44.8          & 30.6          & 11.6          & 18.1          & 45.7          & 18.1          & 35.6          & 33.9          & 30.8          \\ 
\hline                                                                                          
CNN           & 63.7          & 60.1          & 24.7          & 62.6          & 52.6          & 23.4          & 26.1          & 65.4          & 33.2          & 52.3          & 59.1          & 47.6          \\
\cnnmv-4V     & 44.5          & 40.2          & 16.0          & 51.2          & 40.4          & 12.6          & 19.5          & 45.5          & 25.4          & 38.8          & 57.7          & 35.6          \\
\cnnmv-8V     & 38.8          & 42.2          & 16.3          & 47.2          & 35.9          & 11.5          & 20.4          & 44.6          & 24.0          & 34.2          & 50.5          & 33.2          \\
\cnnmv-16V    & 37.6          & 31.3          & 14.4          & 43.4          & 32.7          & 10.6          & 16.5          & 39.4          & 19.8          & 30.9          & 45.4          & 29.3          \\
\cnnmv-24V    & 37.6          & 31.0          & 14.7          & 36.6          & 31.4          & 9.0           & 16.6          & 36.8          & 20.2          & 32.2          & 37.3          & 27.6          \\
\bottomrule 
\end{tabular}
\caption{Per class object 2D bounding box localization
  results (Fig.~\ref{exp:vp} (left) in paper). Pascal3D+~\cite{xiang14wacv} dataset. }
\label{tab:supplDet}
\end{table*}

\section{Simultaneous 2D BB and viewpoint estimation}\label{sec:supplVp}

Tab.~\ref{tab:supVp} reports the simultaneous BB localization and
viewpoint estimation performance on all classes of Pascal3D+, in
addition to the average results presented in Fig.~\ref{exp:vp} (right)
in the paper. Now, we also compare the performance of the second stage
of our pipeline, the \rcnn viewpoint regressors: \rcnnRidge,
\rcnnLasso and \rcnnElNet. We use the average viewpoint precision
(AVP) measure at VP$_V$ viewpoint bins ($V \in \{4,8,16,24\}$).

As seen previously in Fig.~\ref{exp:vp} (right), on average our \rcnn
viewpoint regressors achieve state-of-the-art performance on the
coarse bins: \rcnnElNet with $37.1\%$ AVP$_4$ outperforms the
state-of-the-art \dpmvocvp-4V (23.8\%) and \rcnnRidge ($25.4\%$)
AVP$_8$ is again better than \dpmvocvp-8V ($21.5\%$).  However, at the
finer viewpoint bins \dpmvocvp ($17.3\%$ AVP$_{16}$, $13.6\%$
AVP$_{24}$) excels over our \rcnnRidge ($16.4\%$,
$11.7\%$). Additionally, Tab.~\ref{tab:supVp} suggests that the
tendencies are similar for the different classes as well. Compared to
\vdpm and \dpmvocvp, the \rcnn viewpoint regressors dominate or are at
least comparable on most of the classes, except on {\em car}, {\em
  bus}  and {\em bicycle} where \dpmvocvp is better.

\begin{table*}
\centering
\small
\setlength{\tabcolsep}{3pt}
\begin{tabular}{cCCCCCCCCCCCC}
  AVP          & aeroplane     & bicycle       & boat          & bus           & car           & chair         & diningtable   & motorbike     & sofa          & train         & tvmonitor     & AVG           \\ 
\toprule
\vdpm-4V       & 34.6          & 41.7          & 1.5           & 26.1          & 20.2          & 6.8           & 3.1           & 30.4          & 5.1           & 10.7          & 34.7          & 19.5          \\
\vdpm-8V       & 23.4          & 36.5          & 1.0           & 35.5          & 23.5          & 5.8           & 3.6           & 25.1          & 12.5          & 10.9          & 27.4          & 18.7          \\
\vdpm-16V      & 15.4          & 18.4          & 0.5           & 46.9          & 18.1          & 6.0           & 2.2           & 16.1          & 10.0          & 22.1          & 16.3          & 15.6          \\
\vdpm-24V      & 8.0           & 14.3          & 0.3           & 39.2          & 13.7          & 4.4           & 3.6           & 10.1          & 8.2           & 20.0          & 11.2          & 12.1          \\
\hline                                                                                         
\dpmvocvp-4V   & 37.4          & 43.9          & 0.3           & 48.6          & \textbf{37.0} & 6.1           & 2.1           & 31.8          & 11.8          & 11.1          & 32.2          & 23.8          \\
\dpmvocvp-8V   & 28.6          & \textbf{40.3} & 0.2           & \textbf{38.0} & \textbf{36.6} & 9.4           & 2.6           & 32.0          & 11.0          & 9.8           & 28.6          & 21.5          \\
\dpmvocvp-16V  & \textbf{15.9} & \textbf{22.9} & 0.3           & \textbf{49.0} & \textbf{29.6} & 6.1           & 2.3           & 16.7          & 7.1           & 20.2          & 19.9          & \textbf{17.3} \\
\dpmvocvp-24V  & 9.7           & \textbf{16.7} & 2.2           & \textbf{42.1} & \textbf{24.6} & 4.2           & 2.1           & 10.5          & 4.1           & \textbf{20.7} & 12.9          & \textbf{13.6} \\
\hline                                                                                                        
\rcnnmv-4V     & 40.3          & 32.0          & 11.4          & 51.9          & 28.2          & 9.1           & 10.0          & 36.5          & 27.2          & 36.6          & 54.5          & 30.7          \\ 
\rcnnmv-8V     & 29.9          & 22.4          & 5.4           & 33.3          & 20.3          & 6.1           & 11.0          & 24.3          & 14.8          & 28.2          & 32.2          & 20.7          \\ 
\rcnnmv-16V    & 12.4          & 9.6           & 2.8           & 31.9          & 12.9          & 3.3           & \textbf{7.8}  & 15.6          & 9.8           & \textbf{22.9} & 17.8          & 13.3          \\ 
\rcnnmv-24V    & 7.1           & 7.2           & 1.3           & 24.7          & 10.4          & 2.6           & 5.5           & 8.2           & 6.5           & 18.0          & 13.4          & 9.5           \\ 
\hline                                                                                                           
\cnnmv-4V      & 24.4          & 22.5          & 8.4           & 48.8          & 33.4          & 7.8           & 11.0          & 31.8          & 22.1          & 34.6          & 55.9          & 27.3          \\
\cnnmv-8V      & 14.9          & 17.8          & 5.1           & 35.9          & 23.9          & 5.7           & 9.5           & 21.4          & 15.4          & 27.0          & 38.8          & 19.6          \\
\cnnmv-16V     & 8.4           & 8.2           & 2.3           & 23.1          & 15.4          & 2.8           & 5.3           & 9.9           & 9.6           & 16.9          & 21.7          & 11.2          \\
\cnnmv-24V     & 6.0           & 4.3           & 1.9           & 14.7          & 12.0          & 1.4           & \textbf{6.0}  & 6.6           & 6.5           & 14.9          & 15.4          & 8.1           \\
\hline
\rcnnRidge-4V  & 46.3          & 37.3          & \textbf{13.3} & \textbf{55.7} & 32.1          & \textbf{15.8} & \textbf{18.8} & 50.6          & 30.2          & \textbf{45.0} & \textbf{57.1} & 36.6          \\
\rcnnRidge-8V  & \textbf{37.0} & 29.3          & \textbf{7.0}  & 33.0          & 25.0          & \textbf{10.8} & 10.3          & 30.3          & \textbf{21.4} & 32.1          & \textbf{43.4} & \textbf{25.4} \\
\rcnnRidge-16V & 14.3          & 14.7          & \textbf{7.0}  & 34.0          & 13.0          & \textbf{6.8}  & 5.0           & 19.7          & 14.6          & 22.2          & \textbf{28.9} & 16.4          \\
\rcnnRidge-24V & 11.3          & 11.5          & \textbf{4.2}  & 20.7          & 11.4          & \textbf{4.6}  & 3.0           & 11.5          & 10.5          & 19.4          & \textbf{20.6} & 11.7          \\
\hline
\rcnnLasso-4V  & 46.8          & 40.9          & 13.2          & 55.5          & 33.1          & 14.8          & 18.8          & 49.6          & \textbf{31.2} & 43.8          & 56.8          & 36.8          \\
\rcnnLasso-8V  & 35.1          & 25.3          & 7.0           & 33.1          & 21.1          & 9.9           & \textbf{12.1} & \textbf{35.0} & 19.9          & \textbf{34.9} & 38.9          & 24.8          \\
\rcnnLasso-16V & 15.8          & 12.7          & 2.9           & 32.9          & 15.1          & 5.7           & 5.4           & 19.3          & \textbf{16.3} & 17.7          & 25.7          & 15.4          \\
\rcnnLasso-24V & \textbf{11.9} & 10.1          & 2.0           & 18.9          & 11.2          & 3.3           & 3.6           & 12.5          & \textbf{10.6} & 19.8          & 16.2          & 10.9          \\
\hline
\rcnnElNet-4V  & \textbf{46.9} & \textbf{41.7} & 12.9          & 55.6          & 33.2          & 14.7          & 18.8          & \textbf{52.2} & 31.2          & 43.8          & 56.8          & \textbf{37.1} \\
\rcnnElNet-8V  & 34.6          & 27.5          & 6.6           & 32.8          & 22.3          & 9.9           & 10.3          & 33.3          & 20.2          & 33.8          & 39.8          & 24.6          \\
\rcnnElNet-16V & 15.5          & 10.7          & 3.1           & 36.6          & 12.9          & 5.7           & 5.6           & \textbf{22.8} & 15.3          & 18.0          & 27.1          & 15.8          \\
\rcnnElNet-24V & 11.5          & 7.4           & 2.5           & 21.4          & 11.2          & 3.8           & 3.5           & \textbf{13.3} & 10.3          & 20.4          & 17.7          & 11.2          \\
\bottomrule
\end{tabular}
\caption{Per class simultaneous BB and viewpoint estimation. Multi-view and continuous viewpoint methods (Fig.~\ref{exp:vp} (right) in paper).}
\label{tab:supVp}
\end{table*}

\section{2D to 3D lifting}\label{sec:suppl3Dlift}

\begin{table*}
\centering
\small
\setlength{\tabcolsep}{3pt}
\begin{tabular}{cCCCCCCCCCCCC}
  AVP             & aeroplane     & bicycle       & boat          & bus           & car           & chair         & diningtable   & motorbike     & sofa          & train         & tvmonitor     & AVG           \\ 
\toprule
 \keyReg 4V       & 38.1          & 25.1          & 11.7          & 48.6          & 24.5          & 10.3          & 18.0          & 36.8          & 24.5          & 32.0          & 36.5          & 27.8          \\
 \keyReg 8V       & 23.2          & 13.9          & 8.5           & 28.1          & 13.8          & 5.1           & 12.3          & 20.9          & 14.3          & 27.4          & 24.7          & 17.5          \\
 \keyReg 16V      & 13.8          & 8.1           & 3.5           & 24.8          & 8.9           & 3.2           & 9.0           & 12.5          & 8.5           & 14.6          & 15.6          & 11.1          \\
 \keyReg 24V      & 7.3           & 5.6           & 2.4           & 17.0          & 6.6           & 2.0           & 7.6           & 9.7           & 5.4           & 13.7          & 12.6          & 8.2           \\
\hline
\rcnnl-4V         & 49.0          & \textbf{52.7} & 10.7          & 55.6          & \textbf{40.5} & 8.9           & \textbf{19.9} & 47.8          & 23.8          & 42.9          & 42.5          & 35.8          \\
\rcnnl-8V         & \textbf{36.0} & 36.8          & 8.0           & 35.8          & \textbf{30.7} & 6.1           & \textbf{14.3} & 26.7          & 17.0          & 32.7          & 26.5          & 24.6          \\
\rcnnl-16V        & \textbf{24.1} & 19.6          & 3.1           & \textbf{43.5} & \textbf{24.1} & 3.7           & \textbf{10.7} & 12.8          & 12.7          & \textbf{27.5} & 12.5          & 17.7          \\
\rcnnl-24V        & \textbf{16.7} & 13.7          & 3.8           & 29.0          & \textbf{19.7} & 2.6           & \textbf{8.5}  & 10.2          & 10.7          & 21.9          & 10.4          & 13.4          \\
\hline
\rcnnRidgeL-4V    & \textbf{52.0} & 48.2          & \textbf{13.2} & \textbf{56.1} & 36.2          & \textbf{15.8} & 19.5          & \textbf{51.4} & \textbf{28.2} & \textbf{44.7} & \textbf{54.9} & \textbf{38.2} \\
\rcnnRidgeL-8V    & 35.7          & \textbf{38.4} & \textbf{9.2}  & \textbf{40.9} & 30.1          & \textbf{11.1} & 12.5          & \textbf{39.3} & \textbf{21.3} & \textbf{36.5} & \textbf{28.3} & \textbf{27.6} \\
\rcnnRidgeL-16V   & 20.3          & \textbf{22.1} & \textbf{4.9}  & 39.0          & 22.8          & \textbf{6.9}  & 5.4           & \textbf{20.1} & \textbf{17.0} & 27.0          & \textbf{19.3} & \textbf{18.6} \\
 \rcnnRidgeL-24V  & 16.0          & \textbf{18.7} & \textbf{4.7}  & \textbf{35.7} & 17.9          & \textbf{5.1}  & 3.7           & \textbf{19.6} & \textbf{12.2} & \textbf{25.7} & \textbf{14.9} & \textbf{15.8} \\
\bottomrule
\end{tabular}
\caption{Per class simultaneous 2D BB localization and viewpoint estimation. 3D object detection methods (Fig.~\ref{exp:vp} (right) in paper). 
}
\label{tab:suppl3Dlift}
\end{table*}

\begin{figure*}
\centering
\includegraphics[width=0.245\textwidth]{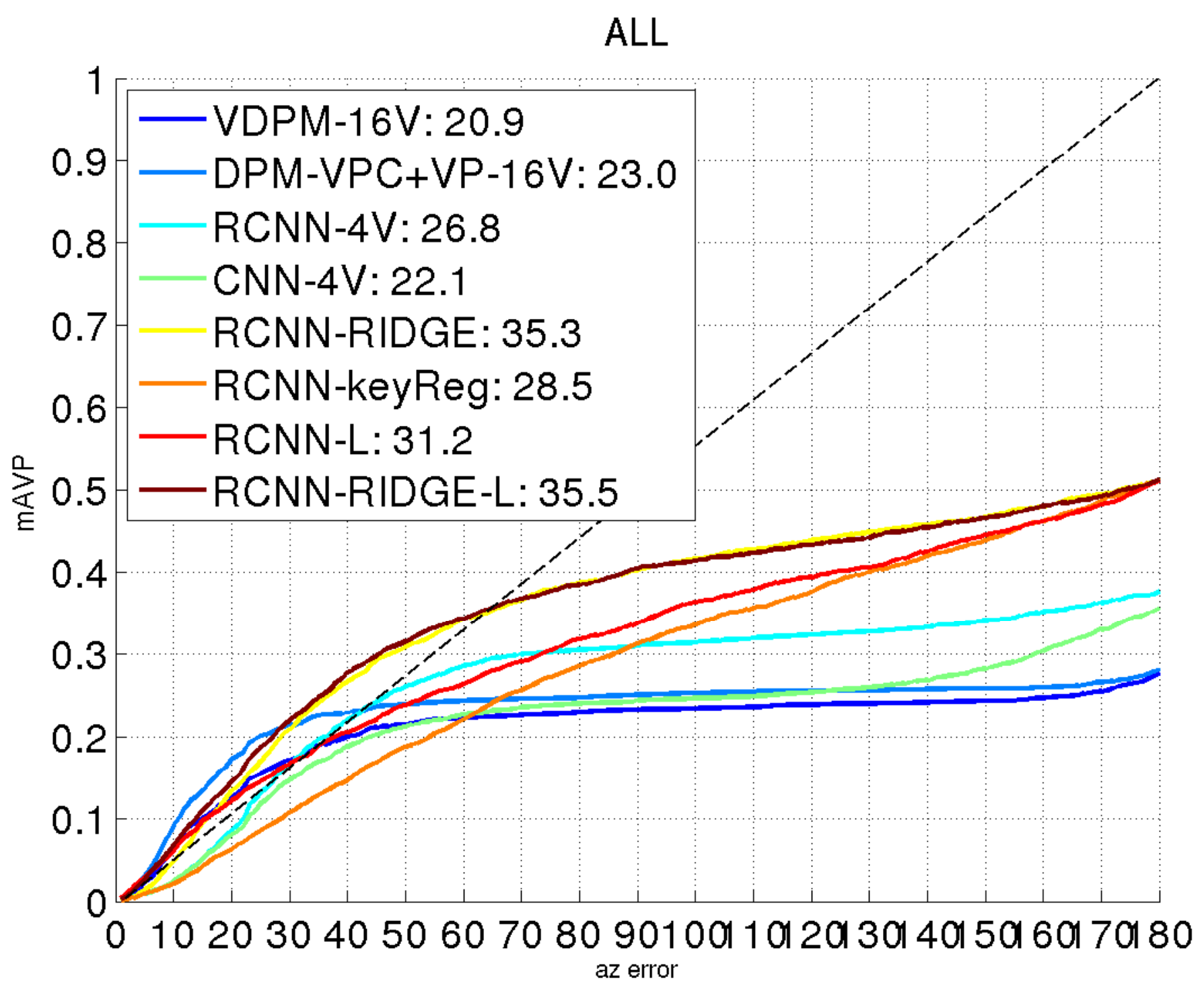}
\includegraphics[width=0.245\textwidth]{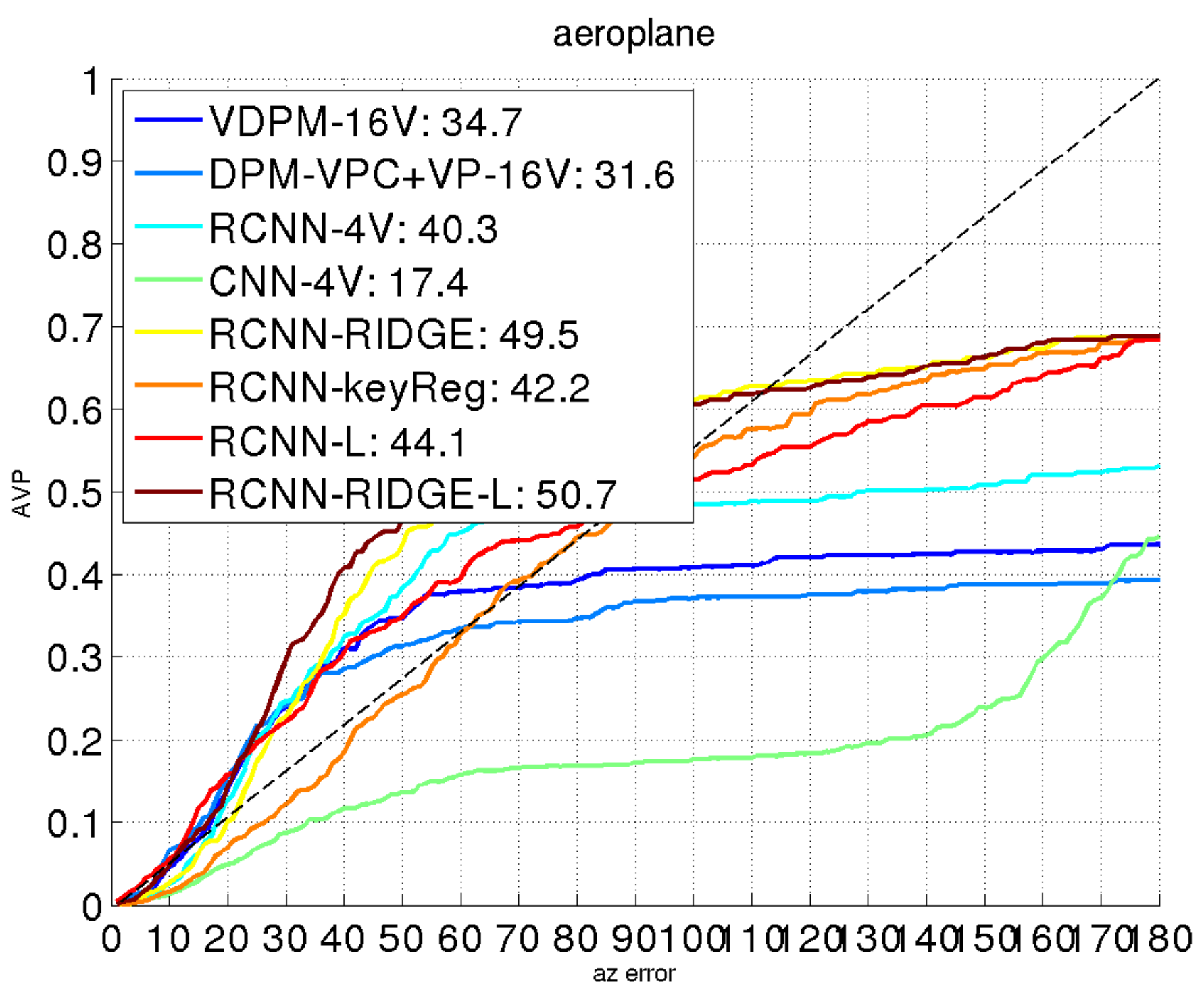}
\includegraphics[width=0.245\textwidth]{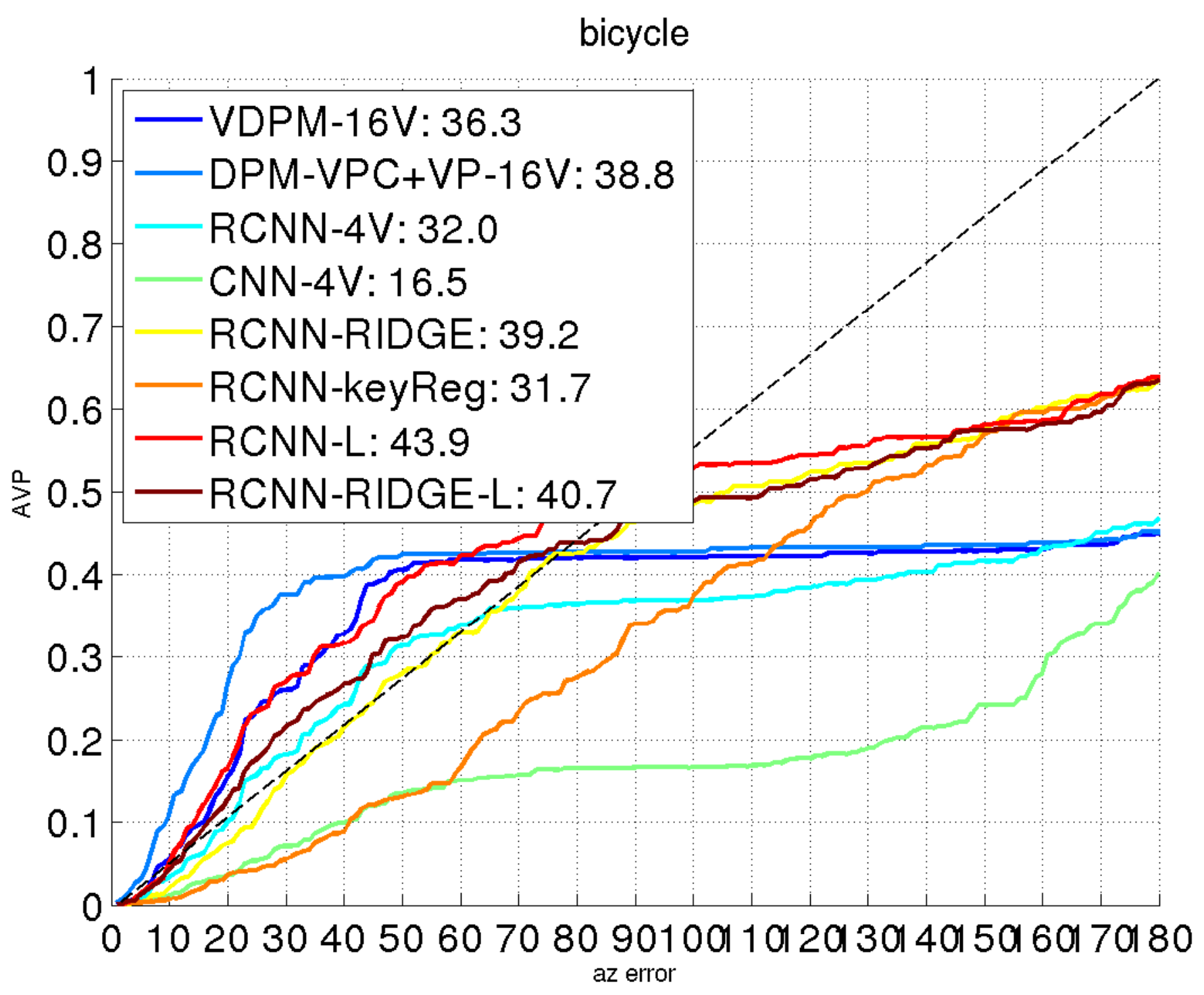}
\includegraphics[width=0.245\textwidth]{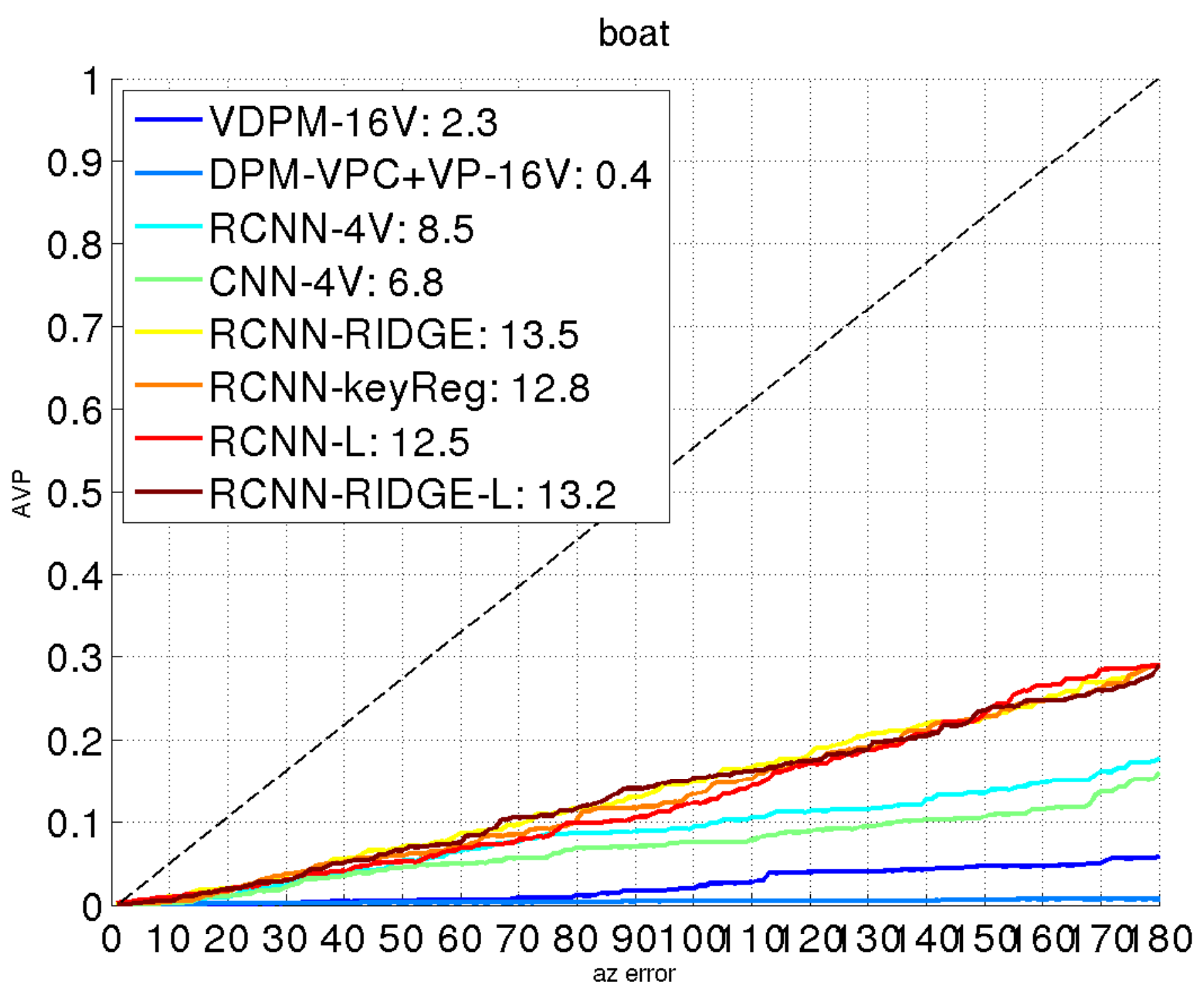}
\includegraphics[width=0.245\textwidth]{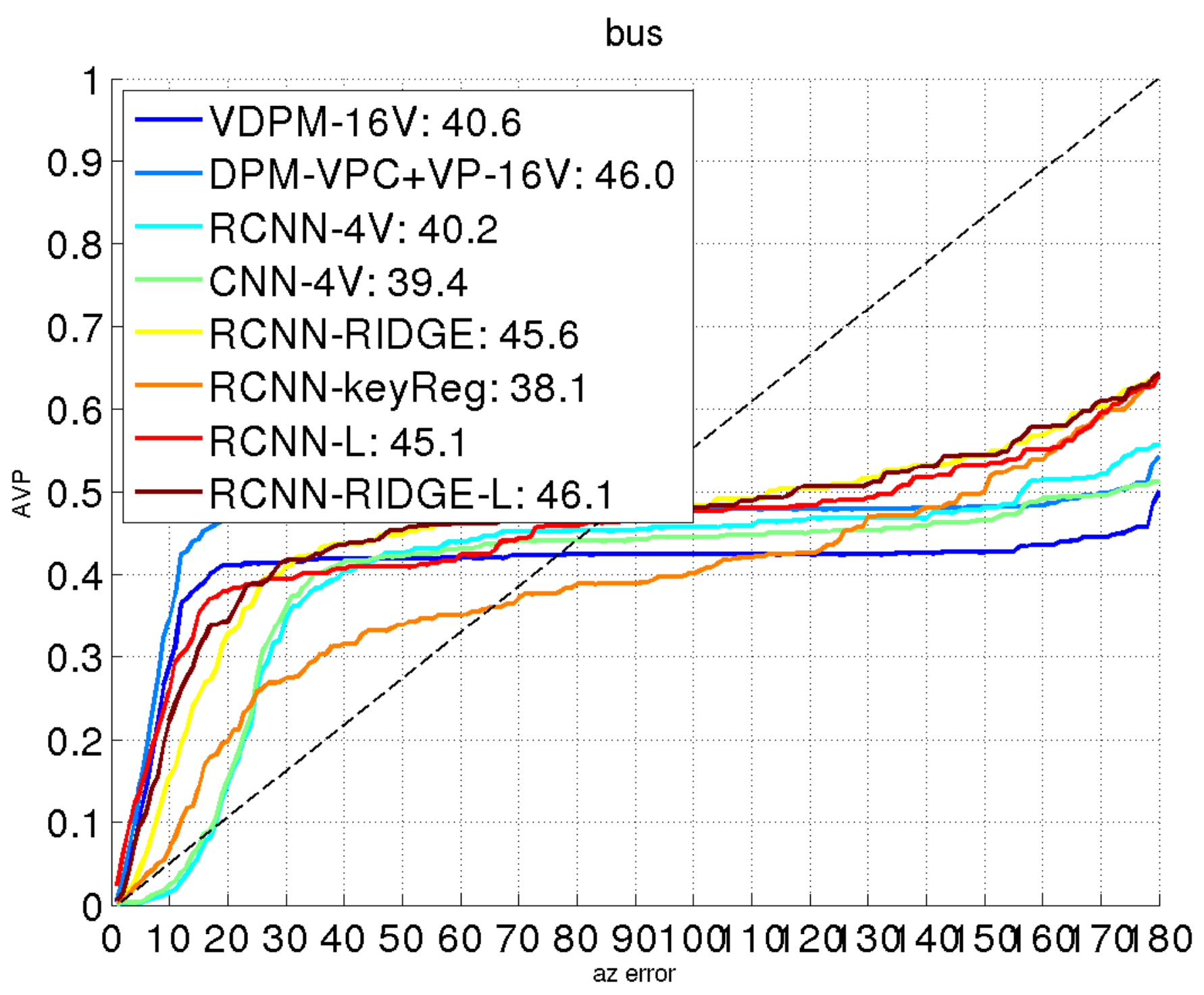}
\includegraphics[width=0.245\textwidth]{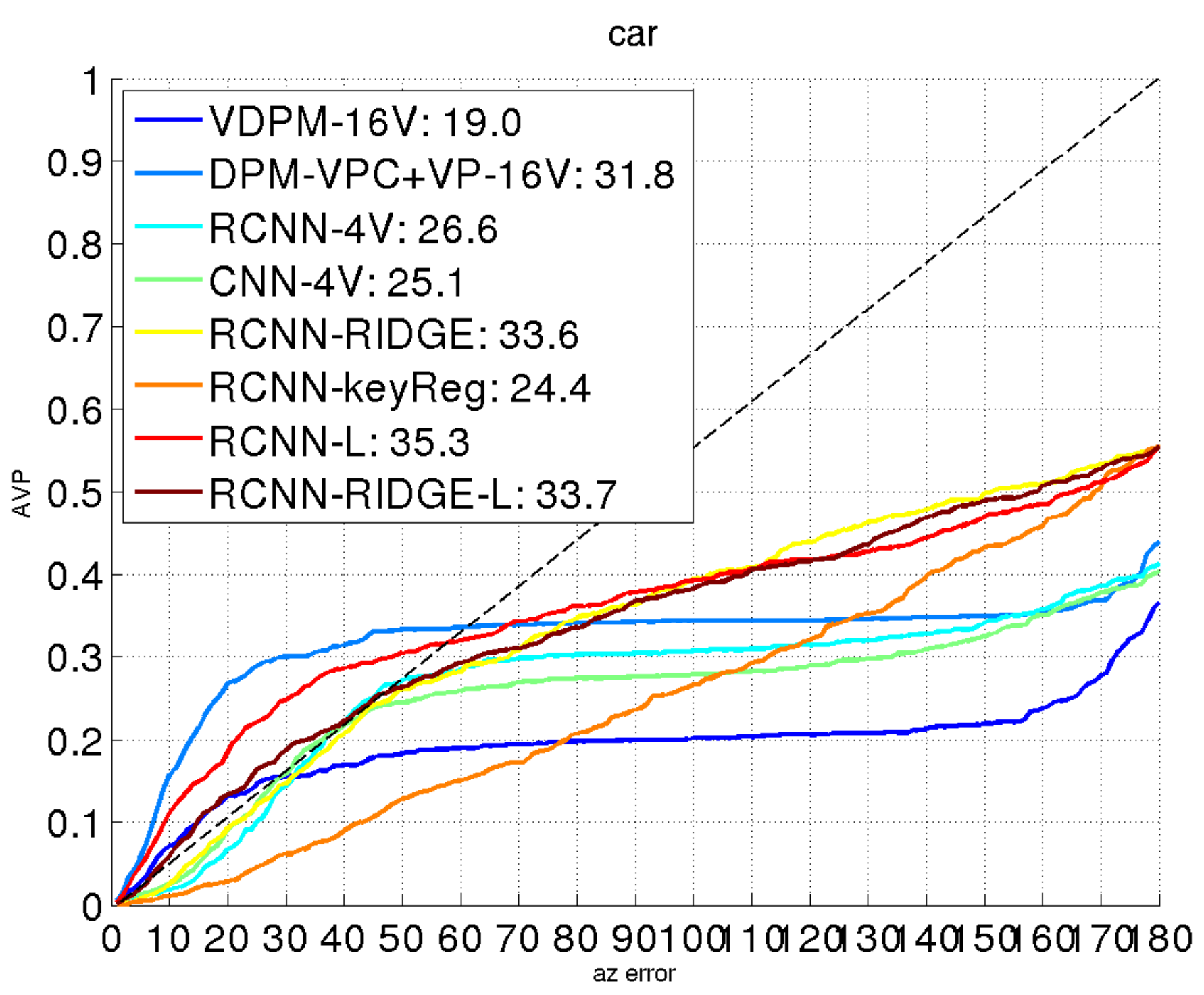}
\includegraphics[width=0.245\textwidth]{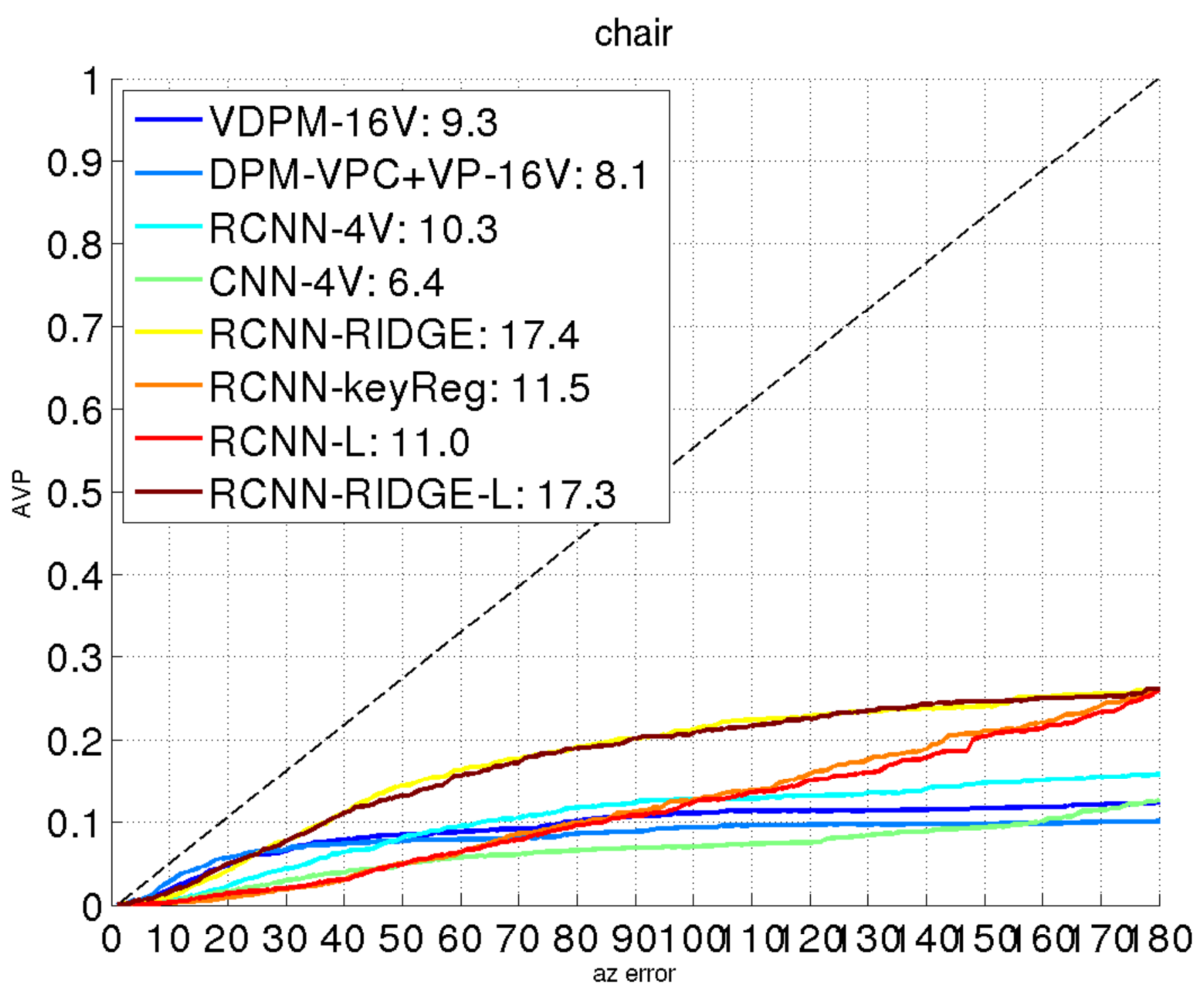}
\includegraphics[width=0.245\textwidth]{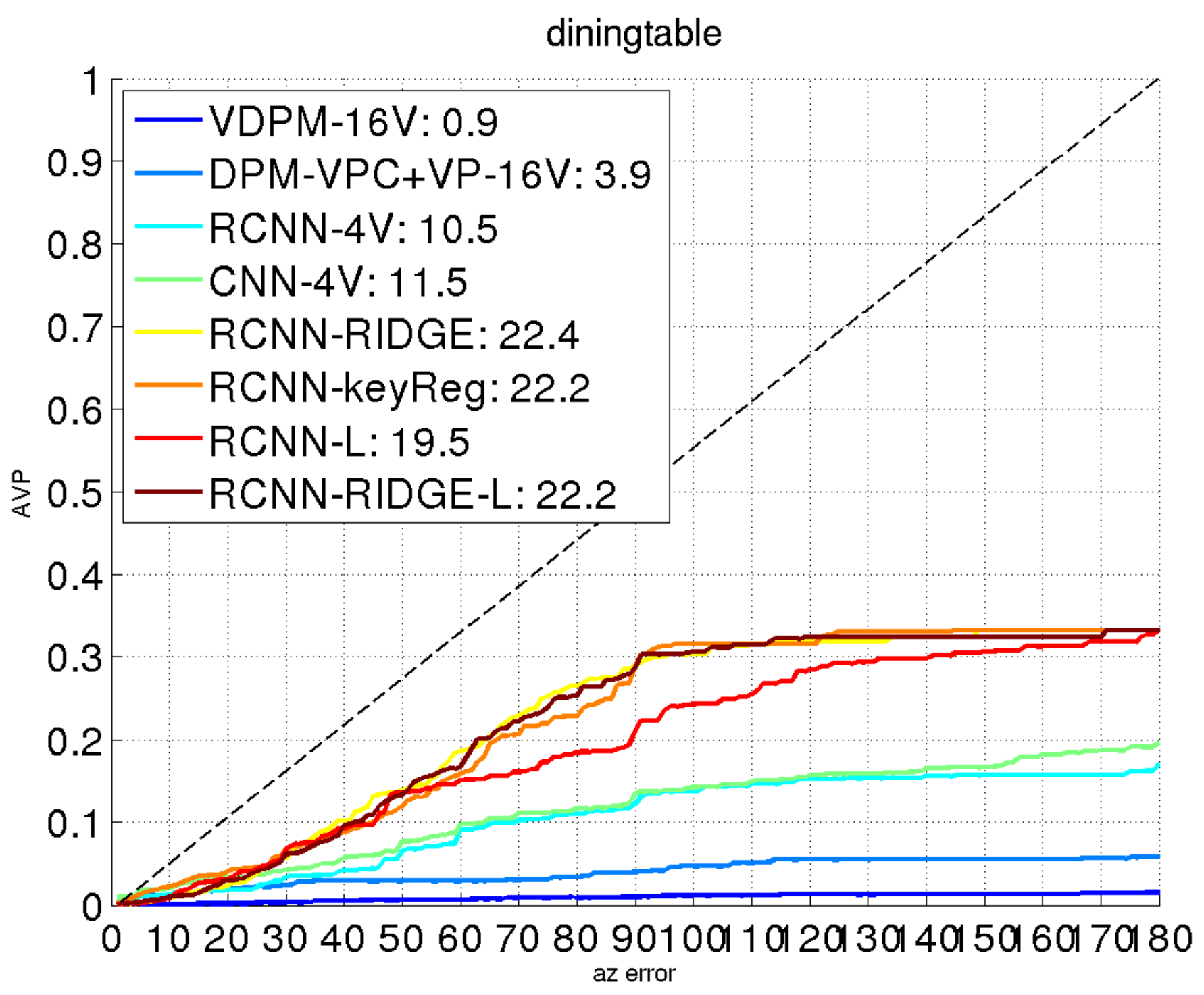}
\includegraphics[width=0.245\textwidth]{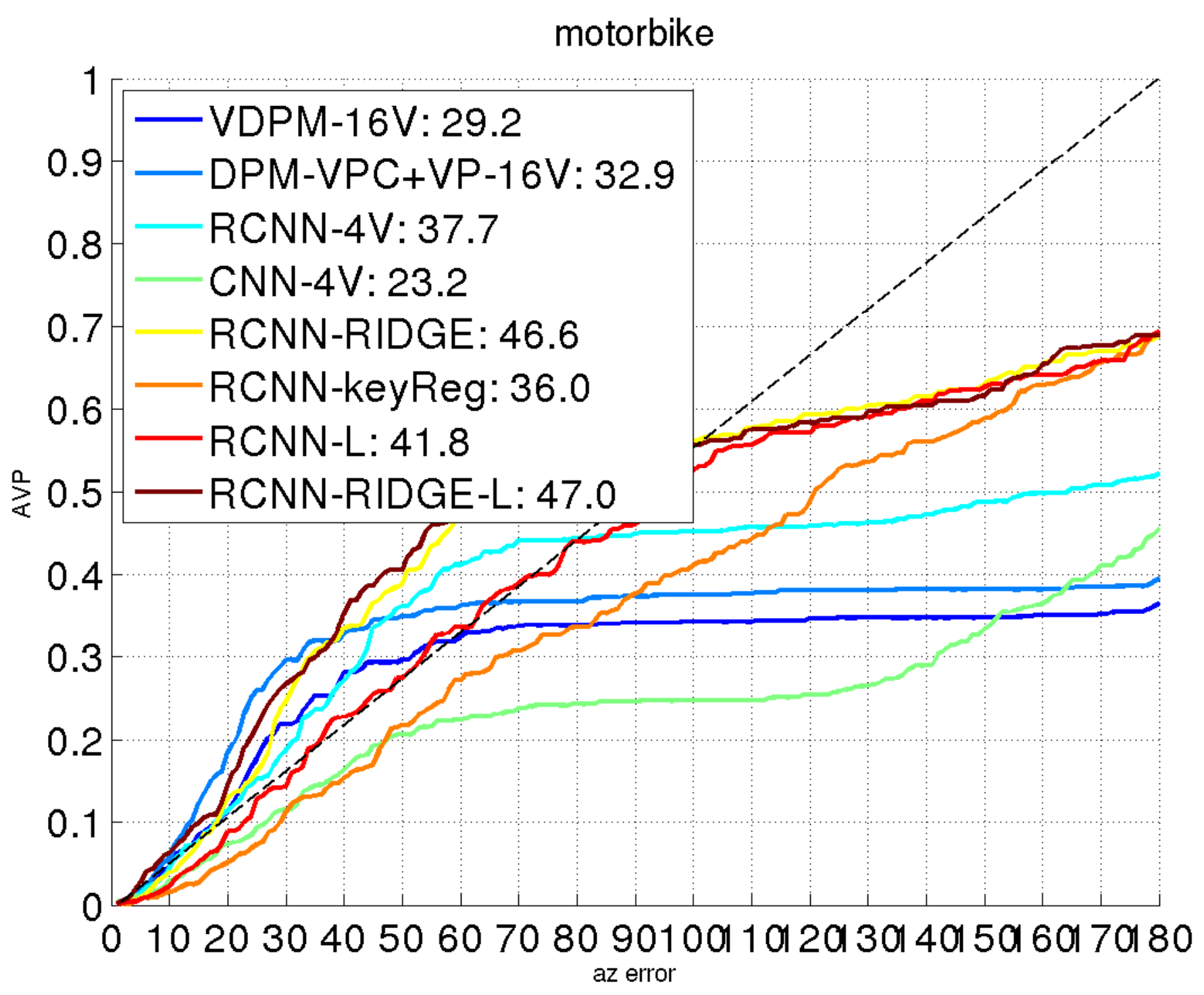}
\includegraphics[width=0.245\textwidth]{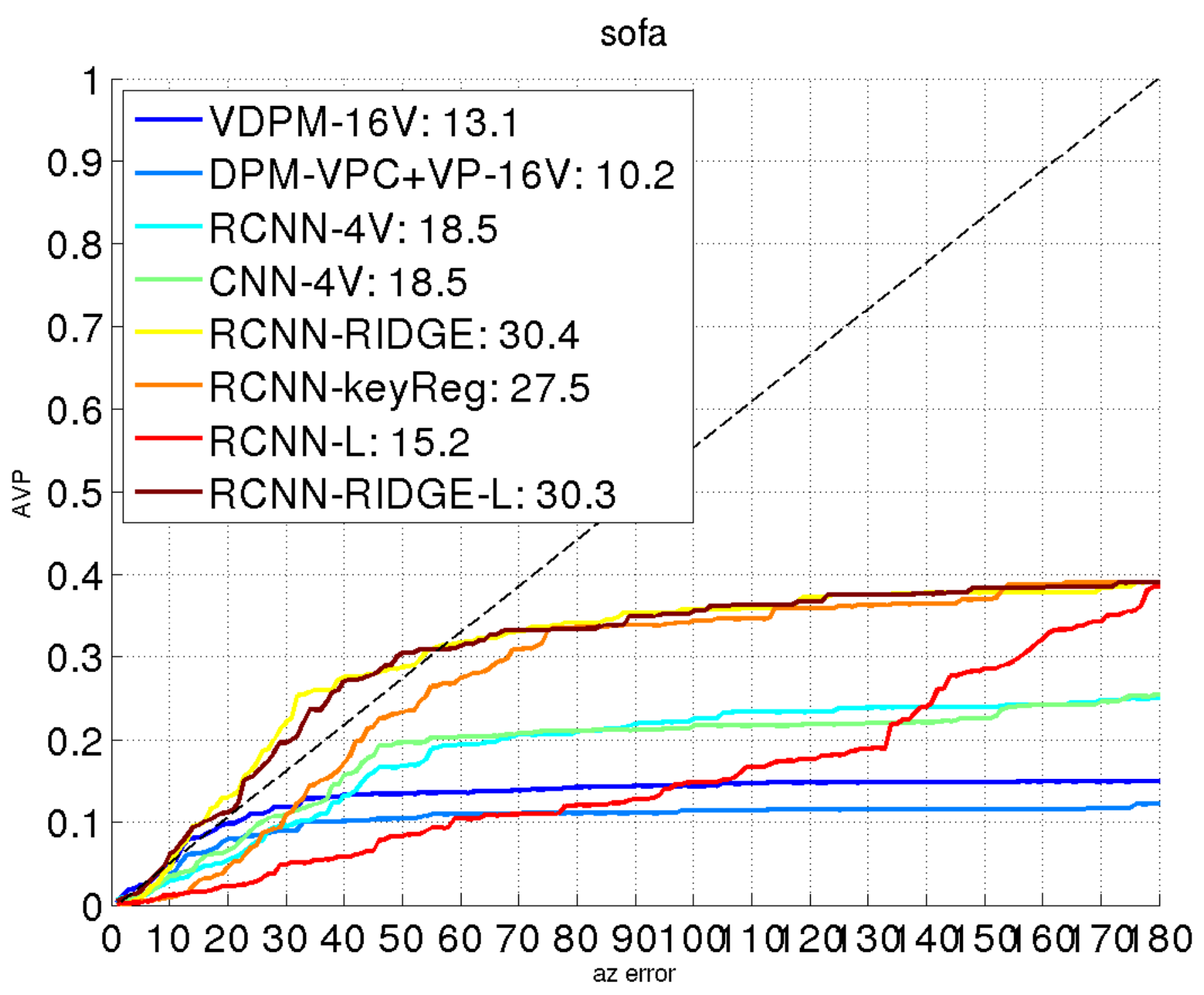}
\includegraphics[width=0.245\textwidth]{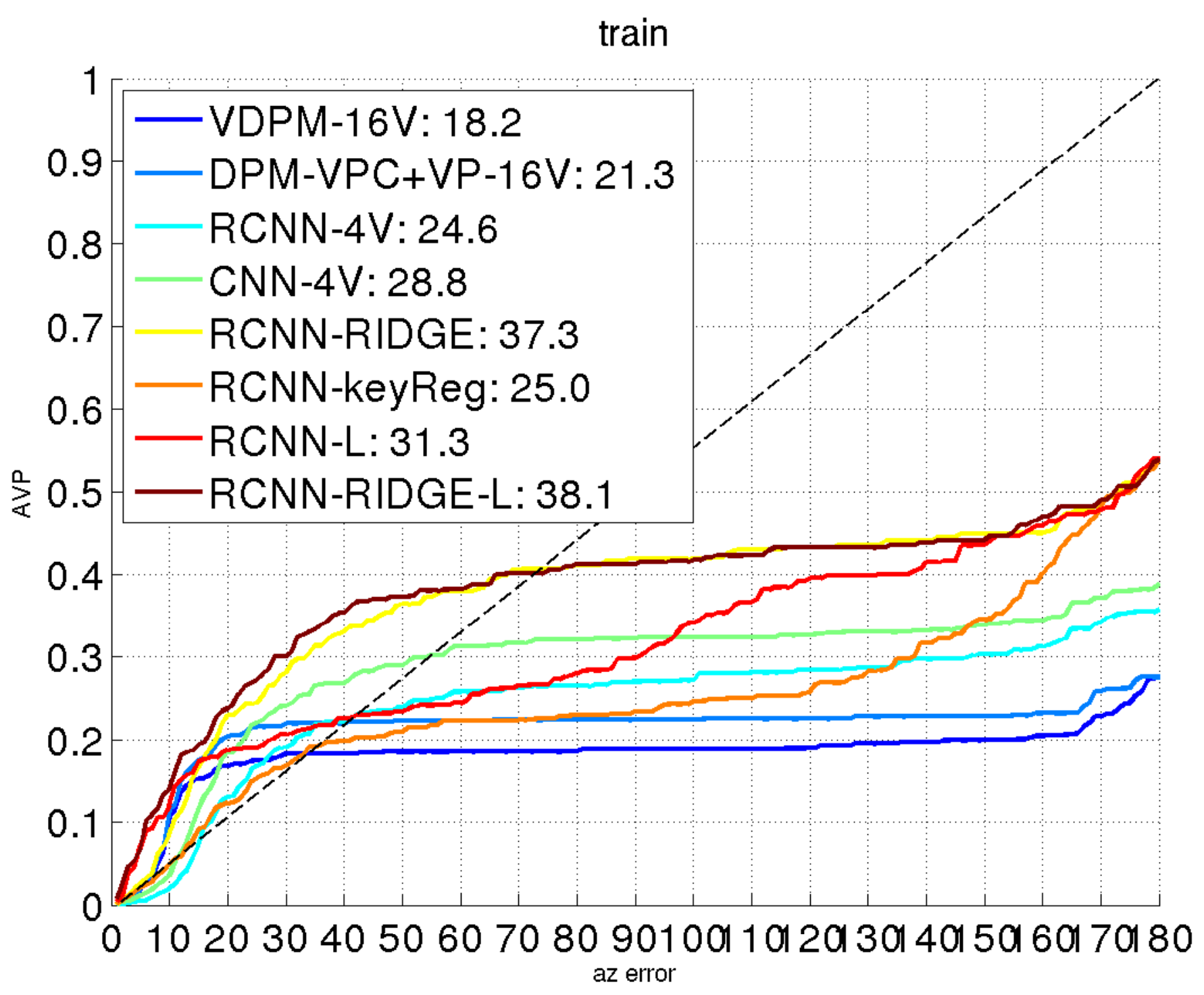}
\includegraphics[width=0.245\textwidth]{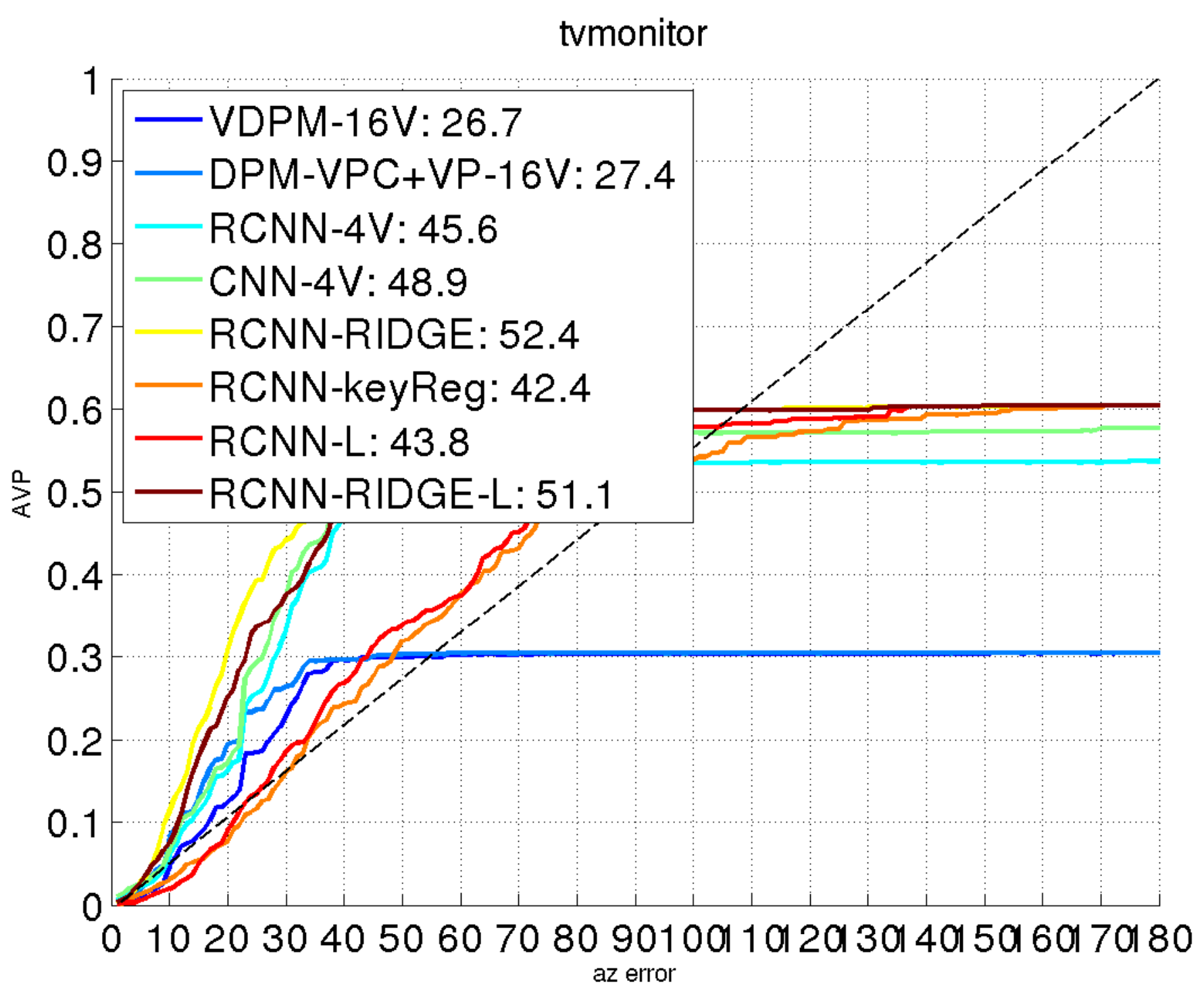}
\caption{Per class simultaneous object localization and viewpoint
  estimation. AVP vs azimuth error curves comparing the 3D lifting
  procedures \rcnnRidgeL and \rcnnl with the best performing
  multi-view and continuous viewpoint predictors (Fig.~\ref{exp:vp}
  (center) and Fig.~\ref{exp:liftPlot} (right) in paper). }
\label{fig:supplAAVP}
\end{figure*}

Tab.~\ref{tab:suppl3Dlift} reports the per-class performance of our
viewpoint guided 3D object detection method \rcnnRidgeL and our 3D
object detection method without any viewpoint guidance \rcnnl, in
comparison to the baseline \keyReg. We again report the AVP$_V$
performance. These results complement the average class results in
Fig.~\ref{exp:vp} (right) in the paper.

On average and across all classes, \rcnnRidgeL and \rcnnl outperform
the baseline \keyReg on all VP$_V$ viewpoint binnings, by large
margins. Moreover, looking into the average results across classes,
\rcnnRidgeL with $38.2\%$, $27.6\%$, $18.6\%$ and $15.8\%$ AVP$_V$
performance outperforms \rcnnl ($35.8\%$, $24.6\%$, $17.7\%$,
$13.4\%$). Focusing on individual classes, \rcnnRidgeL is better on
most of the classes, except on {\em aeroplane}, {\em car} and {\em
  diningtable} where \rcnnl succeeds.

As stated in Sect.~\ref{sec:expLifting} in the paper, in comparison to
the state-of-the-art \vdpm and \dpmvocvp and the \rcnn viewpoint
regressors (Tab.~\ref{tab:supVp}), \rcnnRidgeL outperforms \rcnnRidge
by $1.2\%$ AVP$_4$ and $2.2\%$ AVP$_8$ on average. It is also better
than \dpmvocvp by $1.3\%$ AVP$_{16}$ and $2.2\%$ AVP$_{24}$, achieving
state-of-the-art performance on Pascal3D+. Additionally, focusing on
the individual classes again, \rcnnRidgeL constantly improves over
\rcnnRidge on all viewpoint splits, especially on the fine-viewpoint
binnings (VP$_{16}$ and VP$_{24}$).

\begin{figure*}
\includegraphics[height=1.8cm]{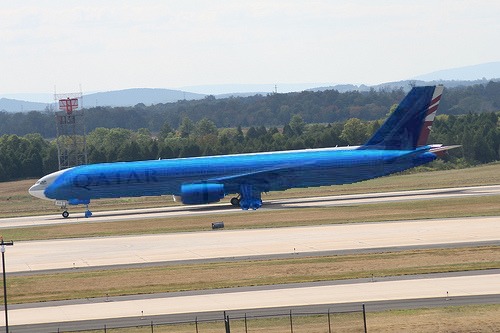}
\includegraphics[height=1.8cm]{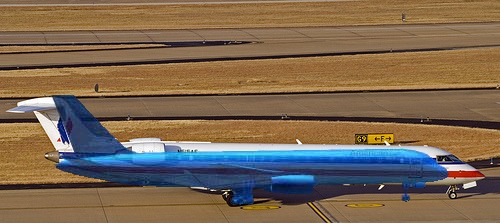}
\includegraphics[height=1.8cm]{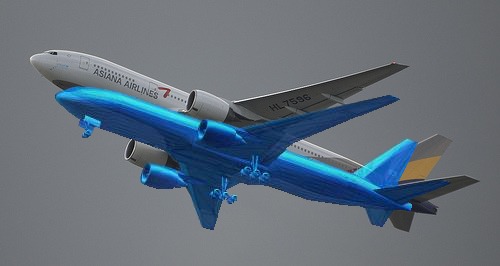}
\includegraphics[height=1.8cm]{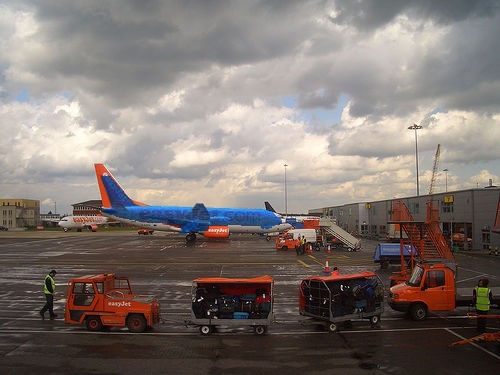}
\includegraphics[height=1.8cm]{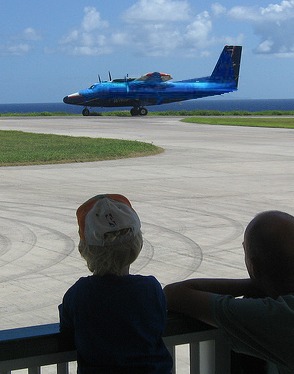}
\includegraphics[height=1.8cm]{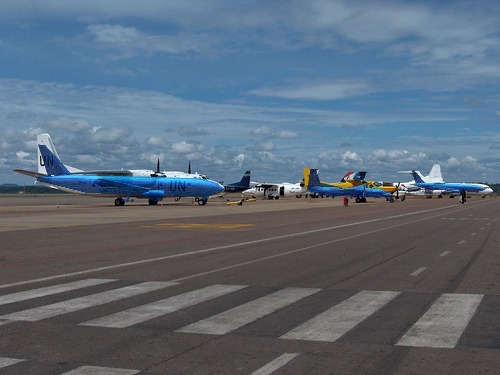}
\includegraphics[height=1.8cm]{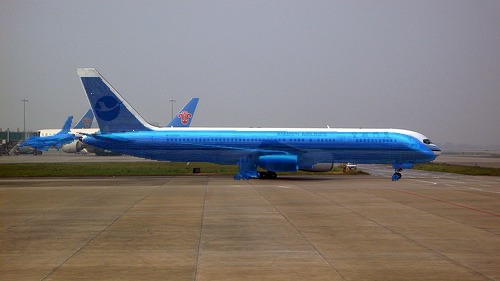}
\includegraphics[height=1.8cm]{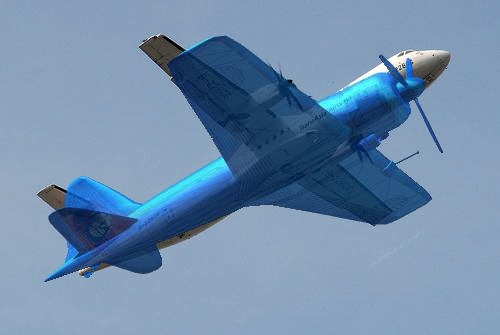}
\includegraphics[height=1.8cm]{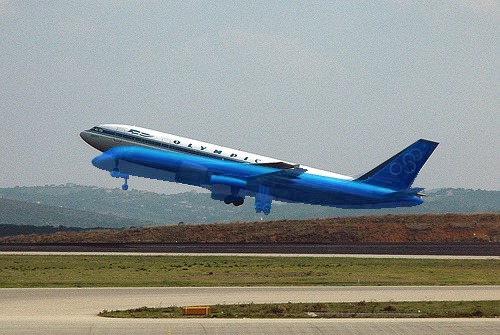}
\includegraphics[height=1.8cm]{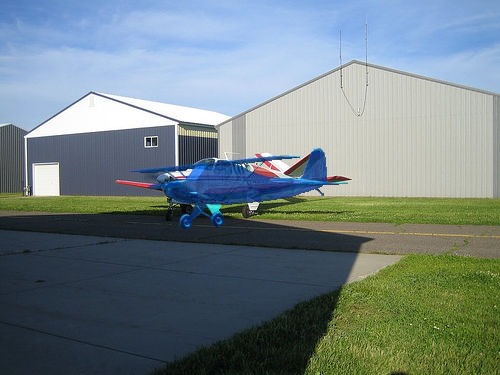}
\includegraphics[height=1.8cm]{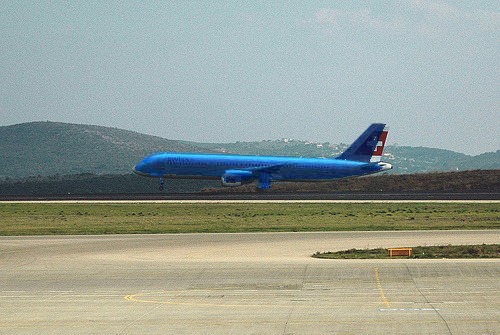}
\includegraphics[height=1.8cm]{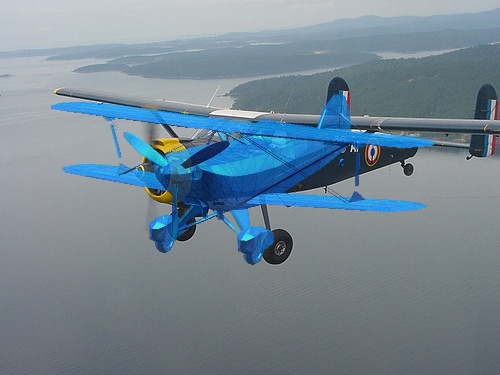}
 \includegraphics[height=1.8cm]{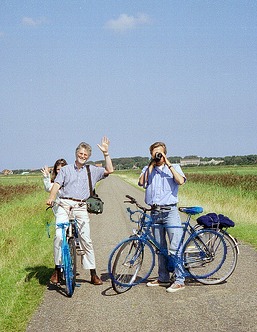}
 \includegraphics[height=1.8cm]{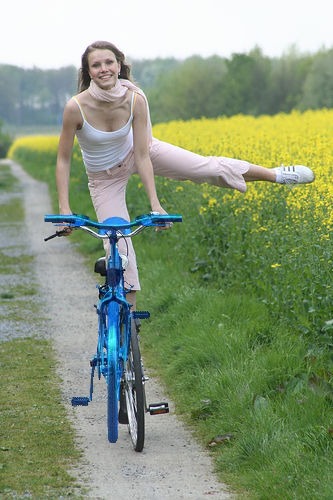}
 \includegraphics[height=1.8cm]{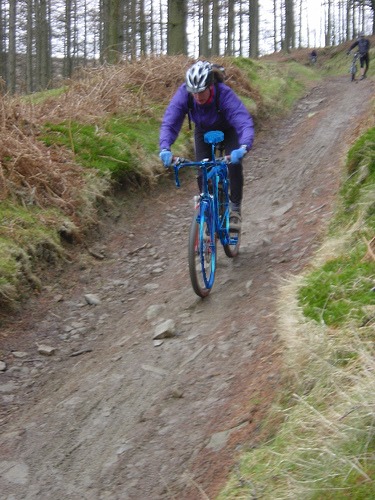}
 \includegraphics[height=1.8cm]{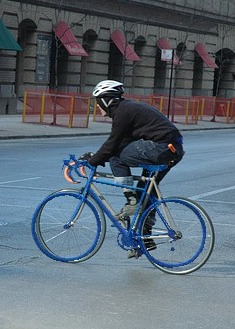}
 \includegraphics[height=1.8cm]{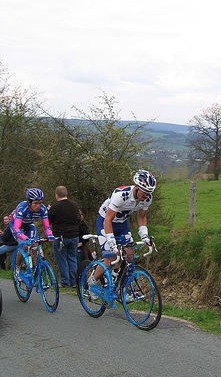}
 \includegraphics[height=1.8cm]{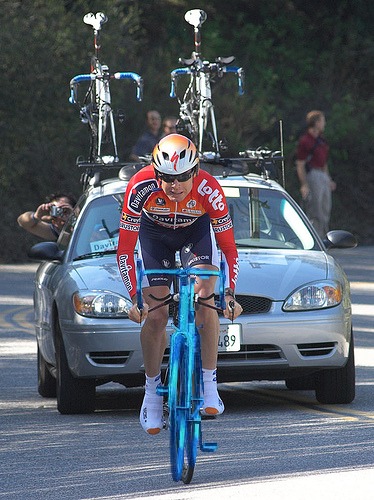}
\includegraphics[height=1.8cm]{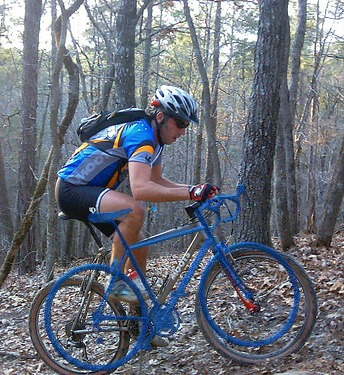}
\includegraphics[height=1.8cm]{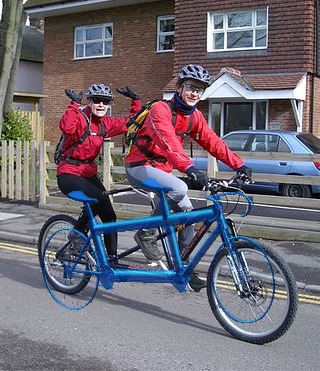}
\includegraphics[height=1.8cm]{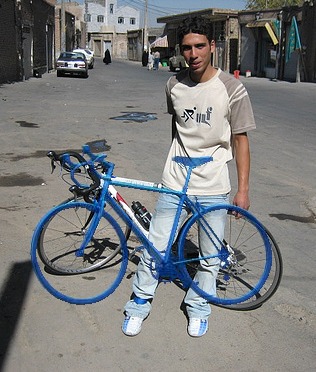}
\includegraphics[height=1.8cm]{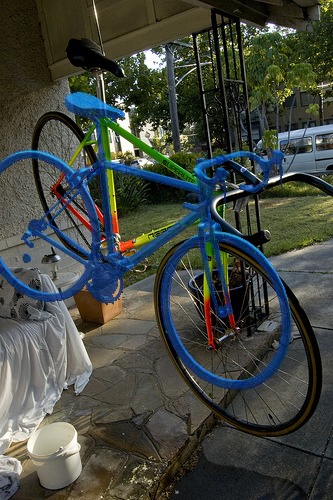}
\includegraphics[height=1.8cm]{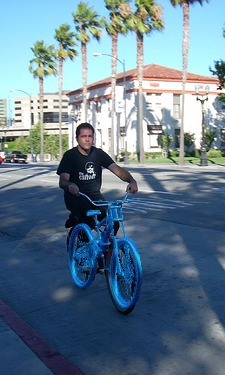}
\includegraphics[height=1.8cm]{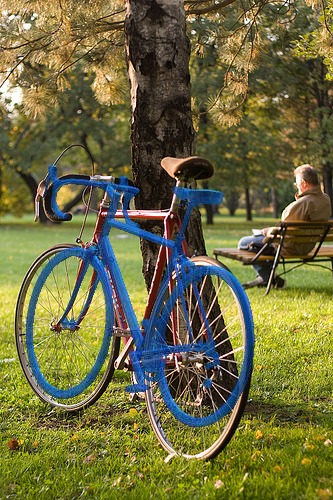}
\includegraphics[height=1.8cm]{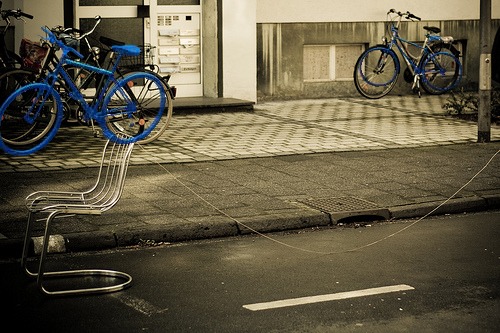}
\includegraphics[height=1.8cm]{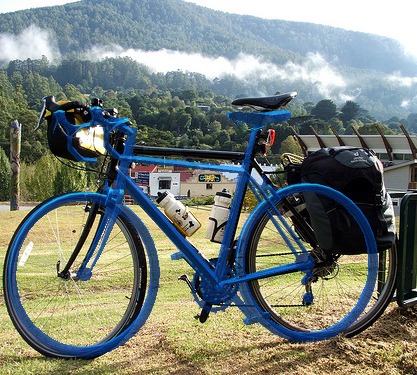}
\includegraphics[height=1.8cm]{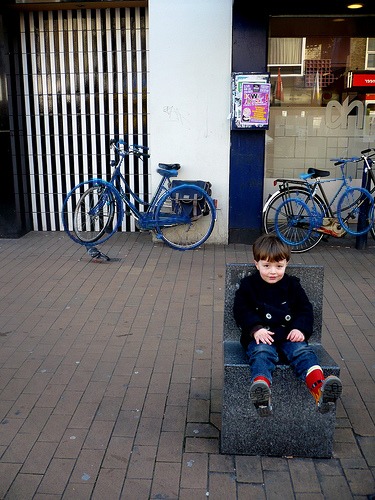}
\includegraphics[height=1.8cm]{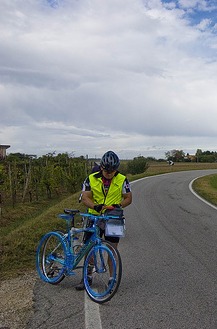}
\includegraphics[height=1.8cm]{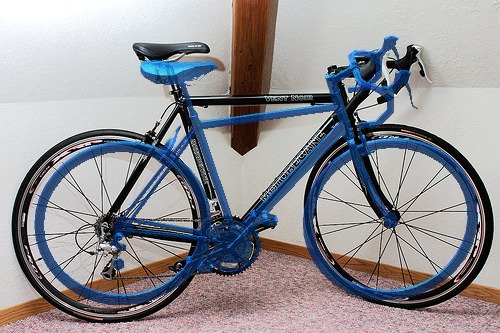}
\includegraphics[height=1.8cm]{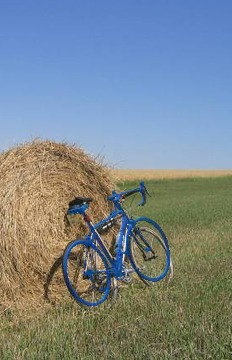}
\includegraphics[height=1.8cm]{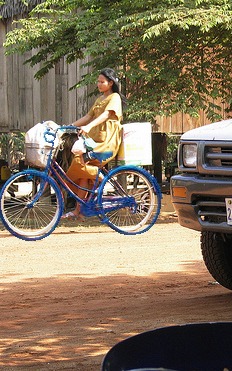}
\includegraphics[height=1.8cm]{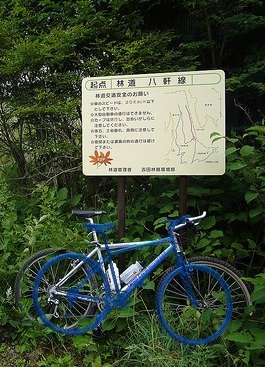}
\includegraphics[height=1.8cm]{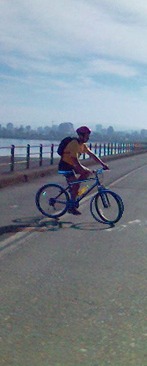}
\includegraphics[height=1.8cm]{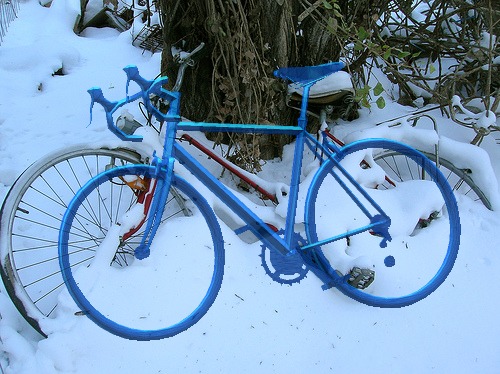}
\includegraphics[height=1.8cm]{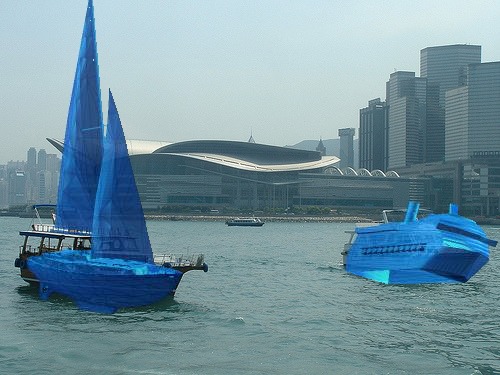}
\includegraphics[height=1.8cm]{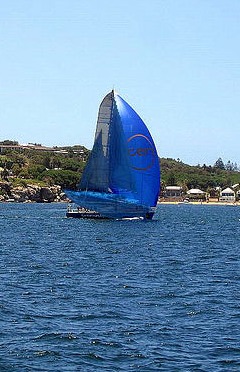}
\includegraphics[height=1.8cm]{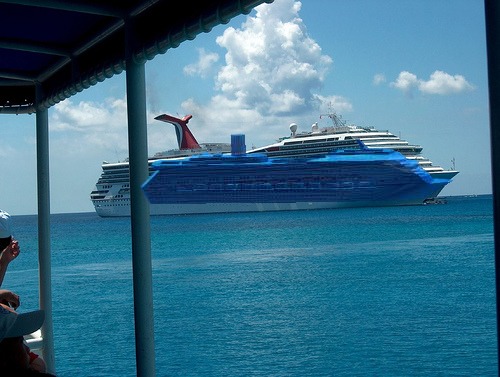}
\includegraphics[height=1.8cm]{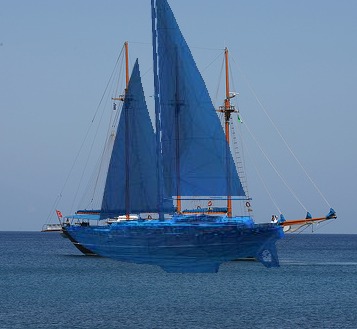}
\includegraphics[height=1.8cm]{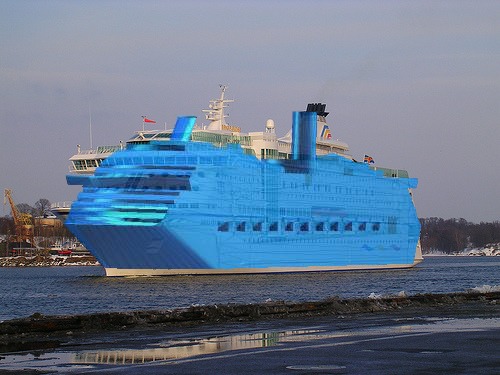}
\includegraphics[height=1.8cm]{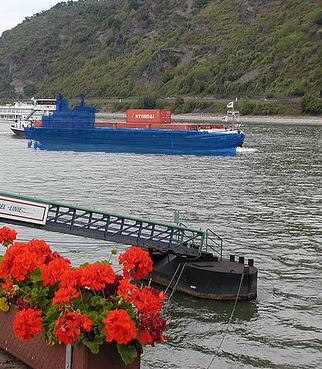}
\includegraphics[height=1.8cm]{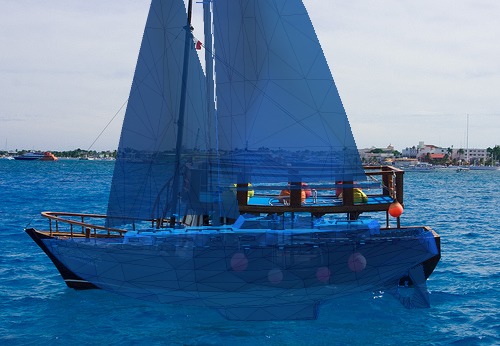}
\includegraphics[height=1.8cm]{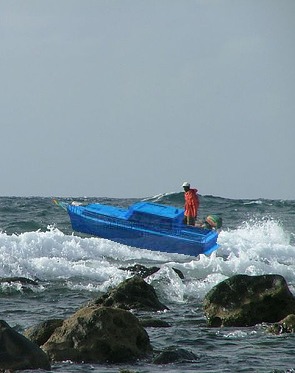}
\includegraphics[height=1.8cm]{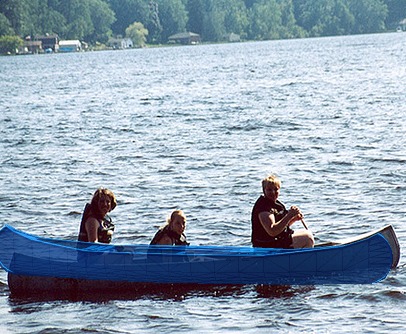}
\includegraphics[height=1.8cm]{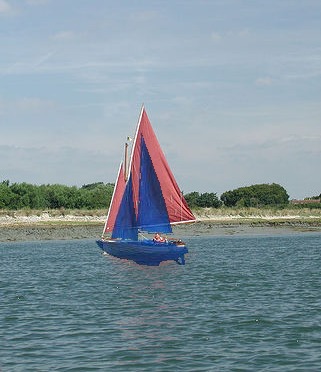}
\includegraphics[height=1.8cm]{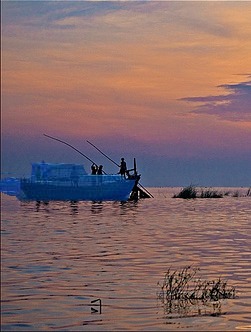}
\includegraphics[height=1.8cm]{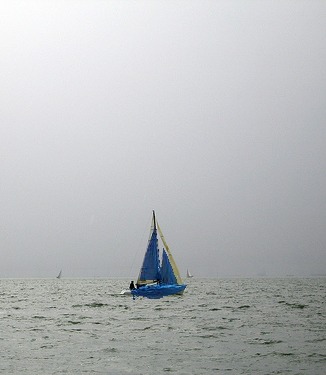}
\includegraphics[height=1.8cm]{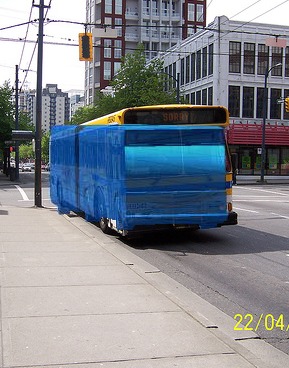}
\includegraphics[height=1.8cm]{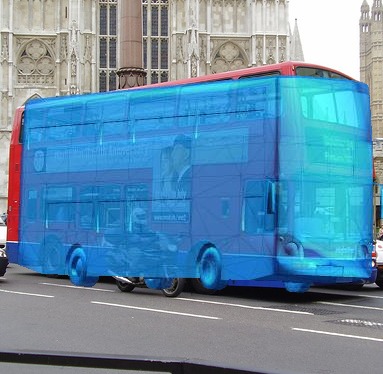}
\includegraphics[height=1.8cm]{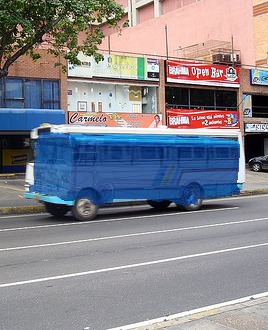}
\includegraphics[height=1.8cm]{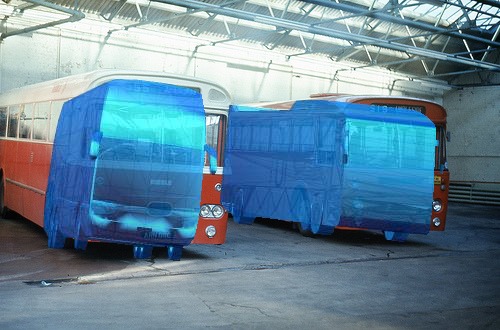}
\includegraphics[height=1.8cm]{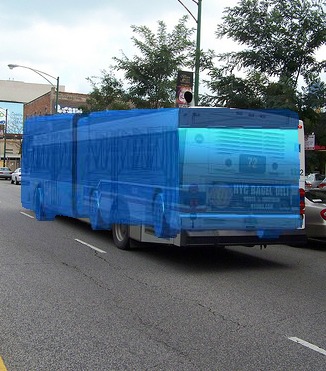}
\includegraphics[height=1.8cm]{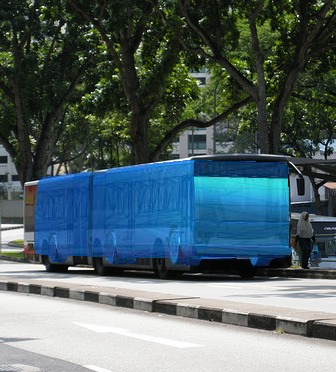}
\includegraphics[height=1.8cm]{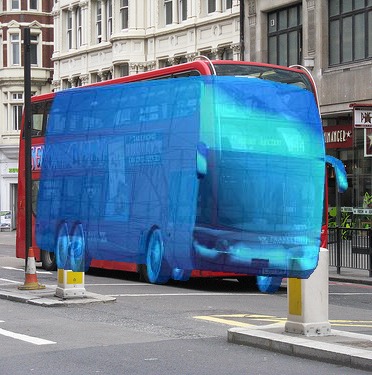}
\includegraphics[height=1.8cm]{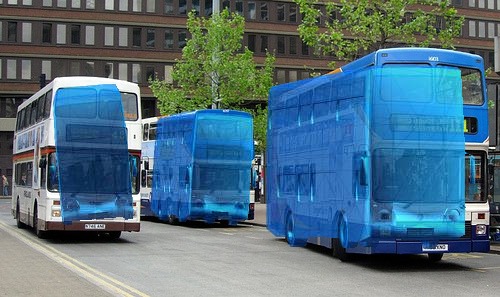}
\includegraphics[height=1.8cm]{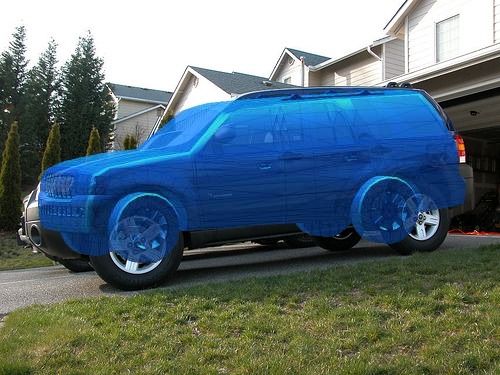}
\includegraphics[height=1.8cm]{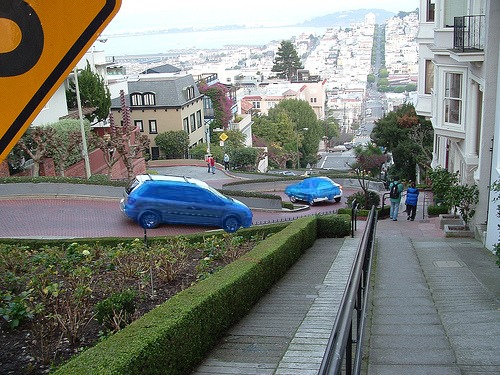}
\includegraphics[height=1.8cm]{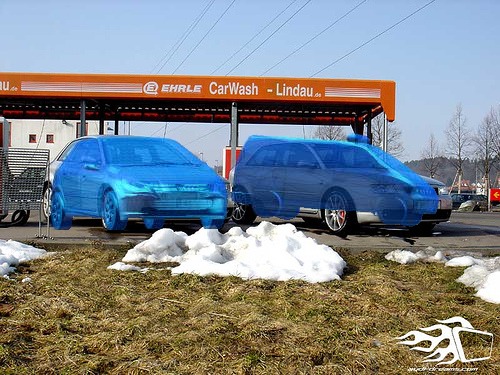}
\includegraphics[height=1.8cm]{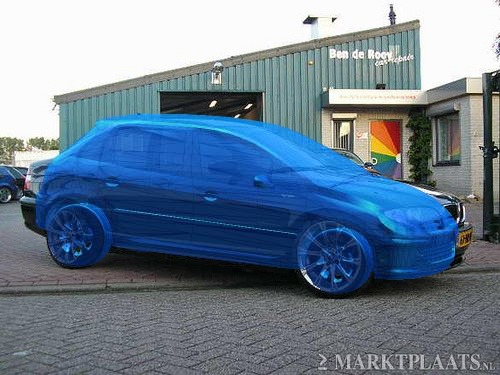}
\includegraphics[height=1.8cm]{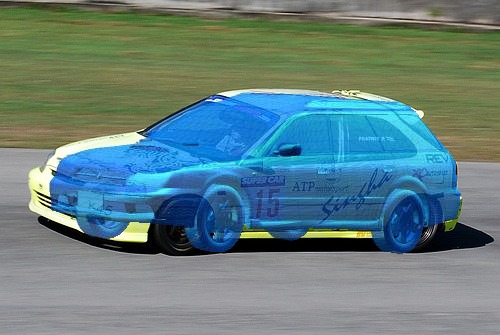}
\includegraphics[height=1.8cm]{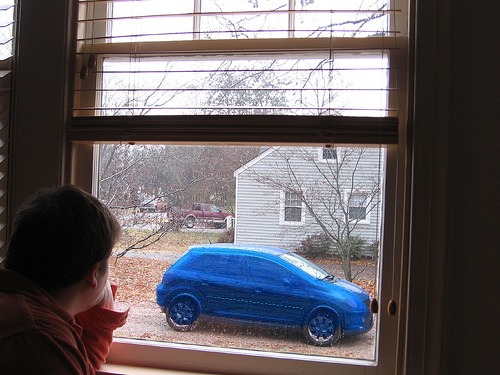}
\includegraphics[height=1.8cm]{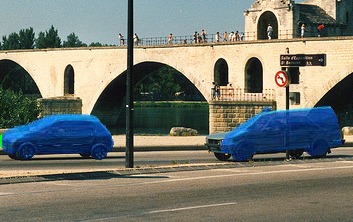}
\includegraphics[height=1.8cm]{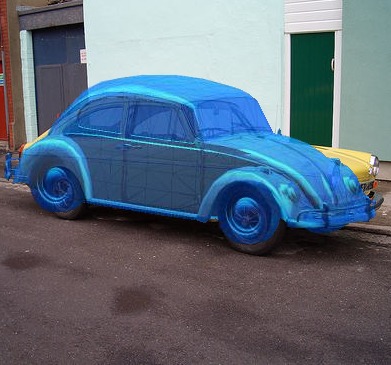}
\includegraphics[height=1.8cm]{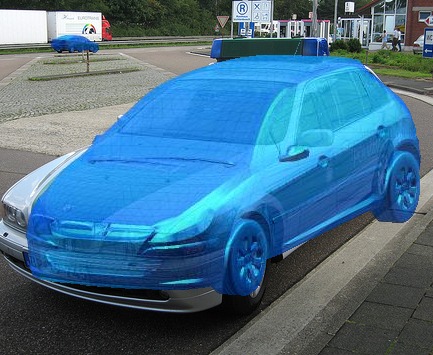}
\includegraphics[height=1.8cm]{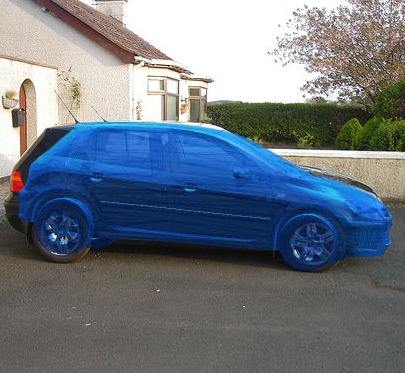}
\includegraphics[height=1.8cm]{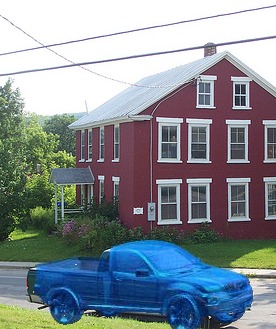}
\includegraphics[height=1.8cm]{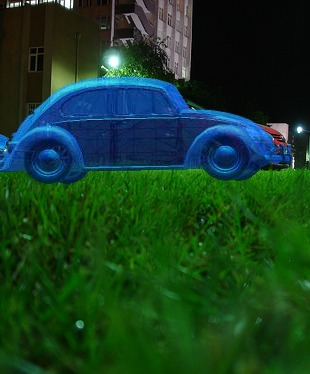}
\includegraphics[height=1.8cm]{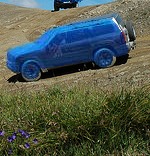}
\includegraphics[height=1.8cm]{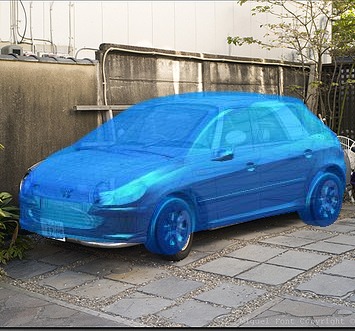}
\includegraphics[height=1.8cm]{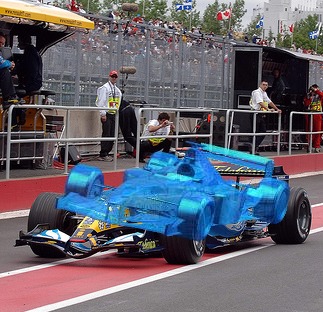}
\includegraphics[height=1.8cm]{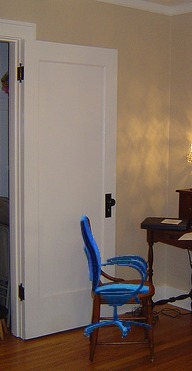}
\includegraphics[height=1.8cm]{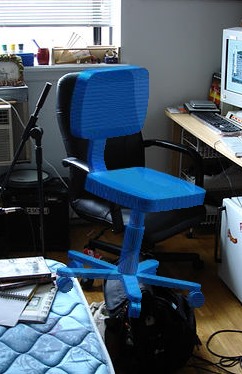}
\includegraphics[height=1.8cm]{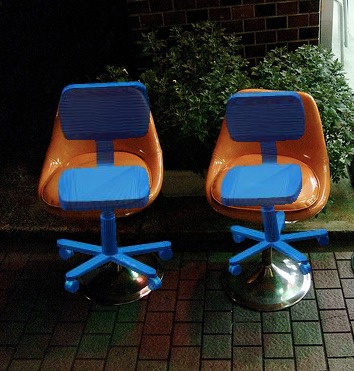}
\includegraphics[height=1.8cm]{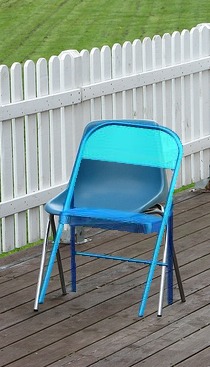}
\includegraphics[height=1.8cm]{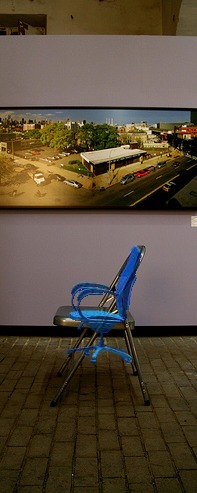}
\includegraphics[height=1.8cm]{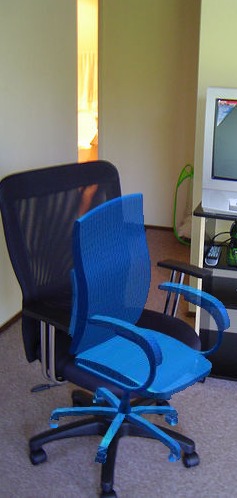}
\includegraphics[height=1.8cm]{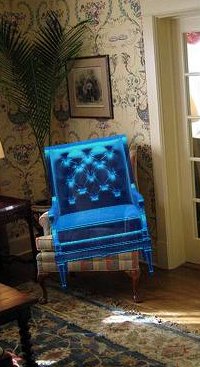}
\includegraphics[height=1.825cm]{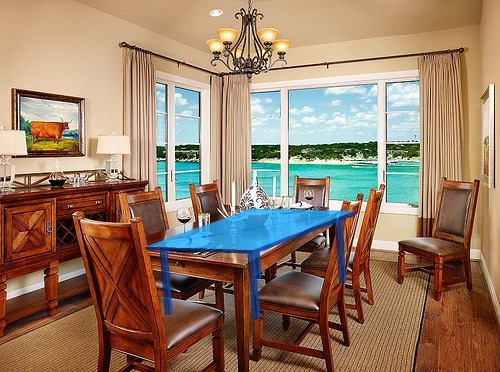}
\includegraphics[height=1.8cm]{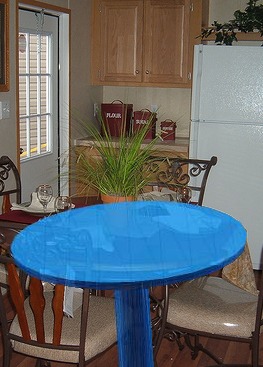}
\includegraphics[height=1.8cm]{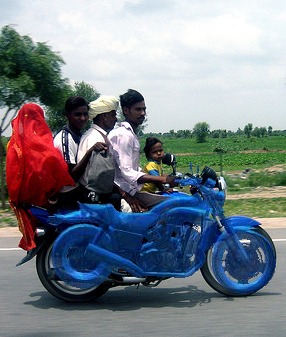}
\includegraphics[height=1.8cm]{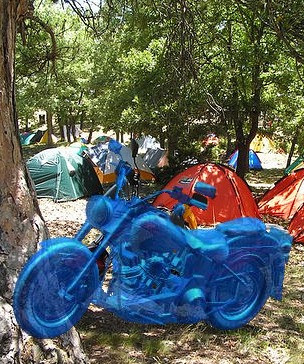}
\includegraphics[height=1.8cm]{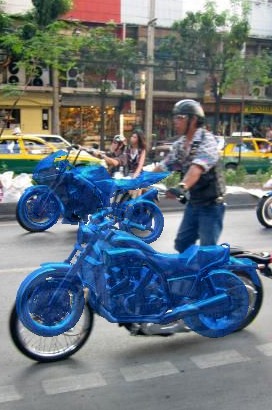}
\includegraphics[height=1.8cm]{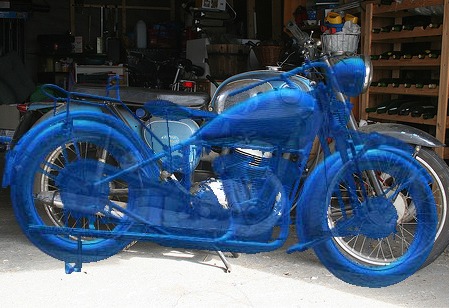}
\includegraphics[height=1.8cm]{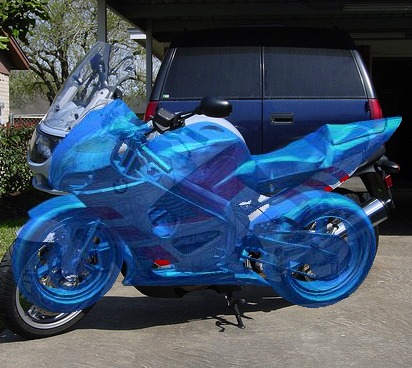}
\includegraphics[height=1.8cm]{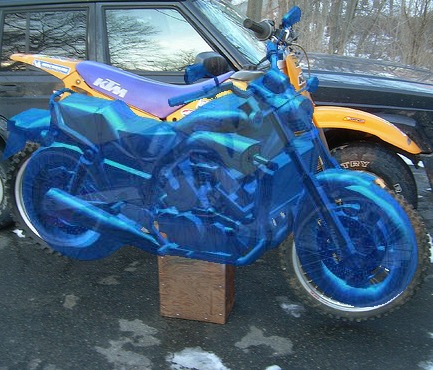}
\includegraphics[height=1.8cm]{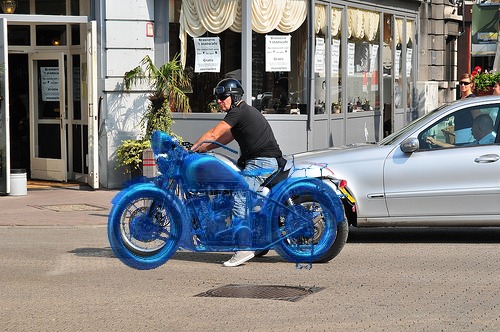}
\includegraphics[height=1.8cm]{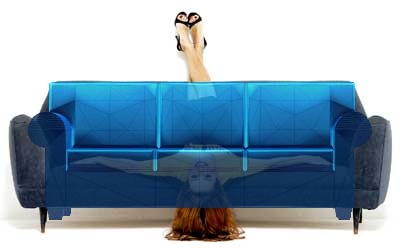}
\includegraphics[height=1.8cm]{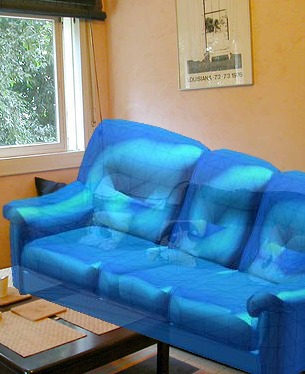}
\includegraphics[height=1.8cm]{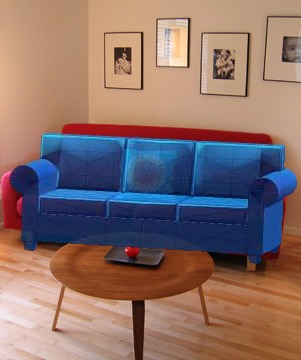}
\includegraphics[height=1.8cm]{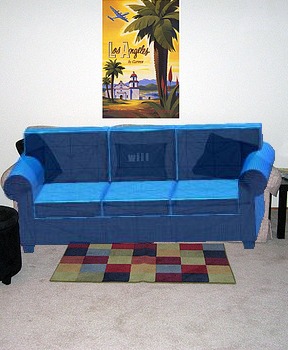}
\includegraphics[height=1.8cm]{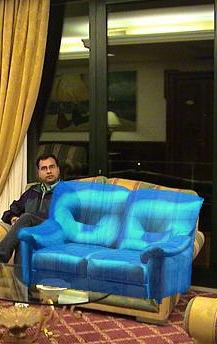}
\includegraphics[height=1.8cm]{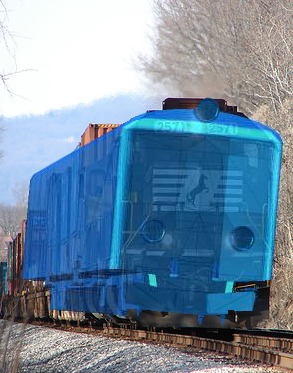}
\includegraphics[height=1.8cm]{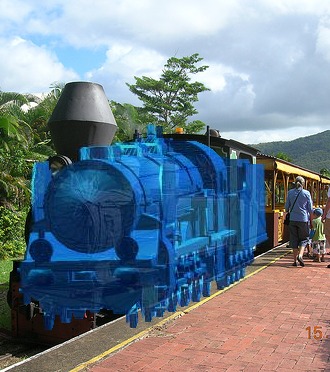}
\includegraphics[height=1.8cm]{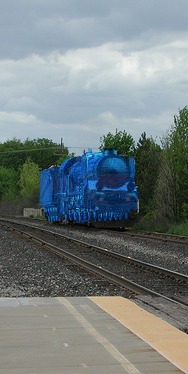}
\includegraphics[height=1.8cm]{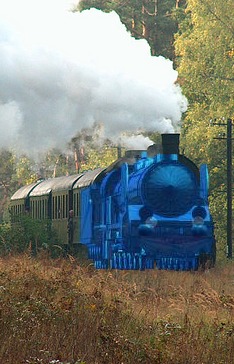}
\includegraphics[height=1.8cm]{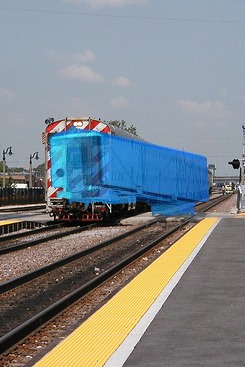}
\includegraphics[height=1.8cm]{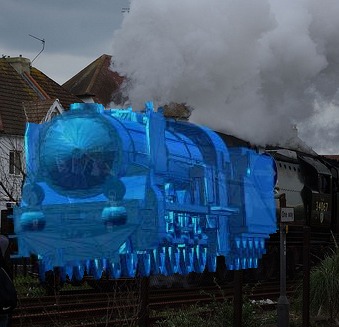}
\includegraphics[height=1.8cm]{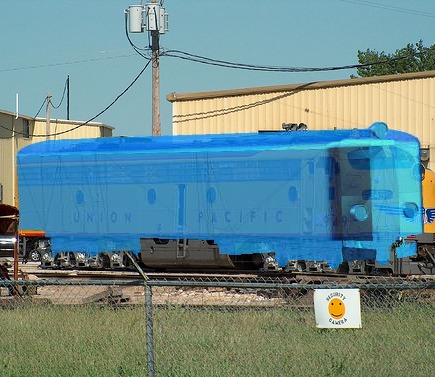}
\includegraphics[height=1.8cm]{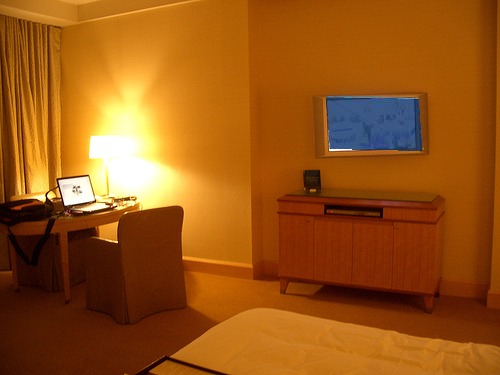}
\includegraphics[height=1.8cm]{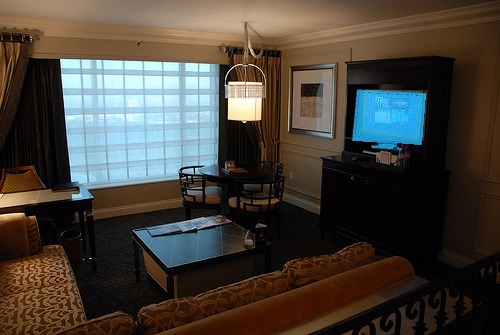}
\includegraphics[height=1.8cm]{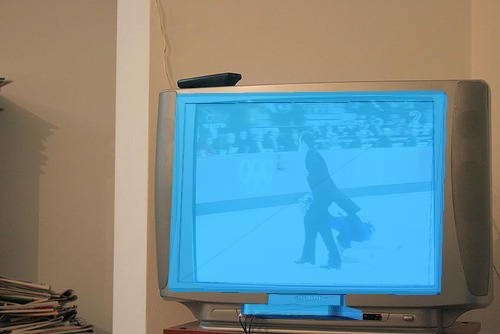}
\includegraphics[height=1.8cm]{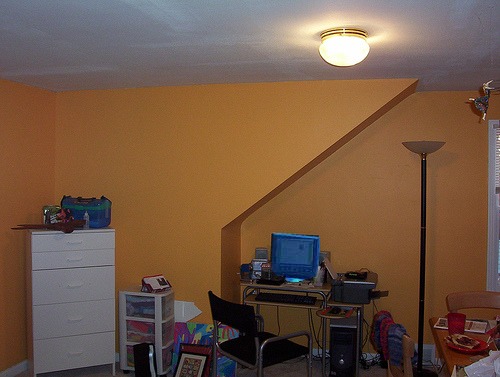}
\includegraphics[height=1.8cm]{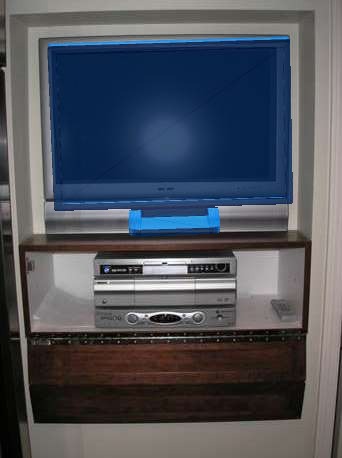}
\caption{3D CAD prototype alignment examples on all Pascal3D+~\cite{xiang14wacv}
  classes. \rcnnRidgeL successfully predicts the 3D shape and viewpoint in many cases.}
\label{fig:supplGoodExamples}
\end{figure*}

\begin{figure*}
\includegraphics[height=1.8cm]{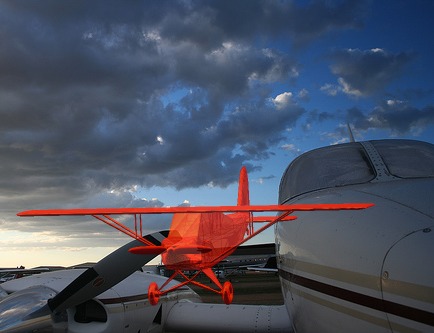}
\includegraphics[height=1.8cm]{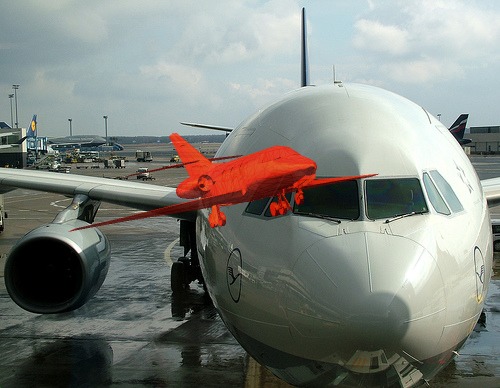}
\includegraphics[height=1.8cm]{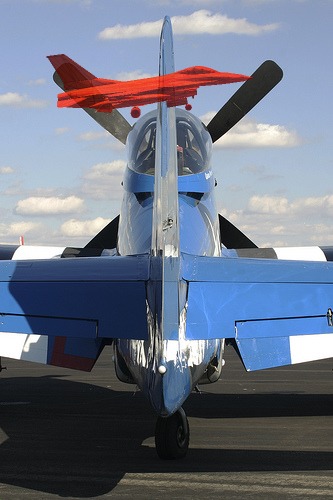}
\includegraphics[height=1.8cm]{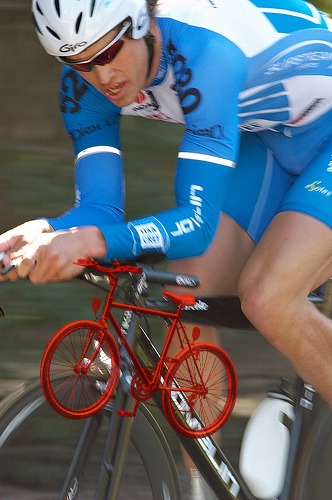}
\includegraphics[height=1.8cm]{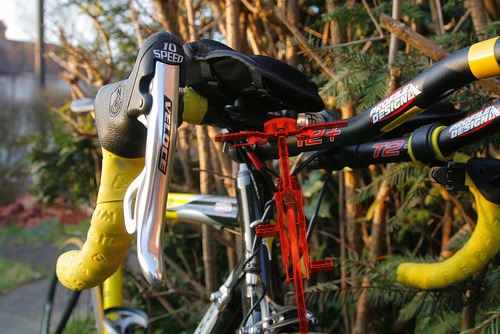}
\includegraphics[height=1.8cm]{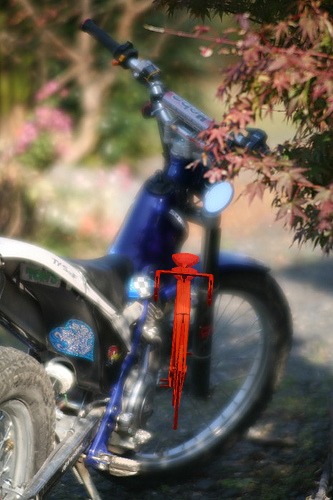}
\includegraphics[height=1.8cm]{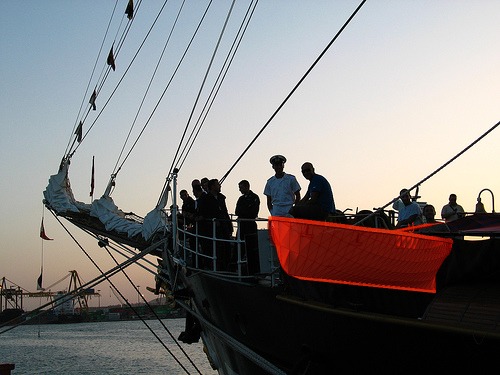}
\includegraphics[height=1.8cm]{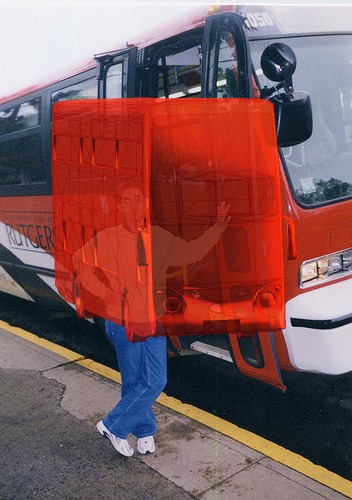}
\includegraphics[height=1.8cm]{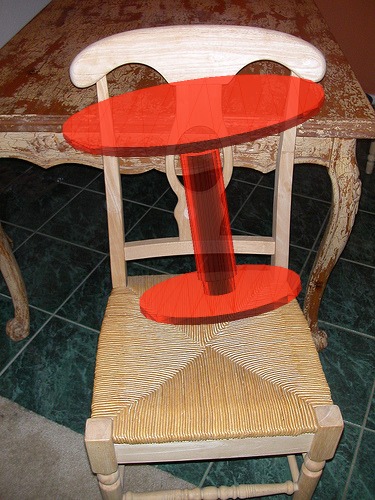}
\includegraphics[height=1.8cm]{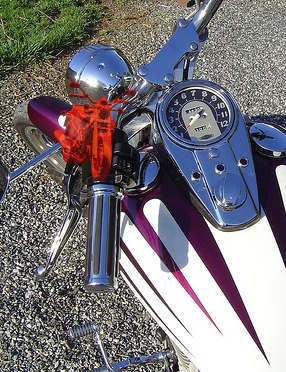}~
\includegraphics[height=1.8cm]{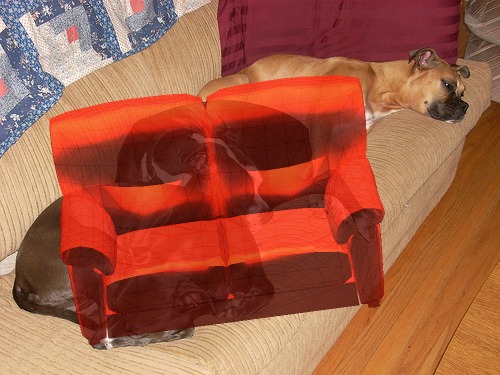}~
\includegraphics[height=1.8cm]{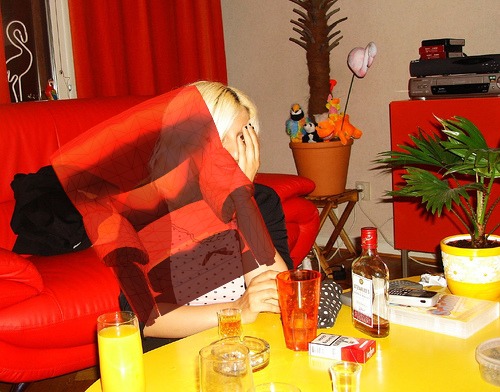}~
\includegraphics[height=1.8cm]{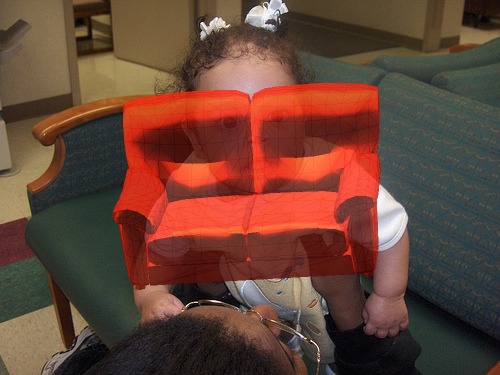}~
\includegraphics[height=1.8cm]{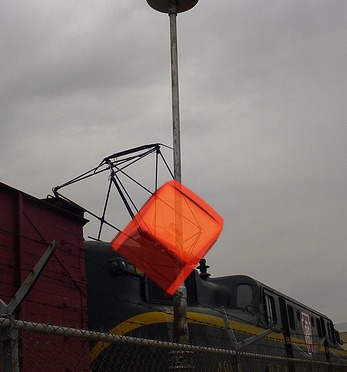}~
\includegraphics[height=1.8cm]{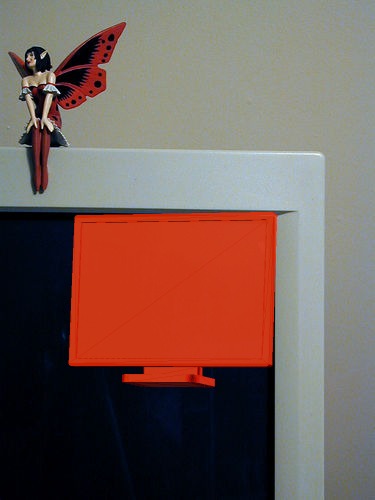}~
\includegraphics[height=1.8cm]{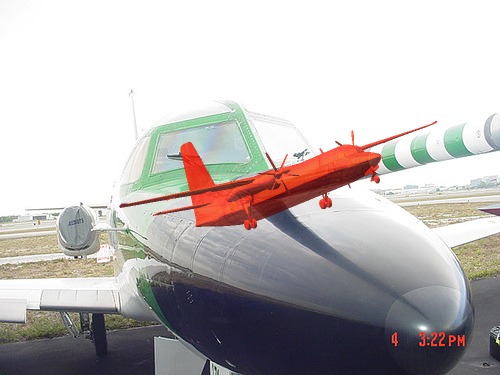}~
\includegraphics[height=1.8cm]{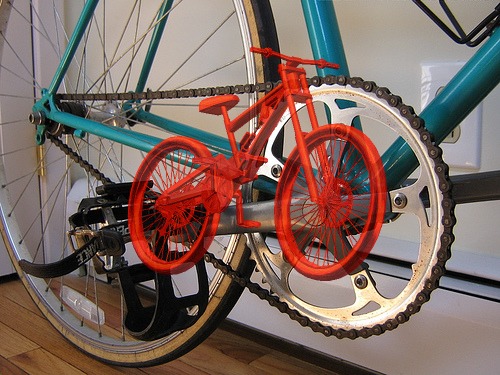}
\\
\includegraphics[height=1.8cm]{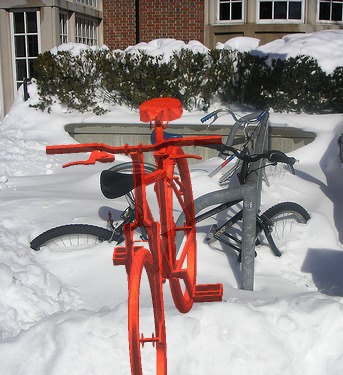}
\includegraphics[height=1.8cm]{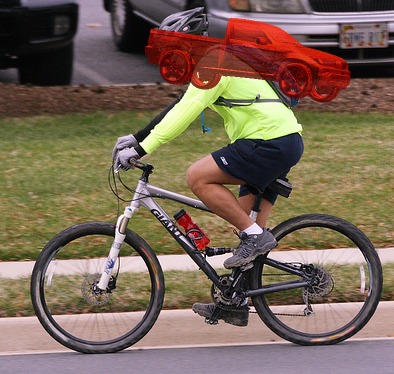}
\includegraphics[height=1.8cm]{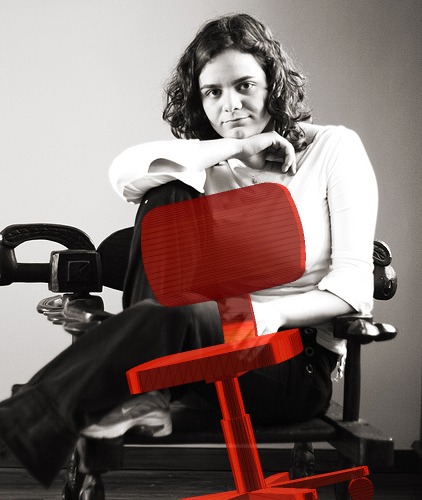}
\includegraphics[height=1.8cm]{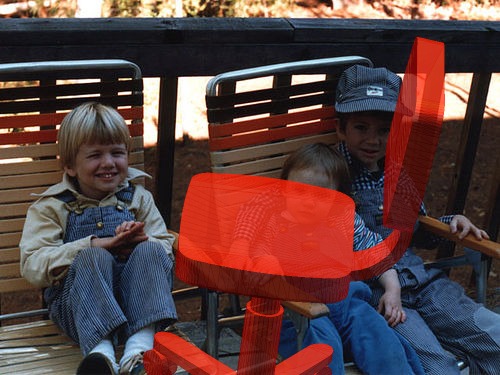}
\includegraphics[height=1.8cm]{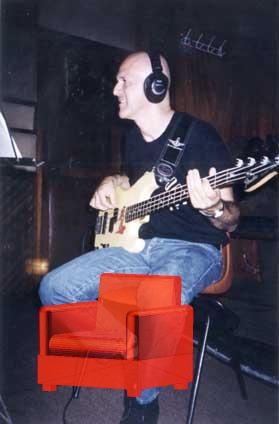}
\includegraphics[height=1.8cm]{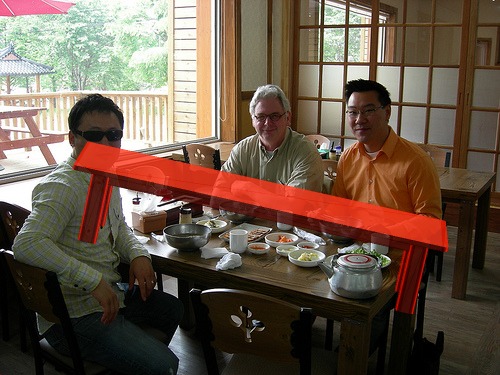}
\includegraphics[height=1.8cm]{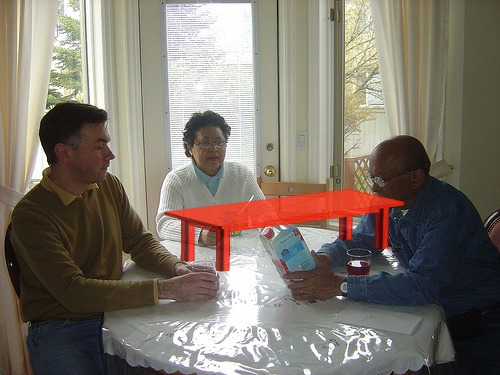}
\includegraphics[height=1.8cm]{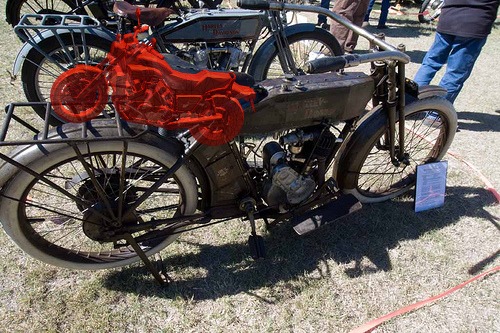}
\includegraphics[height=1.8cm]{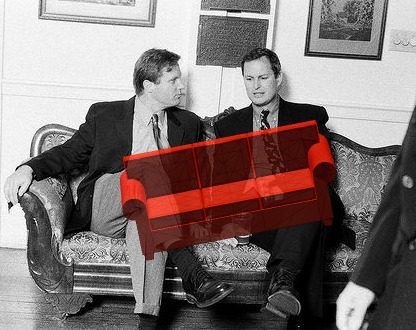}
\includegraphics[height=1.8cm]{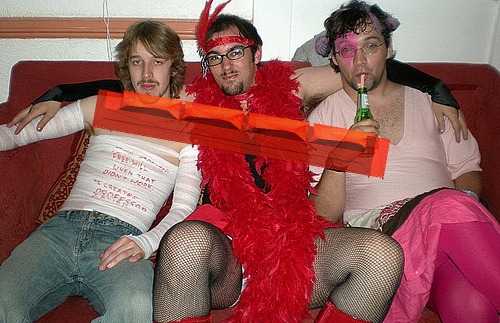}
\includegraphics[height=1.8cm]{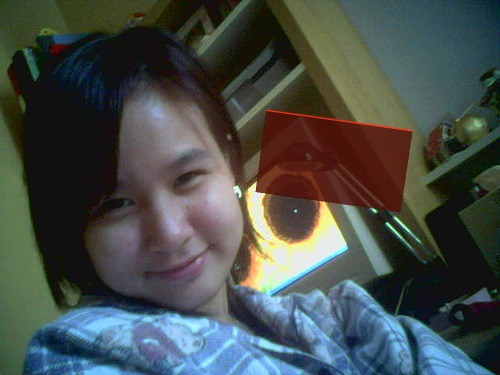}
\includegraphics[height=1.8cm]{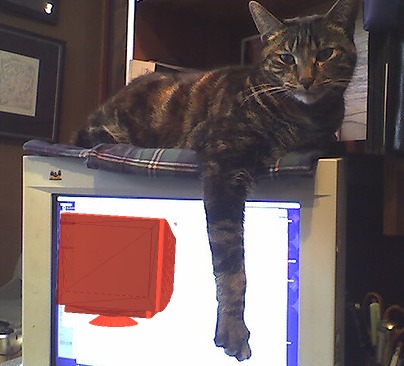}
\includegraphics[height=1.8cm]{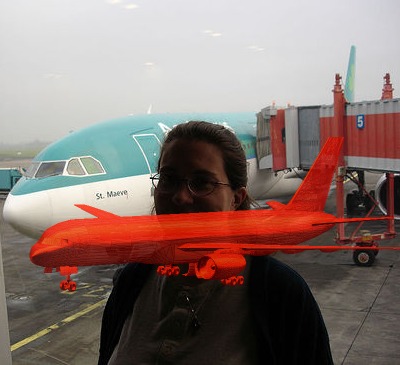}
\includegraphics[height=1.8cm]{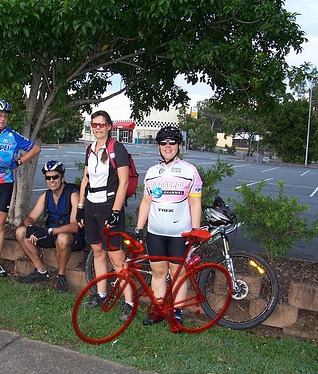}
\includegraphics[height=1.8cm]{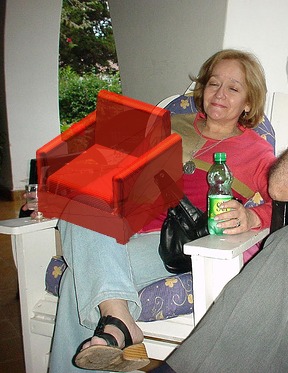}
\includegraphics[height=1.8cm]{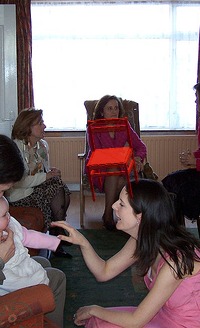}~
\includegraphics[height=1.8cm]{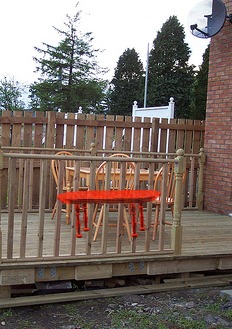}
\\
\includegraphics[height=1.8cm]{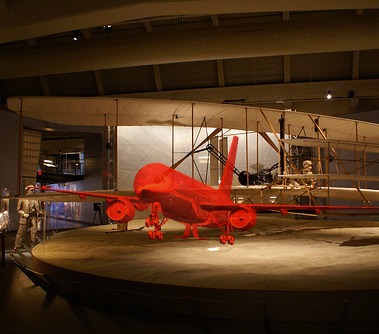}~
\includegraphics[height=1.8cm]{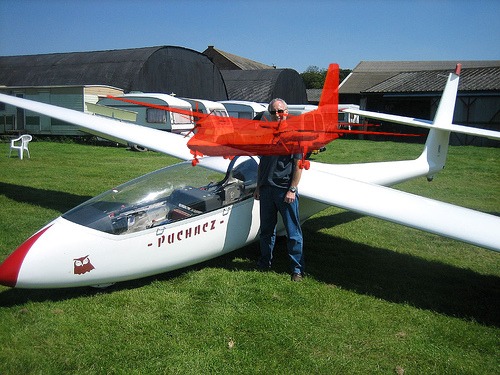}~
\includegraphics[height=1.8cm]{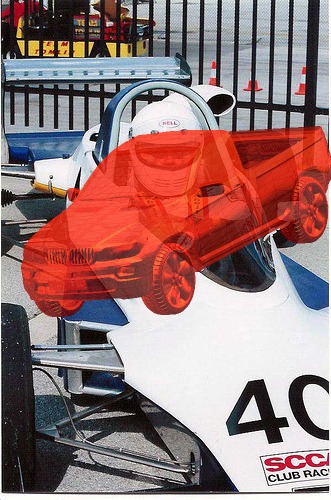}~
\includegraphics[height=1.8cm]{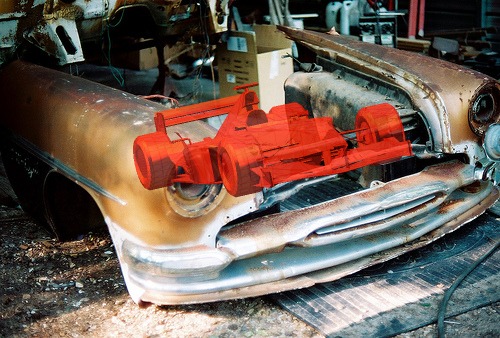}~
\includegraphics[height=1.8cm]{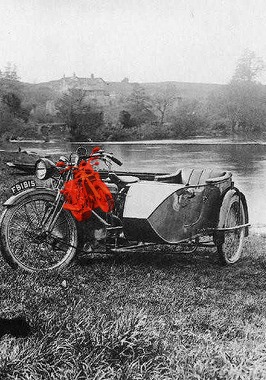}~
\includegraphics[height=1.8cm]{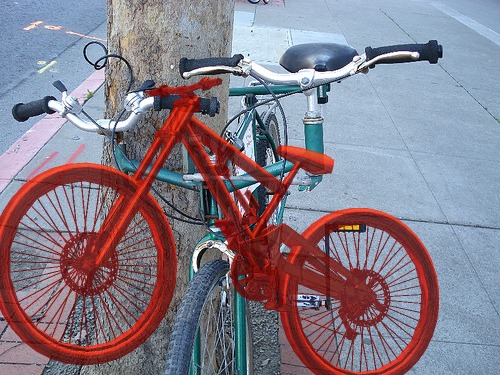}~
\includegraphics[height=1.8cm]{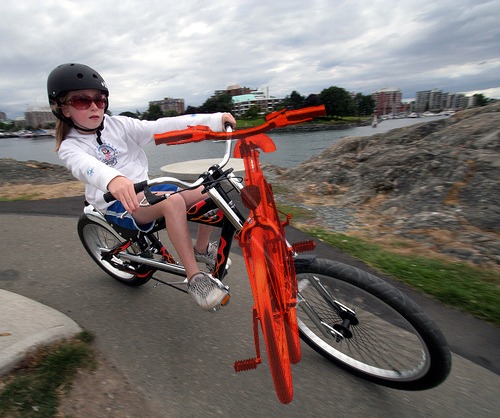}~
\includegraphics[height=1.8cm]{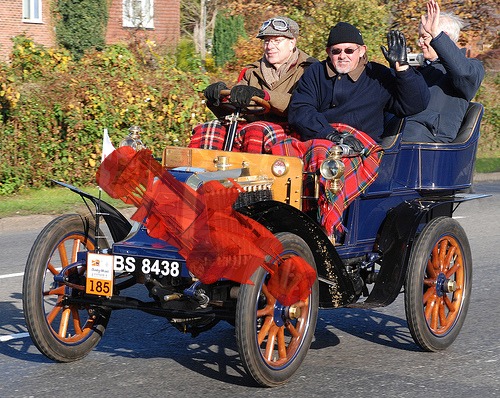}
\\
\includegraphics[height=1.8cm]{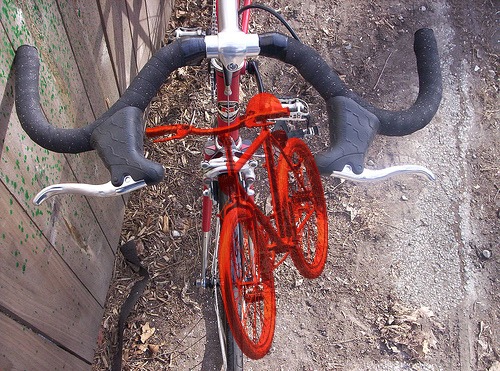}~
\includegraphics[height=1.8cm]{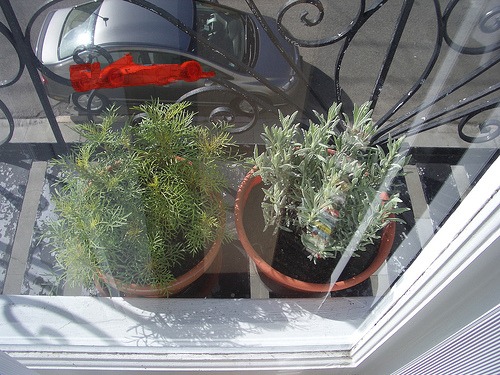}~
\includegraphics[height=1.8cm]{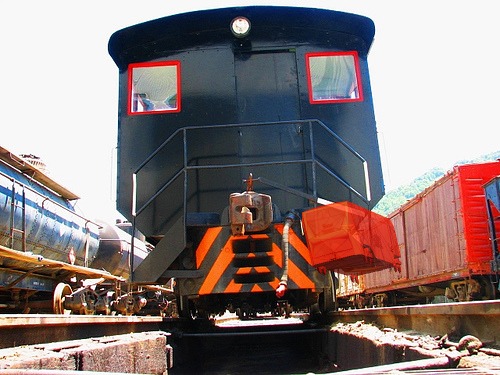}~~
\includegraphics[height=1.8cm]{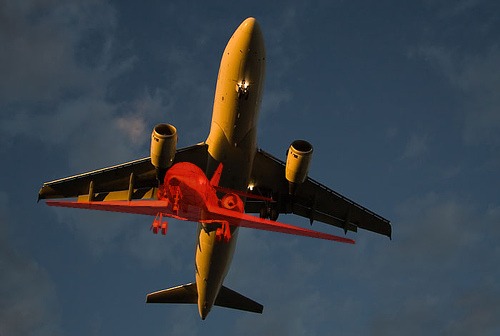}~~
\includegraphics[height=1.8cm]{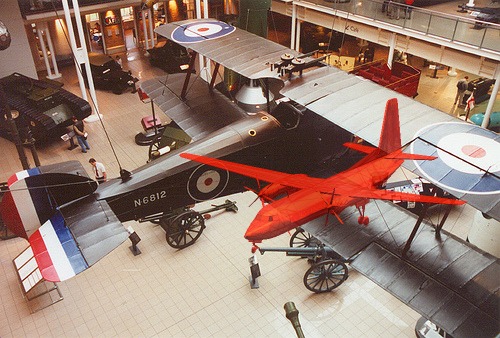}~~
\includegraphics[height=1.8cm]{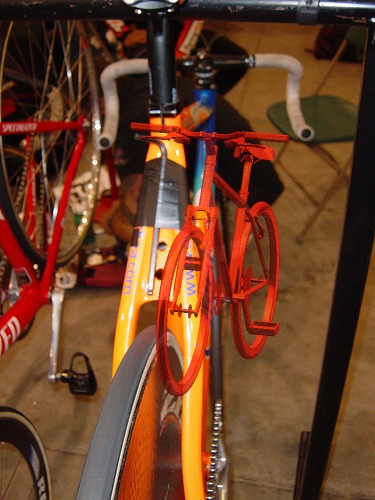}~
\includegraphics[height=1.8cm]{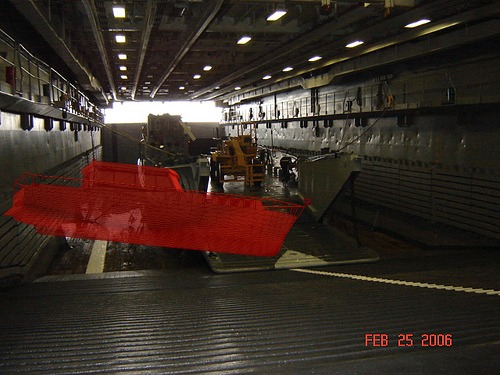}
\\
\includegraphics[height=1.8cm]{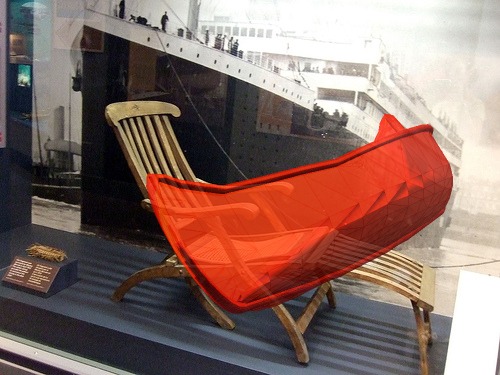}~
\includegraphics[height=1.8cm]{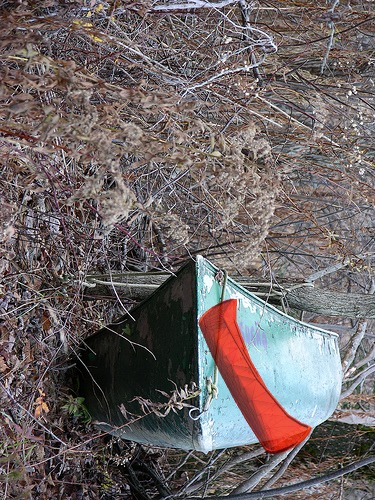}~
\includegraphics[height=1.8cm]{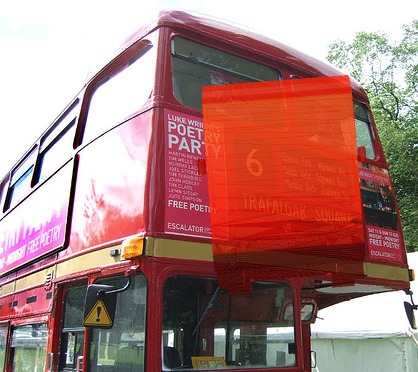}~~
\includegraphics[height=1.8cm]{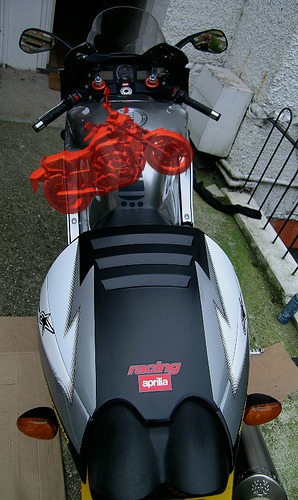}~
\includegraphics[height=1.8cm]{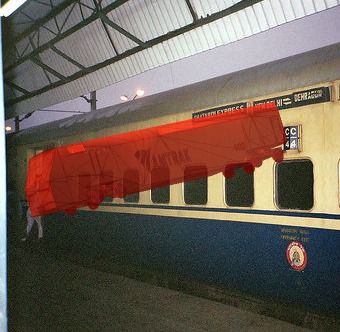}~
\includegraphics[height=1.8cm]{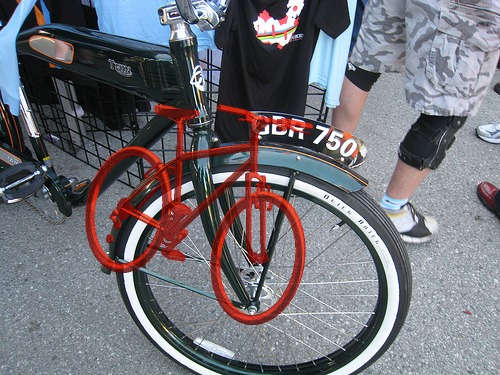}~~
\includegraphics[height=1.8cm]{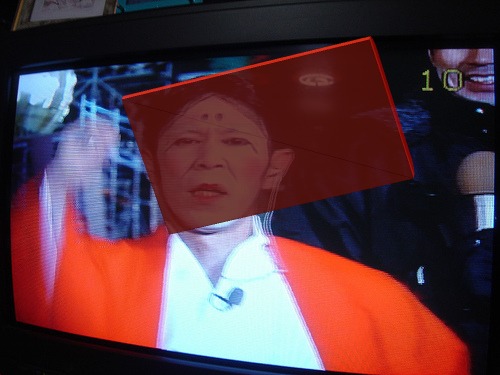}~
\includegraphics[height=1.8cm]{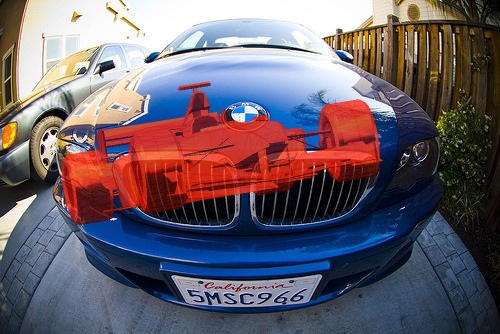}
%
%
%
%
%
%
%
%
%
\caption{3D CAD prototype alignment examples on all Pascal3D+
  classes. The 3D alignment fails mostly due to truncations (row 1 and
  2), occlusions (row 3 and 4), unusual shapes (row 5), viewpoints
  (row 6) and strong perspective effects for objects close to the
  camera (row 7).}
\label{fig:supplBadExamples}
\end{figure*}

Fig.~\ref{fig:supplAAVP} provides AVP vs azimuth error curves,
comparing the 3D lifting methods \rcnnRidgeL, \rcnnl and \keyReg with
the best \vdpm, \dpmvocvp, \rcnnmv, \cnnmv methods and \rcnnRidge in
terms of the AAVP measure (averaged AVP). As seen in Fig.~\ref{exp:vp} (center)
in the paper, \rcnnRidgeL with $35.5\%$ mAAVP has the edge over
\rcnnRidge $35.3\%$. The improvement comes due to the better
performance on the high-precision region $\le 40^{\circ}$.  In
Fig.~\ref{fig:supplAAVP} we observe this tendency across all classes,
except for {\em tvmonitor} where \rcnnRidge ($52.4\%$ AAVP) is
slightly better than \rcnnRidgeL ($51.1\%$). In comparison to the
previous state-of-the-art \rcnnRidgeL is better than \dpmvocvp-16V by
$12.5\%$ (mAAVP) on average. It is also consistently better on all the
classes, thus achieving state-of-the-art performance on Pascal3D+ in
terms of the new AAVP measure as well.

Fig.~\ref{fig:supplGoodExamples} illustrates example 3D CAD prototype
alignments based on the \rcnnRidgeL. While being able to detect the
correct 3D object shape in most of the cases, \rcnnRidgeL also
succeeds in predicting the 3D viewpoint of the objects and can capture
variations in the 3 rotation parameters (azimuth, elevation and
in-plane rotation).

Fig.~\ref{fig:supplBadExamples} illustrates failure cases of the 3D
CAD prototype alignment. Note that the majority of failure cases are
truncated (rows 1 and 2) and occluded objects (rows 3 and
4). \rcnnRidgeL also fails when unusual 3D shape (row 5) or viewpoint
(row 6) are encountered. Objects close to the camera are also an issue
(row 7).

\end{document}